\documentclass[conference]{IEEEtran}
\IEEEoverridecommandlockouts
\usepackage{cite}
\usepackage{amsmath,amssymb,amsfonts}
\usepackage{algorithmic}
\usepackage{graphicx}
\usepackage{textcomp}
\usepackage{xcolor}
\usepackage{float}
\usepackage[caption=false,font=footnotesize]{subfig}
\usepackage{hyperref}
\def\BibTeX{{\rm B\kern-.05em{\sc i\kern-.025em b}\kern-.08em
    T\kern-.1667em\lower.7ex\hbox{E}\kern-.125emX}}

\newcommand{\mycomment}[1]{}

\newcommand{\sm}[1]{\textcolor{black}{#1}}

\makeatletter
\newcommand{\linebreakand}{%
  \end{@IEEEauthorhalign}
  \hfill\mbox{}\par
  \mbox{}\hfill\begin{@IEEEauthorhalign}
}

\begin{document}

\title{TeLL-Me what you can't see. \\
\thanks{This work was supported by the European Union - NextGenerationEU as part of [anonymous].
We thank the partnership and collaboration with the [anonymous] Research Foundation.
}
}

\author{\IEEEauthorblockN{Saverio Cavasin}
\IEEEauthorblockA{\textit{dept. of Mathematics} \\
\textit{University of Padua}\\
Padua, Italy \\
saverio.cavasin@phd.unipd.it}
\and
\IEEEauthorblockN{Pietro Biasetton}
\IEEEauthorblockA{\textit{dept. of Mathematics} \\
\textit{University of Padua}\\
Padua, Italy \\
pietro.biasetton@studenti.unipd.it}
\and
\IEEEauthorblockN{Mattia Tamiazzo}
\IEEEauthorblockA{\textit{dept. of Mathematics} \\
\textit{University of Padua}\\
Padua, Italy \\
mattia.tamiazzo.1@studenti.unipd.it}
\linebreakand 
\IEEEauthorblockN{Simone Milani}
\IEEEauthorblockA{\textit{dept. of Information Engineering} \\
\textit{University of Padua}\\
Padua, Italy \\
simone.milani@dei.unipd.it}
\and
\IEEEauthorblockN{Mauro Conti}
\IEEEauthorblockA{\textit{dept. of Mathematics} \\
\textit{University of Padua}\\
Padua, Italy \\
mauro.conti@unipd.it}
}

\maketitle

\begin{abstract}
During criminal investigations, images of \sm{persons} of interest directly influence the success of identification procedures. However, law enforcement agencies often face challenges related to the scarcity of high-quality images \sm{or their obsolescence}, which can affect the accuracy and \sm{success of people searching} processes.

This paper introduces a novel forensic mugshot augmentation framework aimed at addressing these limitations. Our approach enhances the identification probability of individuals by generating additional, high-quality images through customizable data augmentation techniques, while maintaining the biometric integrity and consistency of the original data. 

Several experimental results show that our method significantly improves identification accuracy and robustness across various forensic scenarios, demonstrating its effectiveness as a trustworthy tool law enforcement applications.
\end{abstract}


\section{Introduction}

    Law enforcement agencies rely heavily on images in identification procedures to localize suspects and missing people. 
    
    Usually data is shared in posters of Person-Of-Interests (POIs), which are mostly structured through a photograph and a list of physical attributes). These collections are developed from a \sm{visual} extraction of information about the depicted subject \sm{performed by a human police operator and supported by witness interviews \cite{5601735}}.
    \sm{Investigations are therefore influenced by the number, quality, and meaningfulness of the visual evidences that the police were able to collect \cite{app14209285}}. 
    If insufficient pictures are available, there is a higher chance for mistakes and confusion. 
    Nevertheless, the nature of these close-up images implies several issues in the recovery of valuable evidence. 
    As a result, \sm{ only a limited number of images turn out to be useful in law enforcement investigations}.

\sm{During the last years, some research works have adopted deep learning (DL) solutions for the enhancement and refinement of forensic sketches and image quality \cite{10744467,app14209285}. Such approaches often rely on a generative network that converts a single visual input into one or more output. The data-driven nature of DL solutions introduces several potential artifacts ranging from a mismatching samples (generated images do not match with the searched samples) to bias and fairnesses issues. In this work, we aim at mitigating such drawbacks by combining multiple picture generative networks with Vision-Language Models (VLMs): the ensemble of multiple generators allows for a better generalization and denoising of the input image, while VLM-generated text prompts enforces the robustness and fidelity of the final samples with respect to persons' ID. } 

To the best of our knowledge, this work introduces the first comprehensive forensic analysis focused on improving witness consultations using mugshot augmentation through \sm{VLMs.} 

Our key contributions are summarized as follows.

\begin{enumerate} 

\item We designed a forensic pipeline to enhance witness consultations from mugshots and generic images, aimed for criminal identification and kidnapping searches.

\item We are the first to propose an automatized forensic analysis \sm{by combining multiple generative networks and Visual Language Models (VLMs), thus} allowing real time guidance for prompt/feature selection and generalization in synthetic image augmentation. 

\item We propose a custom semantic Hamming-like clustering distance to evaluate similarity dynamics and study accuracy and potential biases in the generated VLM descriptions.

\item We developed the first synthetic aging system tailored for forensic and investigative applications, based on LLM prompt optimization and VLM image augmentation to synthetize aging.

\item We tested multiple methods and models for each task, providing empirical baselines and benchmarks. 

\end{enumerate}

    

    In Section~\ref{sec:related_works} we introduce the context of state of the art forensic sketching and face recognition in police investigations;
    Section~\ref{sec:method} describes the collected dataset, the overall structure of the tool, and the models employed by each component;
    in Section~\ref{sec:experiments} we present the experiments performed on the acquired data, the relevant forensic scenarios and discuss the results;
    Section~\ref{sec:conclusion} draws our conclusions, considers the limitations of this work and proposes possible further research orientations and use cases. Eventually, in the Appendix section we show additional evidences and comparisons of our experiments and results.
    
\section{Related Work}
\label{sec:related_works}
    Digital media plays a major role as forensic evidence. It usually provides remarkable insights from sources like surveillance cameras' footage, social networks, and mobile carry-on devices.
    The use of these evidence helps the reconstruction of crime scenes and the identification of the suspects. 
    To better contextualize our work, we will introduce why Generative AI marks an undeniable contribution in this context.   
    
    Witness based forensic sketching is composed by four areas of interest~\cite{forensic_art}: 
        \begin{enumerate}
            \item \textbf{Composite Imagery}: graphic images made up combining individually described physical components. 
            \item \textbf{Image Modification and Identification}: methods of manipulation, enhancement, comparison, and categorization of images.
            \item \textbf{Demonstrative Evidence}: visual supports for case presentation in trials.
            \item \textbf{Reconstruction and Postmortem Identification Aids}: methods to aid in the identification of human physical remains.
        \end{enumerate}
        In this work we will focus on the first and the second.
        Composite imagery and image modification are usually performed by a forensic artist producing a drawing or a digital picture.
        This spread method nevertheless introduces downsides, such as the obvious requirement of a skilled artist as the lack of accuracy likely leads to identification failures.

        Some Computer Vision and Artificial Intelligence assistant have been proposed in the field of forensic art.
        The work from ~\cite{sketchgen1, sketchgen2} allow generating a realistic forensic composite by setting manually the features of the subject, including age, skin color and other physical properties. Starting from a sketching image of the individual, ~\cite{sketchtoimg1, sketchtoimg2, sketchtoimg3} focused on realism improvement. \sm{Other solutions aims at improving the matching probability  by generating multiple different sketch samples from a single low-quality picture  using diffusion models \cite{10744467}}. 
        
        \sm{Indeed, most of the existing solutions aims at characterizing POI's face more accurately via multiple views or skecth generation. So far, ensembles of textual descriptions combined with multiple visual generators was not considered.} \sm{ This approach permits optimizing the generation of realistic pictures of wanted or missing individuals and, at the same time, preventing potential inaccuracies caused by images' manual inspection and classification.}
        Our use of Vision-Language Models (VLMs) allows for the automatic extraction of an individual's physical features from a picture. 
        Moreover, while the production of forensic sketches traditionally requires the expertise of a specialized forensic artist, our proposed models enable the direct generation of new synthetic, high-quality pictures of the subject, preserving their original identity.
        Our framework also enables several modifications to the individual's original physical appearance, such as aging simulation, weight loss or clothing changes.  

        Face recognition has been explored since the 1970s~\cite{facerecog1}, and it is usually based on holistic methods, which aim to extract facial features using Computer Vision techniques.
        
        Recently, the development of Deep Neural Networks has allowed the production of several verification algorithms ~\cite{facerecog2}, which generally outperform older methods, despite implying some bias and limitations \cite{biasface}.

        Synthetic image generation models are able to produce a specific picture by iteratively deblurring an image initialized with Gaussian noise~\cite{diffusionmodels}.
        A notable example is represented by Stable Diffusion~\cite{stablediffusion}, an open-source network specialized in the generation of images from textual prompts. 

        Age-modeling in images have advanced significantly with the use of Deep Learning as well. 
        One prominent approach is the \mbox{Age-cGAN} model~\cite{antipov2017face}, which learns age transformations by using conditional GANs to control aging while maintaining identity consistency. 
        Additionally, large-scale datasets like \mbox{FG-NET}~\cite{panis2016overview} have been crucial in training these models, offering a wide range of facial images across different ages.

\section{Method}
\label{sec:method}

    The \sm{proposed} mugshot-generating technology \sm{can be described by the following key modules:}
    \begin{itemize}
        \item Image enhancement: the quality of the original picture \sm{is improved by means of multiple denoising generative strategy that aim at compensating several quality-impairing elements affecting the original image.}
        \item Linguistic description: \sm{ a textual physical description of the subject is extracted using a LLM in order to guide the enhancement of the final picture}. 
        \item Image augmentation: \sm{LLM output is used as a prompt to guide the VLM image generation (i.e., the new augmented mugshot)}.
    \end{itemize}

    \begin{figure}[ht] 
        \centering
        \includegraphics[width=0.7\linewidth]{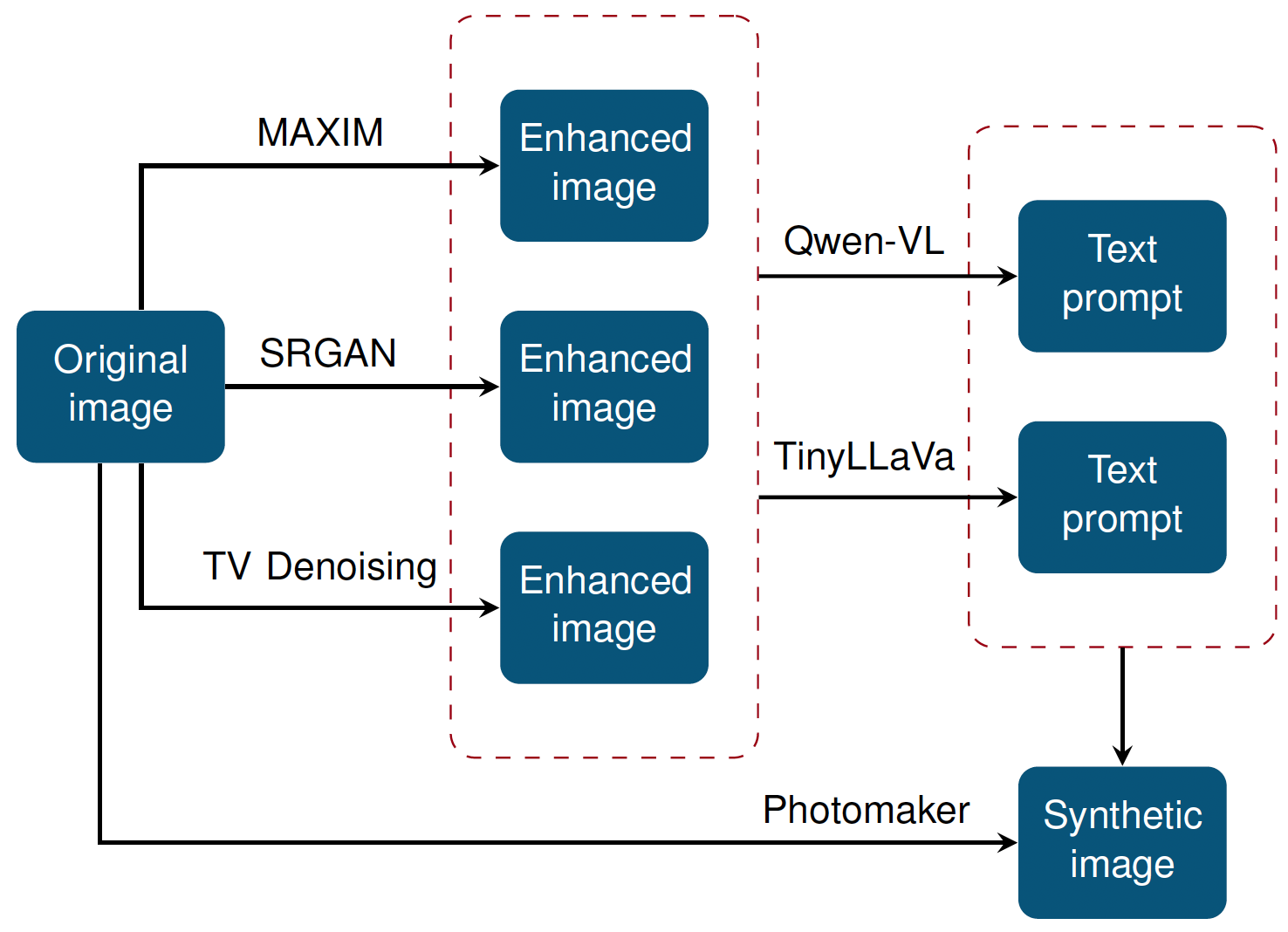}%
        \caption{Automatic synthetic mugshot generation.}       
        \label{fig:strategy}
        \vskip -2ex
    \end{figure}
    
    \subsection{Dataset}
\begin{figure}
            \centering
            \includegraphics[width=.7\linewidth]{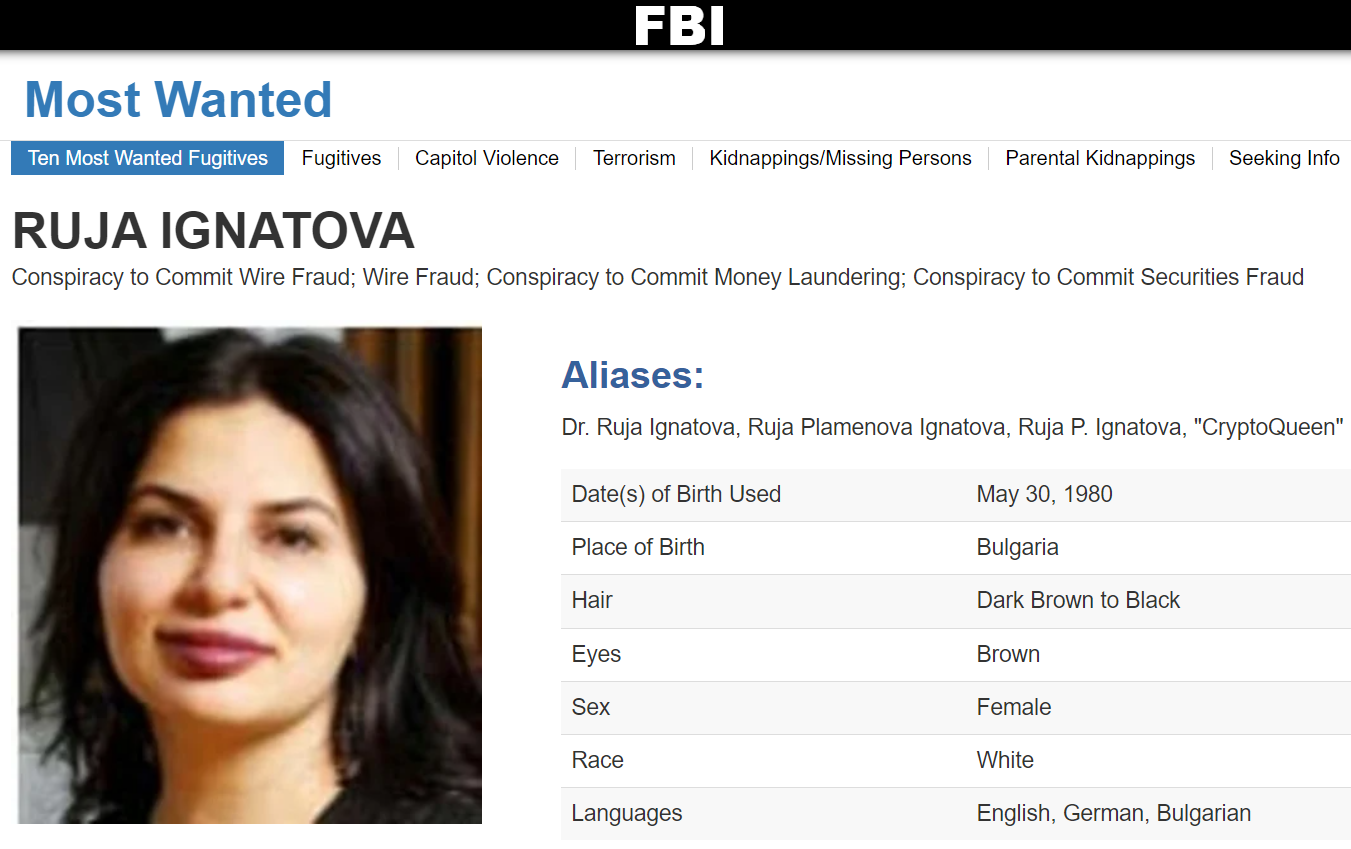}
        \caption{Example of a FBI wanted poster}\label{fig:screenshots_a}
        \vskip -2ex
        \end{figure}

        The images used in developing the tool are mugshots of missing or wanted individuals shared by intelligence and security organizations, like INTERPOL and FBI.
        Law enforcement agencies typically have access to only a few images of the subject, and often their quality is limited.
        To replicate this scenario, we selected low-to-average quality photographs of people (e.g., small pictures, low resolution, noise or compression artifacts, etc.) for which their physical information (forensic categorization) is also available as a benchmark. 
        As shown in \figurename~\ref{fig:screenshots_a}, intelligence agencies specifically summarize pictures and categorization of a relevant subject in mugshot posters. 
              
\begin{figure}
            \centering
            \vspace*{-1ex}\subfloat{\includegraphics[width=.2\linewidth]{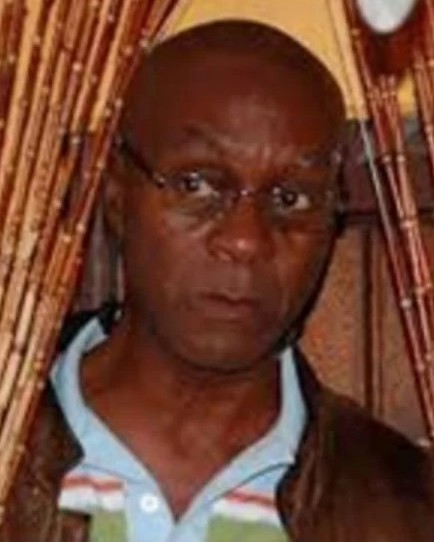}} \quad
            \vspace*{-1ex}\subfloat{\includegraphics[width=.2\linewidth]{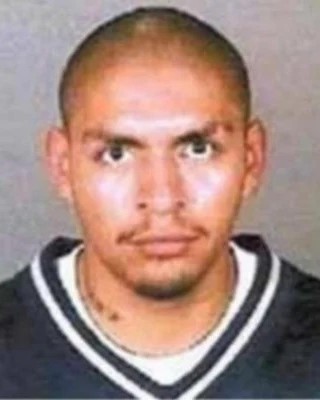}} \quad
            \vspace*{-1ex}\subfloat{\includegraphics[width=.2\linewidth]{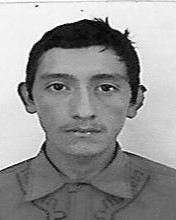}}             
            \caption{Actual mugshots employed in this work.}
            \label{fig:mushots}
            \vskip -2ex
        \end{figure}

        To mitigate biases in the dataset, we selected images of individuals with diverse physical characteristics, including gender, age, and ethnicity.   
        Some examples of mugshots also for this work are shown in \figurename~\ref{fig:mushots}.

        Finally, to investigate the aging ability of our methods, we selected a second dataset specifically for testing this task. We compare our artificial aging system with two different real pictures. One representing the young version and one representing the actual old one.
            Rarely this kind of information is easily accessible for convicted criminals or missing people, so we rooted for “famous” people as celebrities childhood images are frequently available.
            \\
            We spaced within several levels of importance of these “VIPs” to avoid specific training biases of the larger pretrained networks present in our framework (as they are often overtrained on open source data). 

    \subsection{Image enhancement techniques}
        In order to improve the quality of the input pictures, we compared three signal processing methods for image enhancement, addressing different image processing tasks. 
        
        \paragraph{MAXIM} 
            This model~\cite{maxim} is a multi-axis Multi-Layer Perceptron (MLP) for image processing that we tasked as a denoiser via light enhancement.

        \paragraph{Total Variation Denoising}
            This method is based on the principle that reducing the total variation of signals leverages the removal of unwanted details. ~\cite{tvdenoise}:
            \[
                V(y) = \sum_{i,j} \lvert y_{i+1,j} - y_{i,j} \rvert + \lvert y_{i,j+1} - y_{i,j} \rvert.
            \]
            
        \paragraph{SRGAN} Estimates high-resolution images from low-resolution counterparts via a SRGAN~\cite{srgan} for image super-resolution.
            Discriminator trains using an adversarial loss to differentiate between the super-resolved images and the original pictures, while a content loss is used to ensure perceptual similarity. 
            
    \subsection{Vision-Language Models}
        The second component of our tool is represented by the linguistic description of the enhanced images, implemented via a Vision-Language Model.
        This task aims to \textit{automatically} extract physical information about the subject from one picture, without required explicit previous knowledge of their identity.
        We introduce these results in Section~\ref{semantic distance}. 
        
        We compared two small-scale pretrained VLMs for this task:
        
        \paragraph{Qwen-VL}
            The model~\cite{qwenvl} is composed by a pretrained vision encoder, a Vision-Language adapter, and a Large Language Model (Qwen-7B).

            \

        \paragraph{TinyLLaVA}
            The model~\cite{tinyllava} is a small-scale large-multimodal architecture, composed by a pretrained vision encoder and a small-scale LLM, connected by a two-layer
            Multi-Layer Perceptron (MLP).
    
    \subsection{Image augmentation}
        In the final stage of our method, the image of the subject is used to create new synthetic pictures.
        In real-world scenarios, law enforcement officers may aim to produce new images of certain, relevant, individuals. 
        This is to maximize and improve the amount of information at their disposal, and additionally they may even incorporate slight changes to the known physical appearance.
        
        For this task we tuned PhotoMaker~\cite{photomaker}, a state of the art open source generative model that produces synthetic pictures while preserving the identity of the subjects, guided through the input text prompts.
        This model leverages an image encoder to represent the original picture in the image embeddings, and a text encoder to produce the embedding of the input prompt.
        Each image embedding is then fused using a two-layer MLP with the corresponding class extracted from the text embedding. 
        All the combined embeddings are concatenated to form the \textit{stacked ID embedding} that is fed to all cross-attention layers of a diffusion model based on the Stable Diffusion SDXL model~\cite{sdxl}, to generate the synthetic photo-realistic human portrait.   

        \begin{figure}[ht] 
            \centering
            \includegraphics[width=0.8\linewidth]{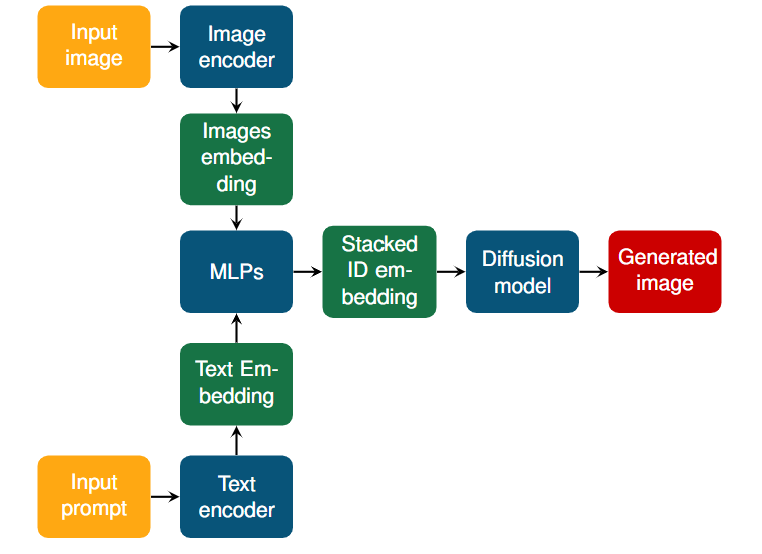}%
            \caption{PhotoMaker structure scheme. In the scheme orange is used for the network's inputs, blue is used to highlight different networks composing the model, green stands for different embeddings and lastly red is used for output.}
            \label{fig:Photomaker}
             \vskip -2ex
        \end{figure}

\section{Experiments}
\label{sec:experiments}

    We considered the outputs of the VLMs on the set of the images and the corresponding additional enhancements.

    To validate the accuracy and precision of such results with respect to the original data, we propose a metric function developed as a “semantic” Hamming distance, which is able to compare both numerical and string results.
    
    In order to generate new pictures during the image augmentation phase, we tested several combinations of prompt and images to obtain complete and visible faces. 
    We tuned the prompt engineering without affecting the quality of the images, and keeping the allowed length limit.
    
    \subsection{Hyperparameters tuning}
    \label{hyperparameters setting}
        
        We fine tuned some of the discussed networks to adapt them for our task. 
        
        In the Total Variation Denoising, the number of iterations was set to 200 empirically. 
        Less iterations implied the lack of a significant improvement for the images, while higher number of iterations caused the loss of peculiar facial features, like wrinkles or freckles.
        
        For MAXIM, we set the \textit{Task} hyperparameter to \textit{Enhancement}, in order to get a model operating with light exposure rather than focusing on denoising or deblurring.
    
        To tune our PhotoMaker instance we set the the number of sample steps to 50 and and the percentage of style strength 20. 
        They influence the ID fidelity of the generated images. 
        A lower value for the first leads to an exaggerated similarity with the original, which is not valuable, since our purpose is to increase the amount of information in our possess. 
        Higher number of sample steps, considerably increased the computational complexity of the process (without providing justifiable improvement).
        
    \subsection{Semantic Hamming distance} 
    \label{semantic distance}
        To evaluate the VLMs, we developed a metric to compare true data (given by the sources) and the results of the linguistic description phase. 
        Such metric allows us not only to compare the two, but also to infer how useful the different enhancements are. 
        During the description phase performed through VLMs, we input prompts containing a list of questions accordingly with the kind of information commonly asked in the process of facial composite production of identi-kit sketching ~\cite{forensic_art}. 
        
        According to the information required in official posters, we took into consideration the following categories: gender, age, ethnical group, hair color, iris color, height, and weight. 
        Therefore, our metric focuses on these categories. 
        We assign interval distances from 0 (corresponding to the case of same information in both sources and prediction), to 1 (corresponding to the case of completely different information).

        \
        
        \paragraph{Numerical values}
            After a preprocessing phase, required to enable the comparison between different units of measure, a threshold $t$ was set for each category belonging to the group (age, height, weight). 
            We therefore defined the real value $a$, the predicted value $b$, and the tolerance value $h =\frac{t}{4}$. 
            We distinguish three different cases:
            \begin{enumerate}
                \item If $\lvert a-b \rvert < h$: the distance is set to $0$.
                \item If $h<\lvert a-b \rvert < t$: the distance is set to $\dfrac{\lvert a-b \rvert - h}{t-h}$.
                \item If $\lvert a-b \rvert > t$: the distance is set to $1$.
            \end{enumerate}

            \
           
        \paragraph{String values}
            When considering string-type data, we focus on semantic similarities basically providing a clustering method. 
            We removed punctuation, capital letters, and summarized synonyms (e.g., in the ethnic group category, the word \textit{Caucasian} was replaced with \textit{White} which is more used by both the Agencies and VLMs). 
            For all categories apart from gender, which already has a binary nature that avoids midway results, we established some labels groupings where distance was set to be less than 1. 
            For the ethnic group category, we established four different clusters, based principally on the similarity of complexion.
            An interesting example in this context may be the label \textit{Hispanic}. 
            Being quite broad, it can be confused with one of the labels among \textit{White}, \textit{Arab}, and \textit{Indian}.
            Official posters often indicate such labels as associated, supporting our structure. 
            Concerning hair color, we established three groups based on their distance in the color spectrum. Following the same criteria, we established three groups for the iris too. 
            We report these groups in Table~\ref{tab:groups}.
            
            \begin{table} 
                \centering
                \caption{Groups used to set distances between elements, often easily misclassified.}
                \label{tab:groups}
                \begin{tabular}{|c|c|}
                    \hline
                        \multicolumn{2}{|c|}{\textbf{Ethnic groups}}              \\
                    \hline
                        \multicolumn{2}{|c|}{African American, African, Aboriginal} \\
                        \multicolumn{2}{|c|}{White, Hispanic}                       \\
                        \multicolumn{2}{|c|}{Hispanic, Arab}                        \\
                        \multicolumn{2}{|c|}{Hispanic, Indian}                      \\
                    \hline
                        \textbf{Hair color}     &   \textbf{Iris color}             \\
                    \hline
                        {Black, Brown}          &   {Black, Brown}                  \\
                        {Blonde, Light brown}   &   {Blue, Green}                   \\
                        {Brown, Light brown}    &   {Green, Brown}                  \\
                    \hline
                \end{tabular}
                \vskip -2ex
            \end{table}
            Our distance is then set according to the following cases:
            \begin{itemize}
                \item Original label and predicted label match: distance is set to be 0.
                \item Original and predicted are different, but belong to the same "cluster": the distance is set to 0.5.
                \item Original label and the predicted are not even in the same group: the distance is set to be 1.
            \end{itemize}

        \paragraph{Accuracy computation}
            Once the measurement of the subject was performed for every selected category, the final distance score was computed as their average.
            The prediction score was finally computed as the percentage of accuracy in the VLM prediction of the subject's features.
            This score has been computed for every picture of the subjects, included the enhancements.

    \subsection{Prompt engineering}
    \label{prompt_eng}
        To test the network for data augmentation, we set a common prompt scheme that enables the user to obtain clearly visible images with the least possible bias. 
        We tested how the generation of new images could be affected by including or not (in the input prompt) different features suggested in the linguistic description phase. 
        We excluded the features not directly concerning the physical appearance of the person from the prompt (e.g., the expression of the eyes, teeth visibility, or clothing). 
        
\begin{figure}
            \centering
            \subfloat[Nose focus]{%
                \includegraphics[width=.2\linewidth]{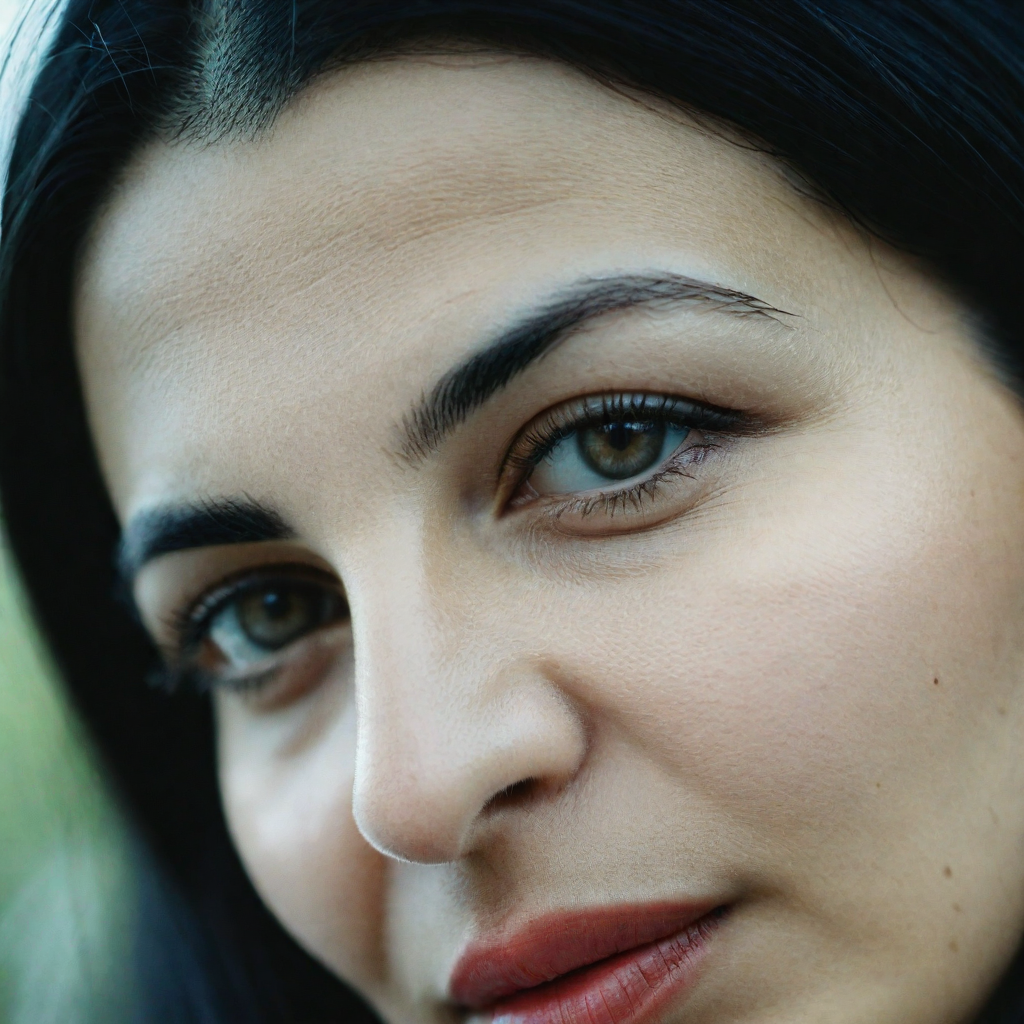}%
                \label{fig:errors_a}}   \quad
            \subfloat[Eyes focus]{%
                \includegraphics[width=.2\linewidth]{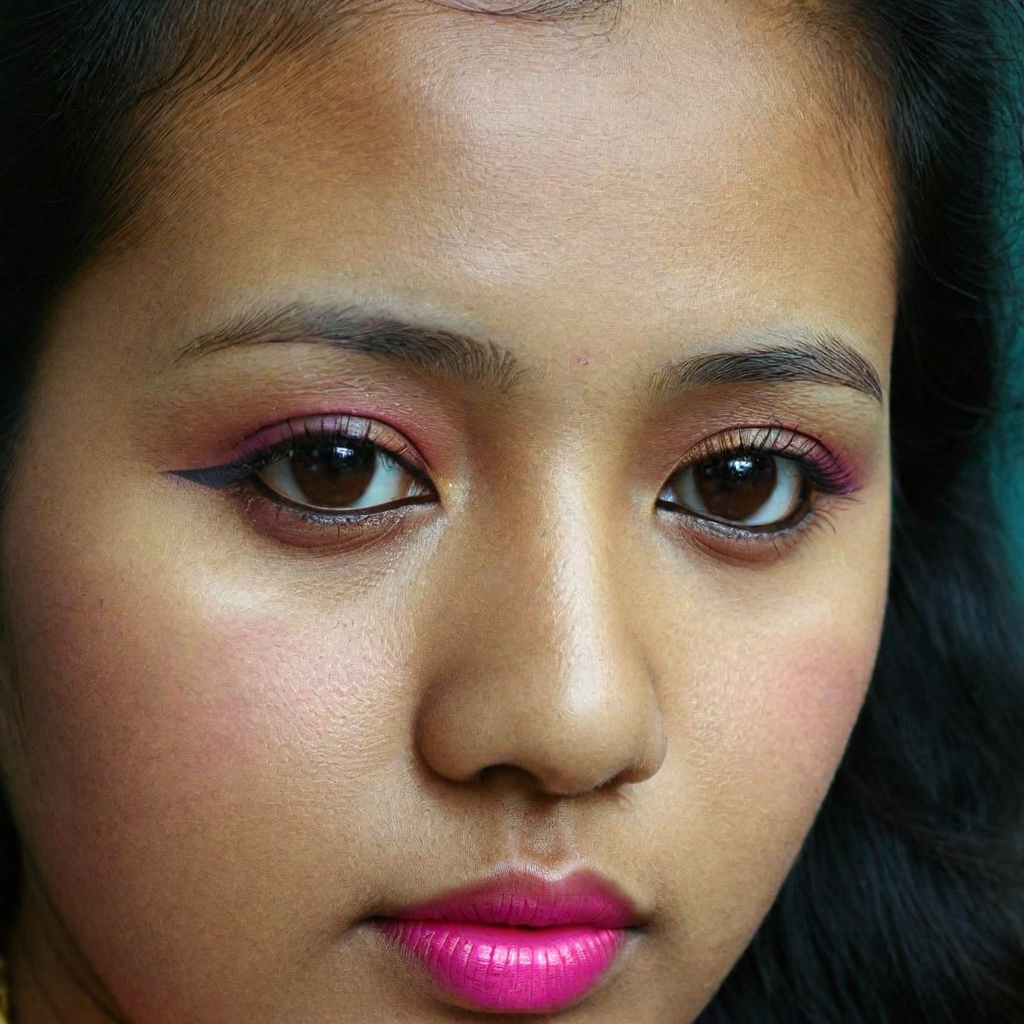}%
                \label{fig:errors_b}}  \quad
            \subfloat[Nose focus]{%
                \includegraphics[width=.2\linewidth]{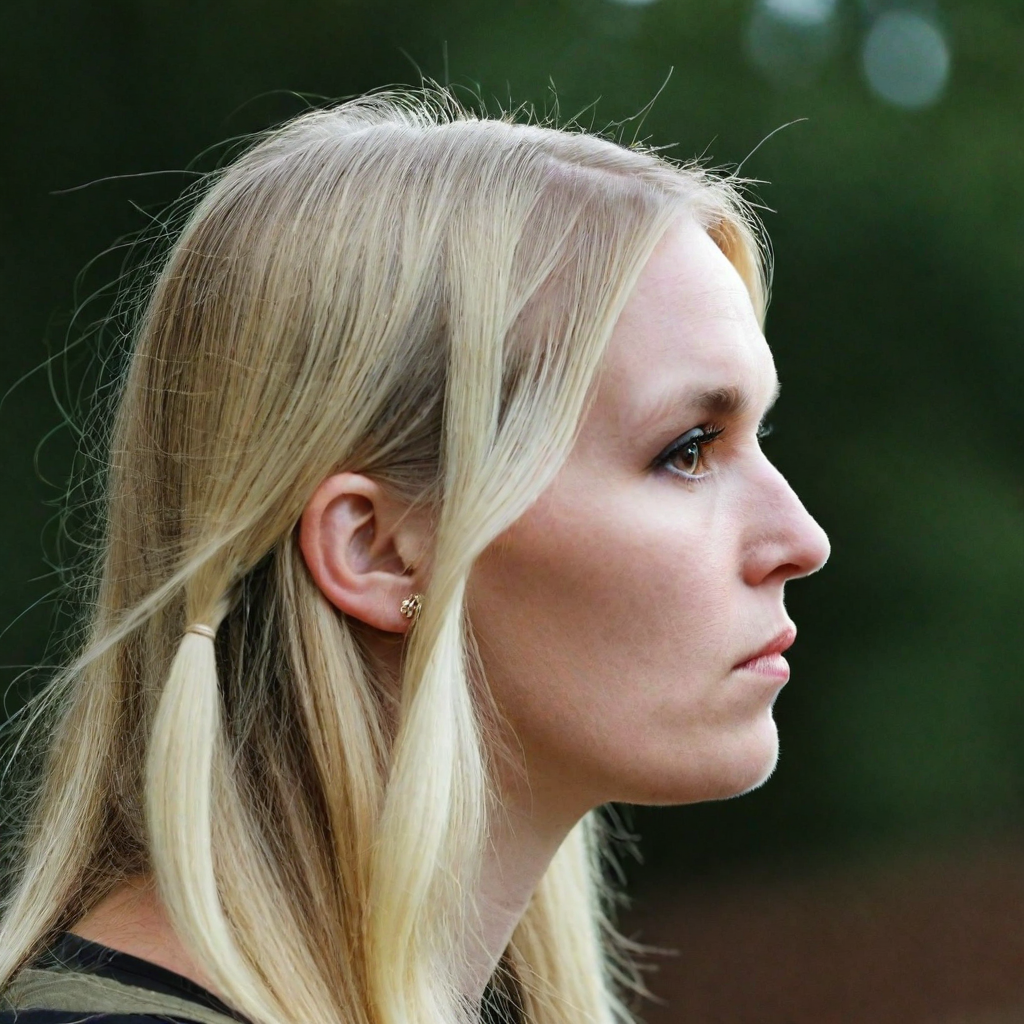}%
                \label{fig:errors_c}}   \quad  
            \subfloat[Beard focus]{%
                \includegraphics[width=.2\linewidth]{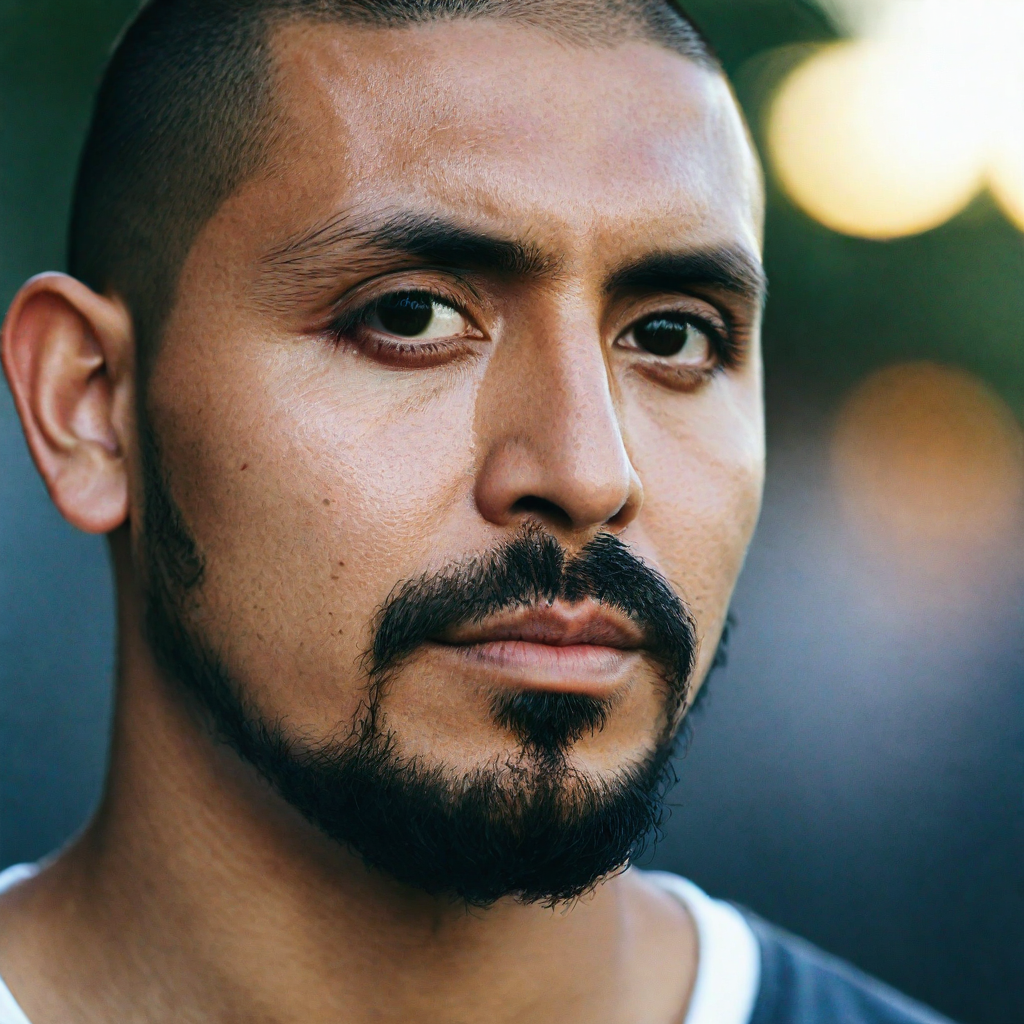}%
                \label{fig:errors_d}}  
            \caption{Specific features generation}
            \label{fig:errors}
            \vskip -2ex
        \end{figure}

        We noticed that the inclusion of overly specific information (e.g eyes and nose) led to close up pictures of the subject, focusing on the details included in the prompt and causing losses in the overall appearance of the individual, as shown in \figurename~\ref{fig:errors}. 
        Similarly, including information concerning the ears provided side profiles mugshot (could be useful, but lacking a-priori generalization). 
      
        Interestingly, facial hair also over proved to shift the output from the required information. Omitting such details in the text prompts allowed for the generated pictures to be rendered in a more faithful manner with respect to the original picture and requirements.
        
        After setting these guidelines, we collected the required data: gender of the subject, age, the ethnic group to which the individual belongs, and the information about hair (in particular length and color). 
        These elements are the ones that proved to affect the output image the most, while not disrupting the overall structure of the pictures.
        
    \subsection{Re-Identification Models}

    To test the similarity between the original samples and our generated information, we employed a chain of  processing units, developed for face recognition and implemented in the following \textbf{DeepFace} architecture ~\cite{serengil2020lightface,serengil2024lightface}:

    \begin{enumerate}
        \item \textbf{Preprocessing and Alignment}. Face normalization, pose and angle correction, facial orientation. This step ensures that faces are captured frontally, for recognition accuracy. 
        \item \textbf{Resizing} provides uniformity to CNNs used in the face detection.
        \item \textbf{Feature Representation} generates a high-dimensional vector (embedding) of the face, capturing unique characteristics.
        \item \textbf{Classification} via similarity metrics (e.g., euclidean), to determine whether two faces match based on the embeddings. 
    \end{enumerate}

The Face detection module we applied is \textbf{FaceNet}~\cite{schroff2015facenet}. It is highly based on Deep-learning classification and it makes use of a Euclidean embedding~\cite{globerson2004euclidean}  of faces that ensures that similar faces share a lower distance while different faces have a bigger one. This result is achieved through the use of a triplet loss~\cite{weinberger2009distance}. 

The other module employed in our tests for comparison is Recognito~\cite{recognito}, a facial recognition system largely based on pre-exiting models like DeepFace. Such model was selected because its algorithm scored among the top systems in both 1:1 verification and 1:N identification scenarios during the Face Recognition Technology Evaluation by NIST~\cite{FRTE}.

    \subsection{Results and confrontation} \label{results}
        We discuss our results distinguishing our 3 main phases: the enhancement and linguistic description of the pictures, the standard mugshot augmentation, and the aging transformation. Additional visual examples can be observed in Appendix B. 
        
\begin{figure}
                \centering
                \subfloat[Original]{%
                    \includegraphics[width=.2\linewidth]{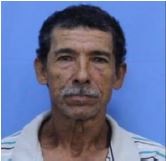}%
                    \label{fig:enhancements_a}} \quad
                \subfloat[MAXIM]{%
                    \includegraphics[width=.2\linewidth]{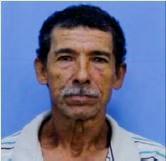}%
                    \label{fig:enhancements_b}} \quad
                \subfloat[SRGAN]{%
                    \includegraphics[width=.2\linewidth]{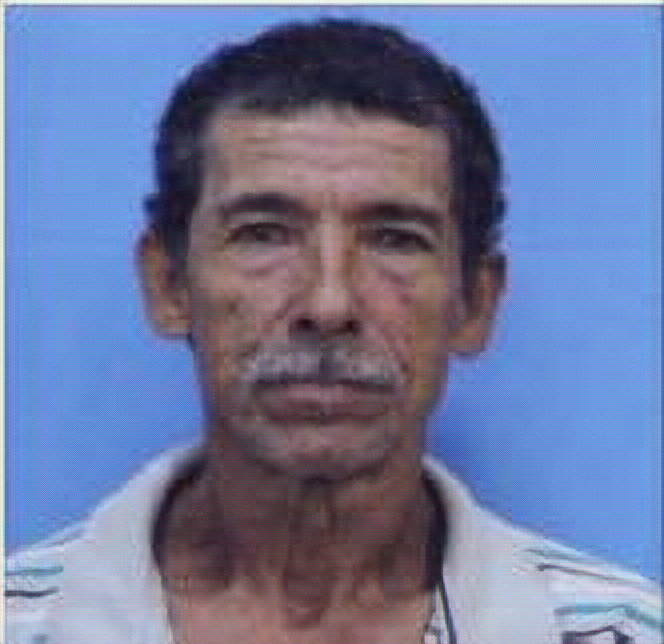}%
                    \label{fig:enhancements_c}}\quad
                \subfloat[TVD]{%
                    \includegraphics[width=.2\linewidth]{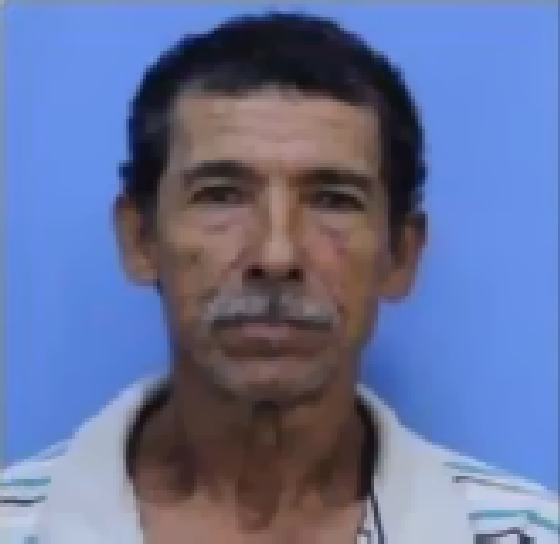}%
                    \label{fig:enhancements_d}}
                    
                \caption{Original picture and enhancements.}
                \label{fig:enhancements}
                \vskip -3ex
            \end{figure}

        \paragraph{Enhancement and description} 
            According to the measurement exposed in Section~\ref{semantic distance}, we collect the estimates of features accuracy predictions from different combination of VLMs and networks for the enhancement of the image in Table~\ref{tab:mytable}.
            From these data, we can highlight how the use of an image enhancement network generally leads to an improvement in the performances of the VLMs. 
            TinyLLaVa proves to be more efficient when jointly applied with enhancement models (with the exception of SRGAN), the Qwen-VL performs better on the original image alone. 
            The use of the MAXIM network leads to more considerable enhancements \figurename~\ref{fig:enhancements}.
            Total Variation Denoising can lead to a degrading effect of specific traits occurs, as introduced in Section~\ref{hyperparameters setting}. 
            
           Qwen-VL provides more accurate age estimates than TinyLLaVa, which often just rounds to multiples of 10, but Qwen-VL is more likely to carry uncertain results in smaller details like the iris color. Both VLMs struggle with identifying specific features like moles, tattoos, and scars, especially in shadowed areas, though image enhancements generally improve their performance except when applied after Total Variation Denoising.
    
            \begin{table}
                \centering
                \caption{VLMs prediction accuracy percentages.}
                \label{tab:mytable}
                \begin{tabular}{|c|c|c|}
                    \hline
                        \textbf{Input pictures}     &   \textbf{Qwen-VL}    & \textbf{TinyLLaVa}    \\
                    \hline
                        Original                    &   $84.381$              & $82.849$                \\
                    \hline
                        MAXIM                       &   $85.140$              & $85.500$                \\
                    \hline
                        SRGAN                       &   $84.537$              & $83.238$                \\
                    \hline
                        Total Variation Denoising   &   $81.568$              & $83.087$               \\
                    \hline
                \end{tabular}
                \vskip -2ex
            \end{table}
            
        \paragraph{Image augmentation} 

To generate new images, we tested the network with both original and enhanced pictures as inputs, according with the prompts derived by the methods described in Section~\ref{prompt_eng}. We observe that enhanced inputs imply similar results, with minimal changes in the subject's pose, regardless of whether the original or enhanced images were used. We additionally tested different combinations of input images (original, original + generated and one for each enhancement). 
    
        As confirmed by the Re-Identification metrics and shown in \figurename~\ref{fig:augmentation} and \figurename~\ref{fig:mugshot_a},\ref{fig:mugshot_b},\ref{fig:mugshot_c} in the Appendix, images generated only from the original mugshot were correctly recognized with higher levels of similarity. Images generated from a combination of real mugshot and synthetic ones introduced higher false positive rate and overall lower similarities and increased artificiality, confirming degradation of the samples associated with iterative generation.

            \begin{figure}
                \centering
                \subfloat[Original ]{%
                    \includegraphics[width=.2\linewidth]{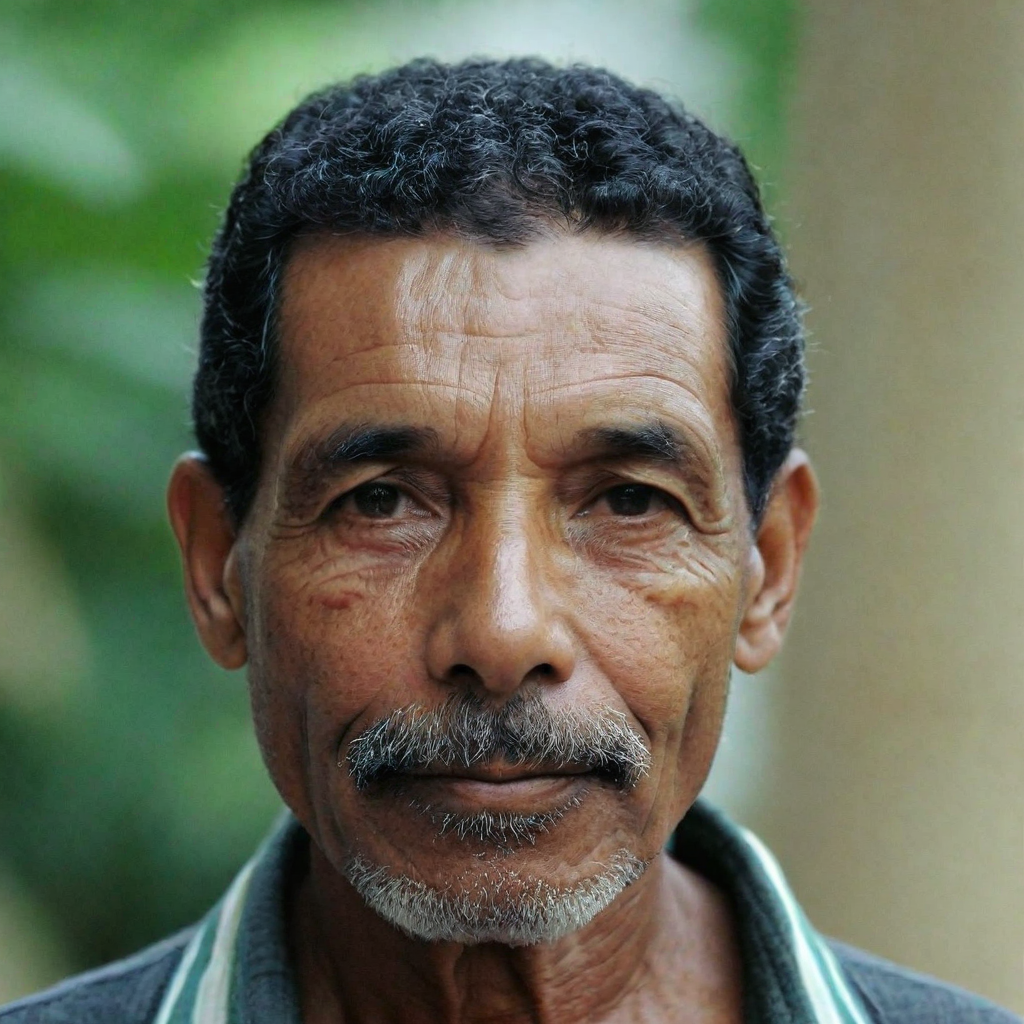}%
                    \label{fig:generation_a}} \quad
                \subfloat[MAXIM ]{%
                    \includegraphics[width=.2\linewidth]{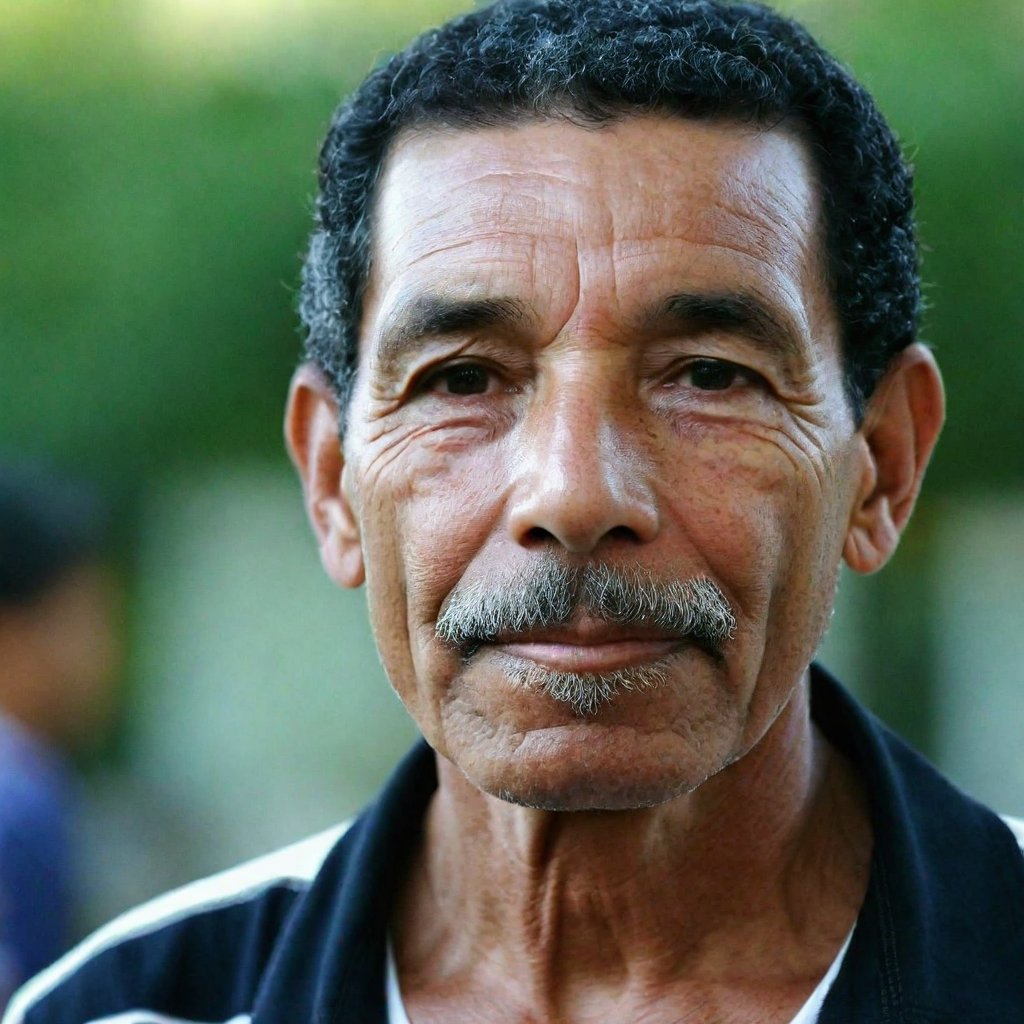}%
                    \label{fig:generation_b}} \quad
                \subfloat[SRGAN]{%
                    \includegraphics[width=.2\linewidth]{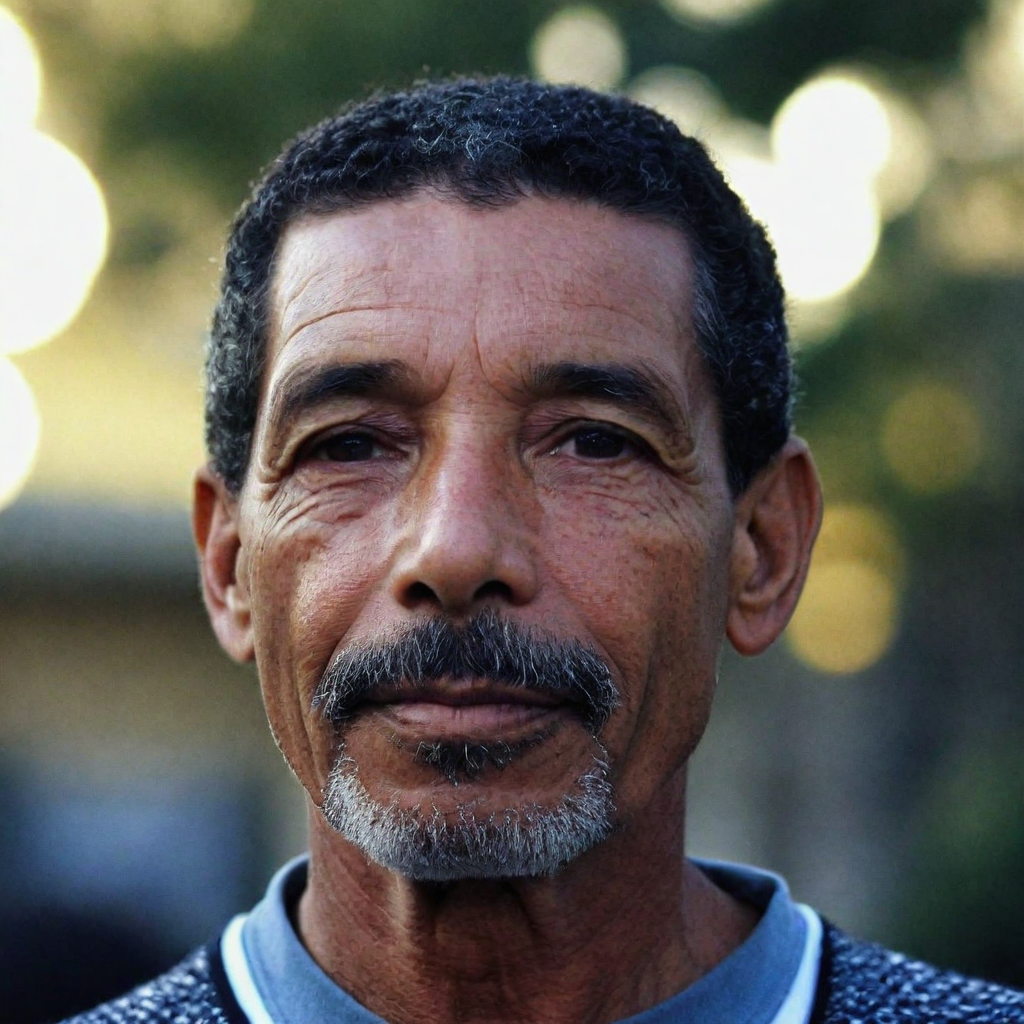}%
                    \label{fig:generation_c}}\quad
                \subfloat[TVD ]{%
                    \includegraphics[width=.2\linewidth]{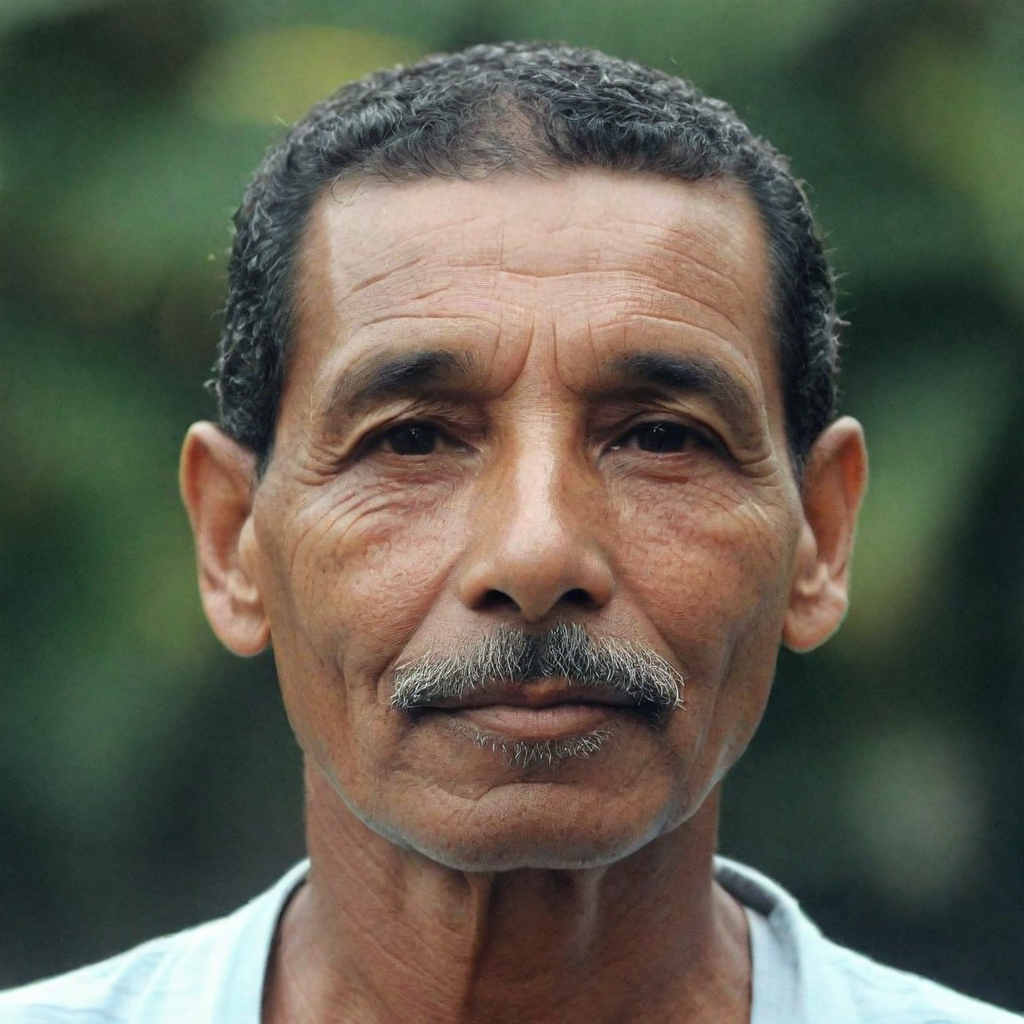}%
                    \label{fig:generation_d}}
                \caption{Images generated from the original picture and different enhancements.}
                \label{fig:generation}
            \end{figure}
            
            \begin{figure}
                \centering
                \subfloat[Original image]{%
                    \includegraphics[width=.25\linewidth]{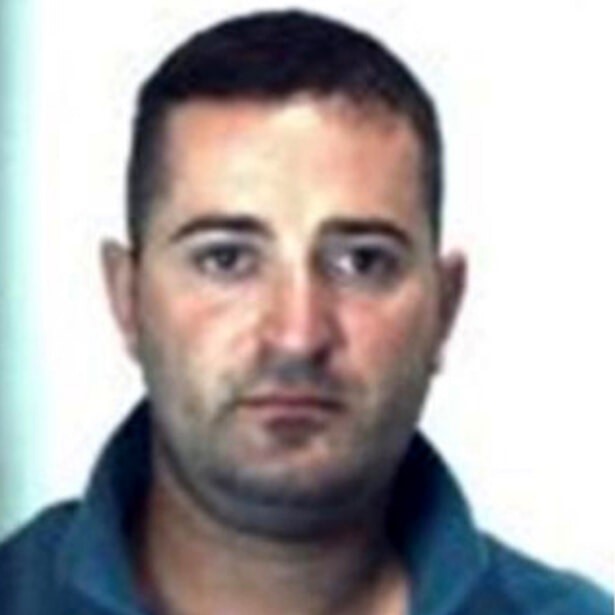}%
                    \label{fig:augmentation_a}} \quad
                \subfloat[Generated from original]{%
                    \includegraphics[width=.25\linewidth]{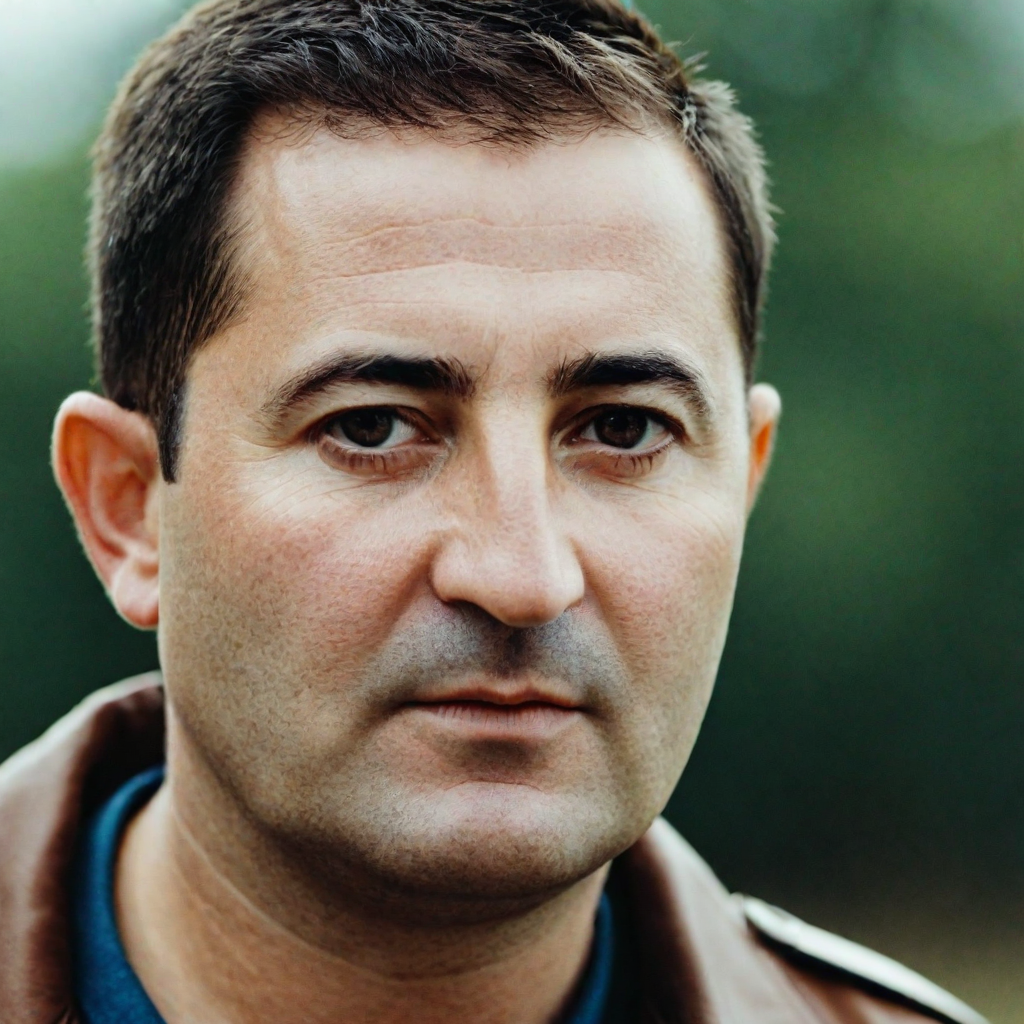}%
                    \label{fig:augmentation_b}} \quad
                \subfloat[Generated from generated]{%
                    \includegraphics[width=.25\linewidth]{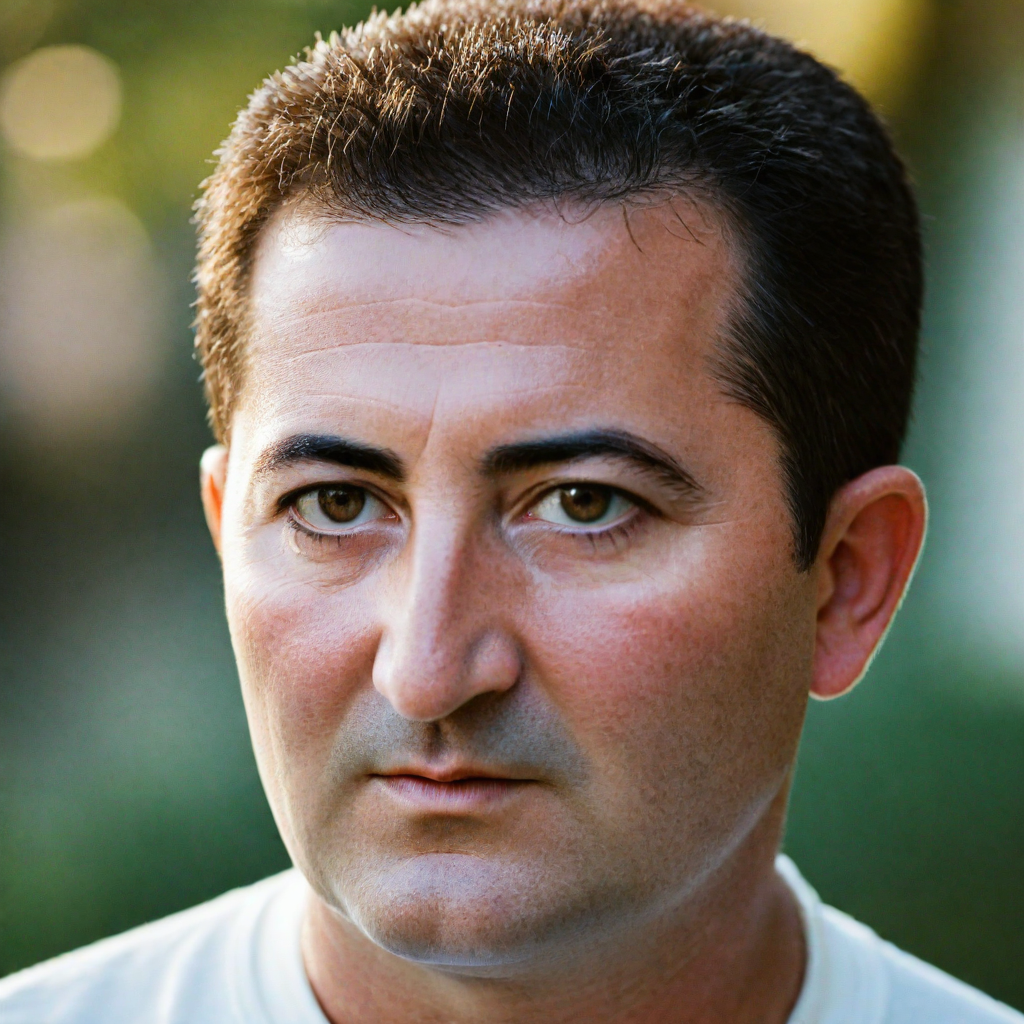}%
                    \label{fig:augmentation_c}}
                \caption{Example of different level of data augmentation showing the degradation of samples between ~\protect\subref{fig:augmentation_b} and ~\protect\subref{fig:augmentation_c}.}
                \label{fig:augmentation}
            \end{figure}

\paragraph{Aging}

            A natural development in image augmentation required by law enforcement is the passing of time. Several years can pass before being able to collect new information about missing of wanted individuals. This eventually can compromise the usability of some picture and posters. Throughout the years, face features undergo a process of development that may cause an overall solid alteration of the face. Such changes are likely to make the photographic evidences less meaningful.
            
            In our experiments we represent each subject with a pair of images depicting them at two different ages (the age gaps are variable among the dataset to improve generality). 

\begin{figure}
                \centering
                \vspace*{-10pt}
                \subfloat[Standard]{%
                    \includegraphics[width=.21\linewidth]{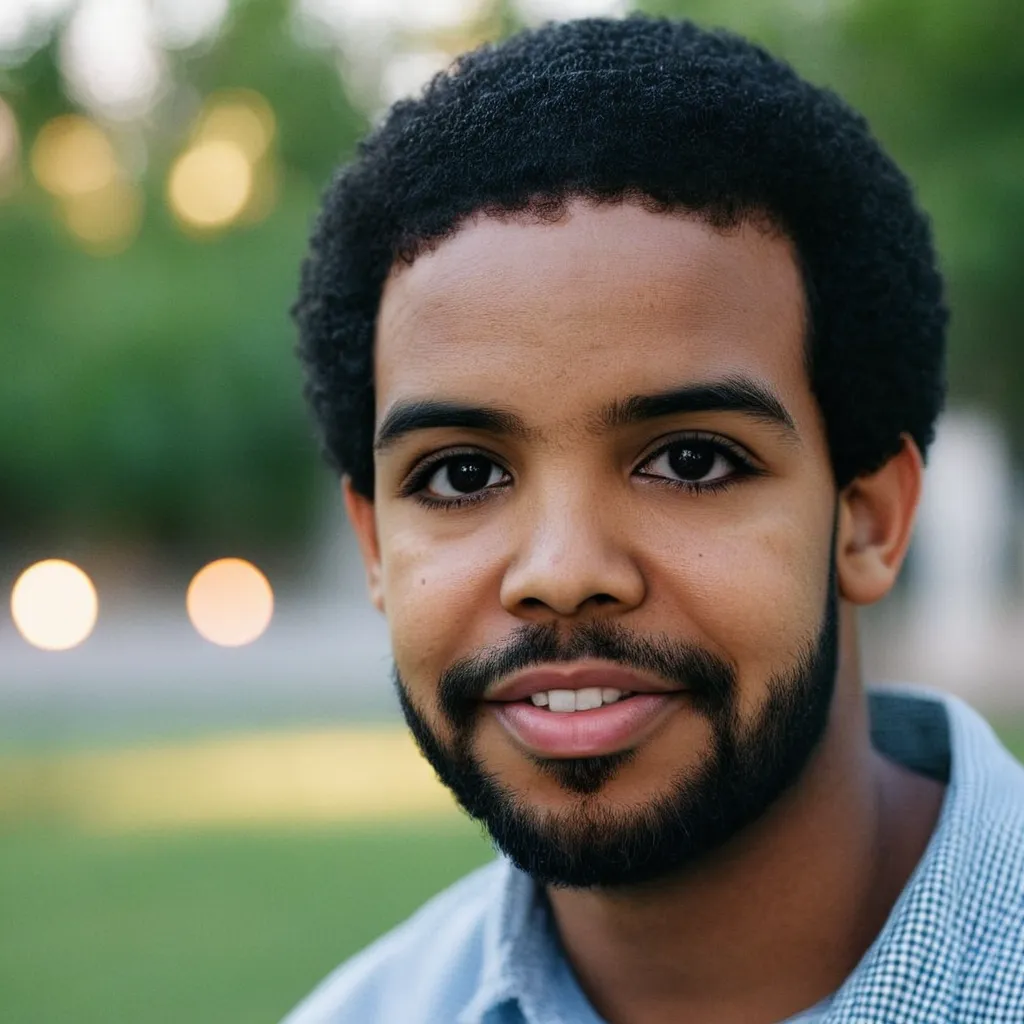}%
                    \label{fig:gen_a}} \quad
                \subfloat[$\neg$ \textit{Baby}]{%
                    \includegraphics[width=.21\linewidth]{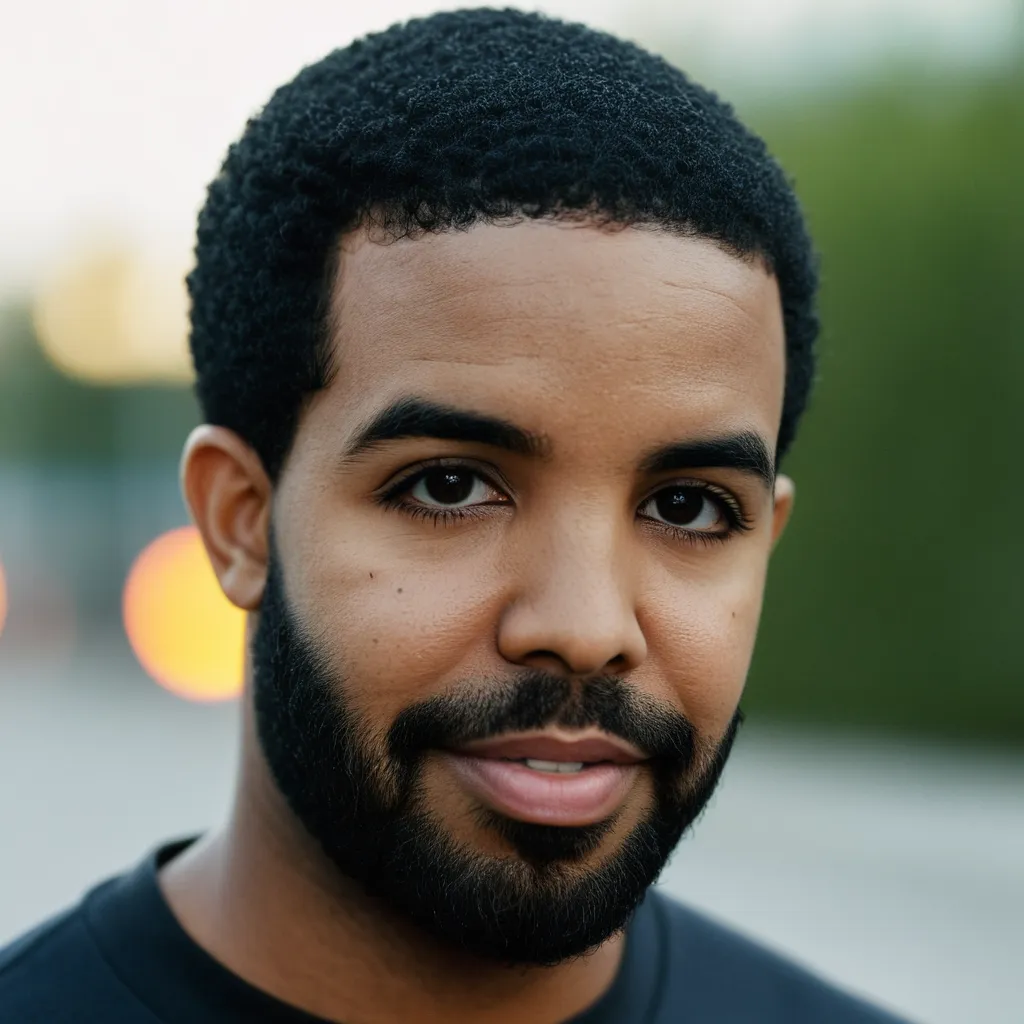}%
                    \label{fig:gen_b}} \quad
                \subfloat[Standard]{%
                    \includegraphics[width=.21\linewidth]{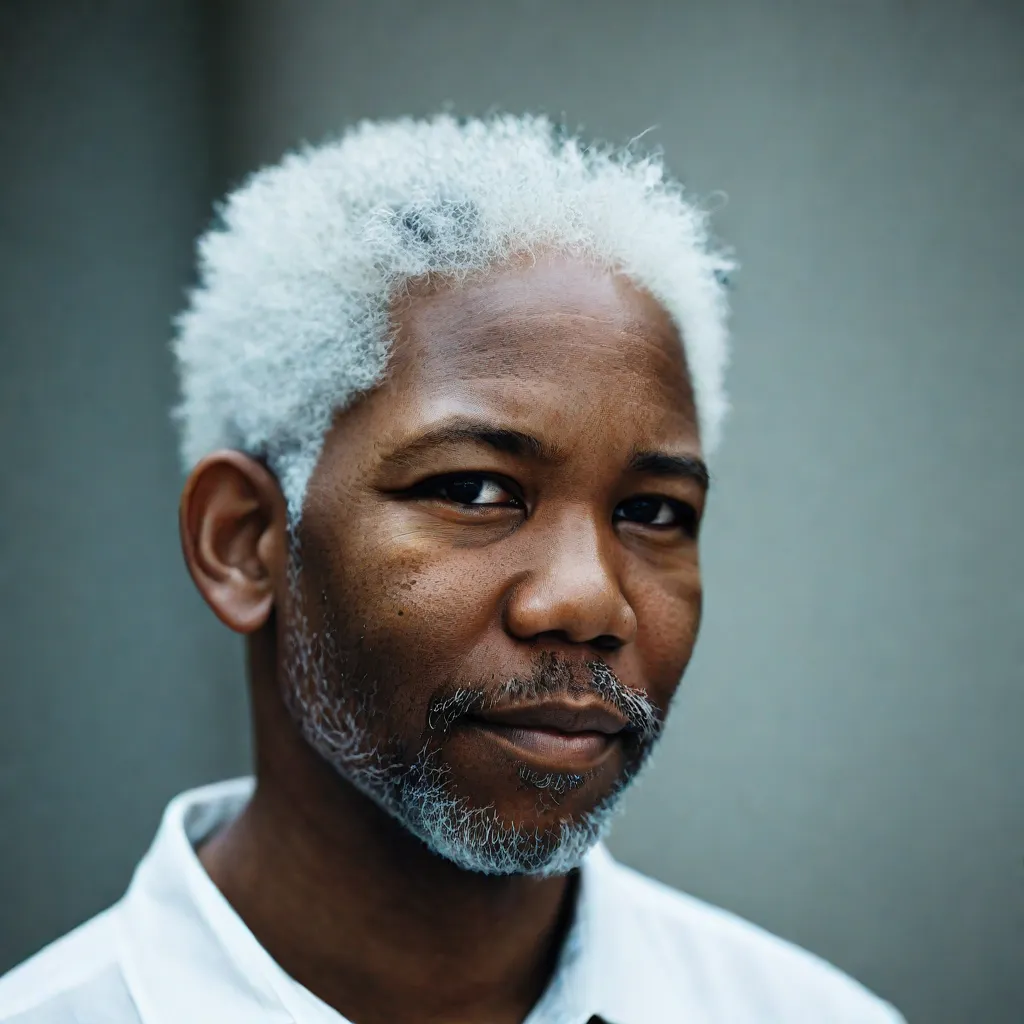}%
                    \label{fig:gen_c}} \quad
                \subfloat[\textit{Wrinkles}]{%
                    \includegraphics[width=.21\linewidth]{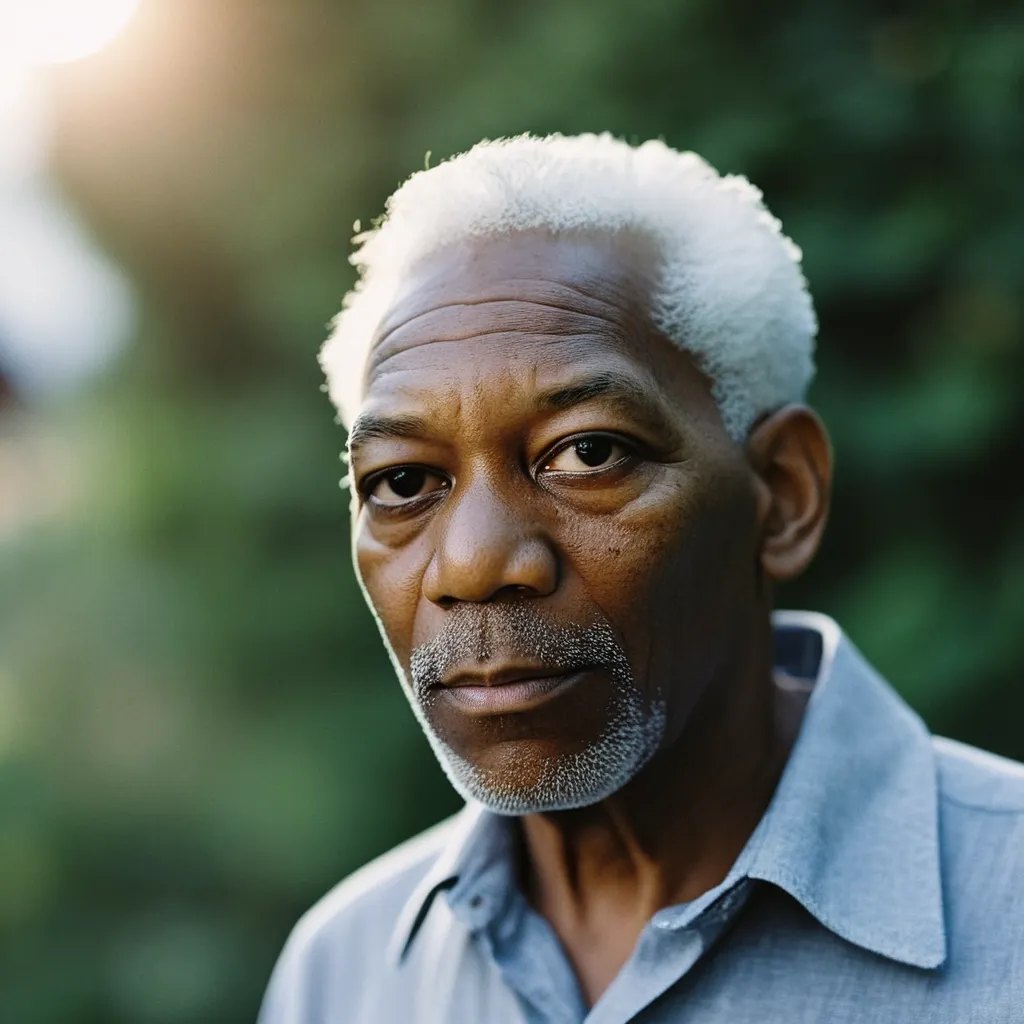}%
                    \label{fig:gen_d}}
                \caption{Prompt influence in the final results, showing difference between standard prompts in~\protect\subref{fig:gen_a} and~\protect\subref{fig:gen_c} and modified ~\protect\subref{fig:gen_b} and~\protect\subref{fig:gen_d}.}
                \label{fig:aging_prompts}
            \end{figure}

We observe better results with the inclusion of specific key prompt terms, especially tuning the negatives (the use of concepts that the model needs to avoid). 
References to “\textbf{wrinkles}” improved significantly the quality of elderly pictures. Same principle leveraged the negative prompt to reduce "children traits" with the inclusion of terms like “\textbf{child}”, “\textbf{baby}” as shown in \figurename~\ref{fig:aging_prompts}.
            
\begin{figure*}
                \centering
                \subfloat[Original image.]{%
                   \includegraphics[width=.12\linewidth]{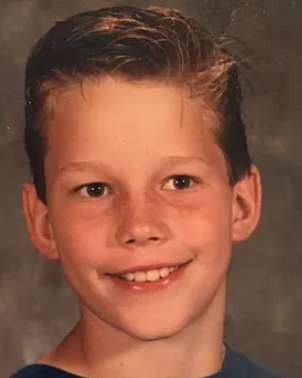}%
                    \label{fig:age_g}} \quad
                \subfloat[Target image.]{%
                    \includegraphics[width=.12\linewidth]{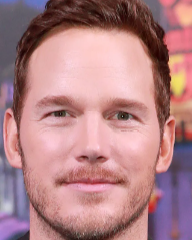}%
                    \label{fig:age_h}} \quad
                \subfloat[Aged image.]{%
                    \includegraphics[width=.12\linewidth]{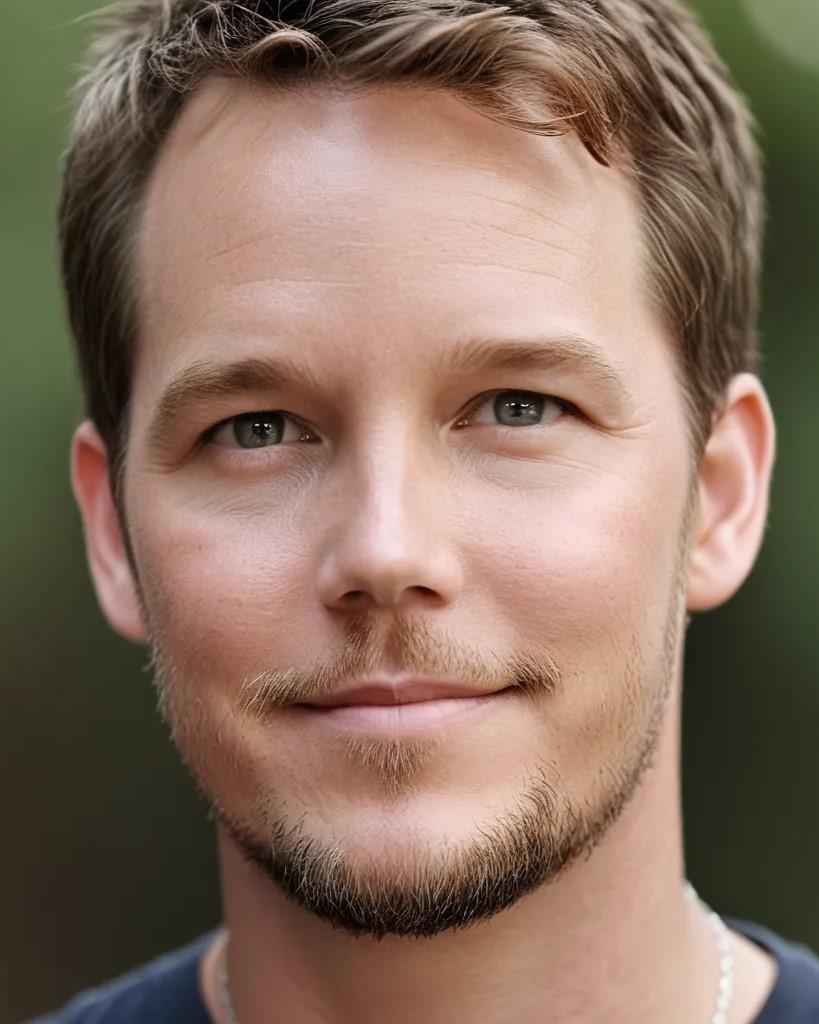}%
                    \label{fig:age_i}}\quad
                \subfloat[Original image.]{%
                    \includegraphics[width=.12\linewidth]{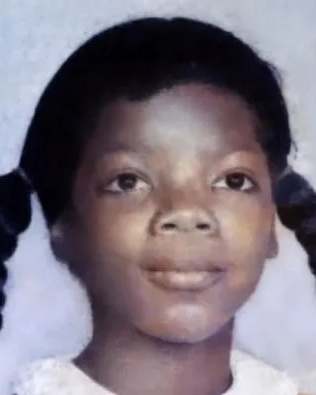}%
                    \label{fig:age_j}} \quad
                \subfloat[Target image.]{%
                    \includegraphics[width=.12\linewidth]{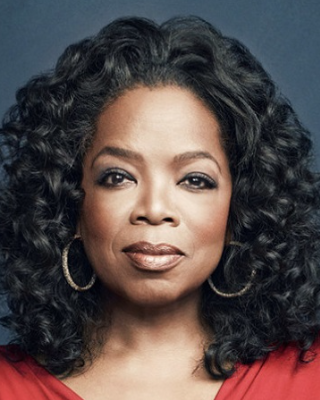}%
                    \label{fig:age_k}} \quad
                \subfloat[Aged image.]{%
                    \includegraphics[width=.12\linewidth]{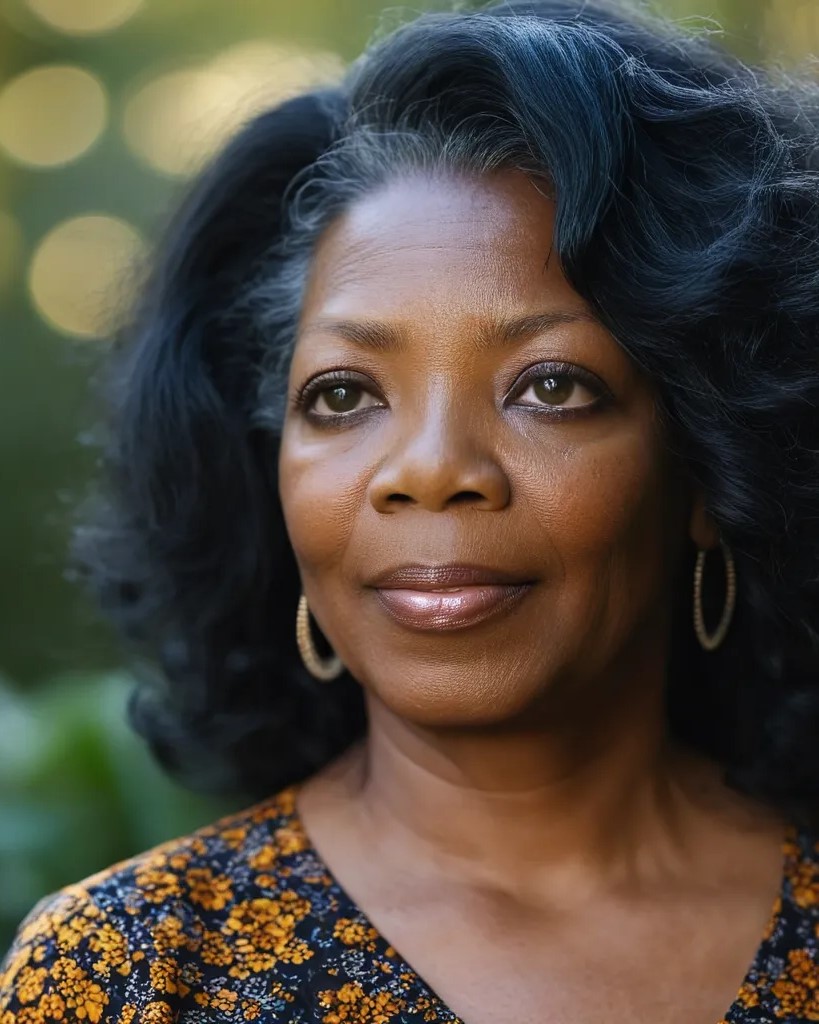}%
                    \label{fig:age_l}}
                \caption{Example of the aging task performed by our models: we introduce the starting images of the subject in a young age ~\protect\subref{fig:age_a},~\protect\subref{fig:age_d}, the target pictures of the subject aged~\protect\subref{fig:age_b},~\protect\subref{fig:age_e}, and the pictures generated ~\protect\subref{fig:age_c},~\protect\subref{fig:age_f}.}
                \label{fig:aging_examples}
                 \vskip -3ex
            \end{figure*}

            We can observe a really high and comparable recognition accuracy ($\sim$ 90\% confidence) of the aged subjects w.r.t. the performances recorded when testing the generated mugshots ($\sim$ 90\% confidence) (as we show in detail with the different distance metrics and confidences in the confusion matrices \figurename~\ref{fig:ageing_a} and \figurename~\ref{fig:ageing_b} in the Appendix). Nevertheless, the network still successfully recognize the two subjects as the same person in almost every case. Examples of successful cases are reported in \figurename~\ref{fig:aging_examples}. 
            We notice two main causes of insuccess: expected output as pre-adolescent children with most face features under transformation or really old people as input, whose main face features and peculiarities started to undergo unpredictable substantial changes.
            
\section{Conclusion}
\label{sec:conclusion}
    We successfully design and implement a mugshot generator that enhances real low-quality evidence pictures to perform automatic linguistic description and achieve several layers of augmentation. Additionally, we investigated the aging processes to extend the capabilities of forensic technologies and investigations. Finally, we proposed, developed and compared different methods to validate the trustworthiness of such systems in possible law enforcement applications.
    Future works should focus on improving model robustness in several challenging scenarios. We noticed that images of people belonging to ethnic groups 
    are rendered with lower accuracy, often generating images of people with a lighter skin tone than the target one, and sometimes altering their facial features.
    
    A possible improvement suggest labeling based on people's skin tone rather than their ethnic group, as previously done in different works~\cite{fair_classification}.
    However, this approach may represent an obstacle for the evaluation of the linguistic description, since intelligence organizations usually provide the subject's ethnicity, rather than their complexion. 
    
    Additionally, investigating de-aging techniques to assist in aligning witness testimony in investigations that trace back several years (with suspects that naturally aged in the meanwhile).
    
    Our preliminary results show a significant deterioration of performance w.r.t. the other mugshot tasks, highlighted by the increase of the false negative and false positive rate in the process of re-identification. In such instances (especially if the target individual is 15 or less) the models are prone to copy facial traits of the developed adults, and paste them in the younger version (as shown in Appendix B). This lack of generalization implies many uncanny and overall inaccurate results that leave huge room for improvement.
    
    Additional research may focus on different types of real world inputs, like surveillance images obtained from CCTV cameras. 
    This scenario represents instances where the suspect of a crime is present in the camera footage, but their identity is unknown. Future extensions of this work may also propose the development and administration of surveys and interviews with field experts to improve quality assessment of synthetic data.

\bibliography{IEEEabrv,egbib}

\begin{thebibliography}{10}
\providecommand{\url}[1]{#1}
\csname url@samestyle\endcsname
\providecommand{\newblock}{\relax}
\providecommand{\bibinfo}[2]{#2}
\providecommand{\BIBentrySTDinterwordspacing}{\spaceskip=0pt\relax}
\providecommand{\BIBentryALTinterwordstretchfactor}{4}
\providecommand{\BIBentryALTinterwordspacing}{\spaceskip=\fontdimen2\font plus
\BIBentryALTinterwordstretchfactor\fontdimen3\font minus \fontdimen4\font\relax}
\providecommand{\BIBforeignlanguage}[2]{{%
\expandafter\ifx\csname l@#1\endcsname\relax
\typeout{** WARNING: IEEEtran.bst: No hyphenation pattern has been}%
\typeout{** loaded for the language `#1'. Using the pattern for}%
\typeout{** the default language instead.}%
\else
\language=\csname l@#1\endcsname
\fi
#2}}
\providecommand{\BIBdecl}{\relax}
\BIBdecl

\bibitem{5601735}
B.~Klare, Z.~Li, and A.~K. Jain, ``Matching forensic sketches to mug shot photos,'' \emph{IEEE Transactions on Pattern Analysis and Machine Intelligence}, vol.~33, no.~3, pp. 639--646, 2011.

\bibitem{app14209285}
S.~Giuliani, F.~Tosti, P.~Lopes, C.~Ciampini, and C.~Nardinocchi, ``Design of a multi-vision system for a three-dimensional mug shot model to improve forensic facial identification,'' \emph{Applied Sciences}, vol.~14, no.~20, 2024.

\bibitem{10744467}
K.~K. Jain, S.~Grosz, A.~M. Namboodiri, and A.~K. Jain, ``Clip4sketch: Enhancing sketch to mugshot matching through dataset augmentation using diffusion models,'' in \emph{2024 IEEE International Joint Conference on Biometrics (IJCB)}, 2024, pp. 1--10.

\bibitem{forensic_art}
K.~T. Taylor, \emph{Forensic art and Illustration}.\hskip 1em plus 0.5em minus 0.4em\relax CRC Press, 2001.

\bibitem{sketchgen1}
S.~Mahajan, V.~Humbe, A.~Raorane, and A.~Deshmukh, ``Forensic face sketch artist system,'' \emph{International Journal for Research in Applied Science and Engineering Technology}, vol.~10, pp. 1660--1681, 08 2022.

\bibitem{sketchgen2}
F.~Reynaud and A.~Fortunato, ``Forensic sketch {AIrtist},'' \url{https://lablab.ai/event/openai-whisper-gpt3-codex-dalle2-hackathon/eagleai/forensic-sketch-airtist}, Dec 2022, accessed: April 2024.

\bibitem{sketchtoimg1}
S.~Nikkath~Bushra and K.~Uma~Maheswari, ``Crime investigation using dcgan by forensic sketch-to-face transformation (stf)- a review,'' in \emph{2021 5th International Conference on Computing Methodologies and Communication (ICCMC)}, 2021, pp. 1343--1348.

\bibitem{sketchtoimg2}
D.~J. Salim and B.-S. Lin, ``Everyone is a forensic artist: Sketch-to-photo transformation for human face,'' in \emph{2021 IEEE 4th International Conference on Knowledge Innovation and Invention (ICKII)}, 2021, pp. 118--122.

\bibitem{sketchtoimg3}
\BIBentryALTinterwordspacing
S.~Devakumar and G.~Sarath, ``Forensic sketch to real image using dcgan,'' \emph{Procedia Comput. Sci.}, vol. 218, no.~C, p. 1612–1620, jan 2023. [Online]. Available: \url{https://doi.org/10.1016/j.procs.2023.01.139}
\BIBentrySTDinterwordspacing

\bibitem{facerecog1}
\BIBentryALTinterwordspacing
W.~Zhao, R.~Chellappa, P.~J. Phillips, and A.~Rosenfeld, ``Face recognition: A literature survey,'' \emph{ACM Comput. Surv.}, vol.~35, no.~4, p. 399–458, dec 2003. [Online]. Available: \url{https://doi.org/10.1145/954339.954342}
\BIBentrySTDinterwordspacing

\bibitem{facerecog2}
\BIBentryALTinterwordspacing
M.~Wang and W.~Deng, ``Deep face recognition: A survey,'' \emph{Neurocomputing}, vol. 429, pp. 215--244, 2021. [Online]. Available: \url{https://www.sciencedirect.com/science/article/pii/S0925231220316945}
\BIBentrySTDinterwordspacing

\bibitem{biasface}
H.~F. Menezes, A.~S.~C. Ferreira, E.~T. Pereira, and H.~M. Gomes, ``Bias and fairness in face detection,'' in \emph{2021 34th SIBGRAPI Conference on Graphics, Patterns and Images (SIBGRAPI)}, 2021, pp. 247--254.

\bibitem{diffusionmodels}
F.-A. Croitoru, V.~Hondru, R.~T. Ionescu, and M.~Shah, ``Diffusion models in vision: A survey,'' \emph{IEEE Transactions on Pattern Analysis and Machine Intelligence}, vol.~45, no.~9, pp. 10\,850--10\,869, 2023.

\bibitem{stablediffusion}
R.~Rombach, A.~Blattmann, D.~Lorenz, P.~Esser, and B.~Ommer, ``High-resolution image synthesis with latent diffusion models,'' 2022.

\bibitem{antipov2017face}
G.~Antipov, M.~Baccouche, and J.-L. Dugelay, ``Face aging with conditional generative adversarial networks,'' in \emph{2017 IEEE international conference on image processing (ICIP)}.\hskip 1em plus 0.5em minus 0.4em\relax IEEE, 2017, pp. 2089--2093.

\bibitem{panis2016overview}
G.~Panis, A.~Lanitis, N.~Tsapatsoulis, and T.~F. Cootes, ``Overview of research on facial ageing using the fg-net ageing database,'' \emph{Iet Biometrics}, vol.~5, no.~2, pp. 37--46, 2016.

\bibitem{maxim}
Z.~Tu, H.~Talebi, H.~Zhang, F.~Yang, P.~Milanfar, A.~Bovik, and Y.~Li, ``{MAXIM}: Multi-axis {MLP} for image processing,'' 2022.

\bibitem{tvdenoise}
L.~I. Rudin, S.~Osher, and E.~Fatemi, ``Nonlinear total variation based noise removal algorithms,'' \emph{Physica D: Nonlinear Phenomena}, vol.~60, no.~1, pp. 259--268, 1992.

\bibitem{srgan}
C.~Ledig, L.~Theis, F.~Huszar, J.~Caballero, A.~Cunningham, A.~Acosta, A.~Aitken, A.~Tejani, J.~Totz, Z.~Wang, and W.~Shi, ``Photo-realistic single image super-resolution using a generative adversarial network,'' 2017.

\bibitem{qwenvl}
J.~Bai, S.~Bai, S.~Yang, S.~Wang, S.~Tan, P.~Wang, J.~Lin, C.~Zhou, and J.~Zhou, ``{Qwen-VL}: A versatile vision-language model for understanding, localization, text reading, and beyond,'' 2023.

\bibitem{tinyllava}
B.~Zhou, Y.~Hu, X.~Weng, J.~Jia, J.~Luo, X.~Liu, J.~Wu, and L.~Huang, ``{TinyLLaVA}: A framework of small-scale large multimodal models,'' 2024.

\bibitem{photomaker}
Z.~Li, M.~Cao, X.~Wang, Z.~Qi, M.-M. Cheng, and Y.~Shan, ``Photomaker: Customizing realistic human photos via stacked {ID} embedding,'' 2023.

\bibitem{sdxl}
D.~Podell, Z.~English, K.~Lacey, A.~Blattmann, T.~Dockhorn, J.~Müller, J.~Penna, and R.~Rombach, ``{SDXL}: Improving latent diffusion models for high-resolution image synthesis,'' 2023.

\bibitem{serengil2020lightface}
S.~I. Serengil and A.~Ozpinar, ``Lightface: A hybrid deep face recognition framework,'' in \emph{2020 Innovations in Intelligent Systems and Applications Conference (ASYU)}.\hskip 1em plus 0.5em minus 0.4em\relax IEEE, 2020, pp. 1--5.

\bibitem{serengil2024lightface}
\BIBentryALTinterwordspacing
S.~Serengil and A.~Ozpinar, ``A benchmark of facial recognition pipelines and co-usability performances of modules,'' \emph{Journal of Information Technologies}, vol.~17, no.~2, pp. 95--107, 2024. [Online]. Available: \url{https://dergipark.org.tr/en/pub/gazibtd/issue/84331/1399077}
\BIBentrySTDinterwordspacing

\bibitem{schroff2015facenet}
F.~Schroff, D.~Kalenichenko, and J.~Philbin, ``Facenet: A unified embedding for face recognition and clustering,'' in \emph{2015 IEEE Conference on Computer Vision and Pattern Recognition (CVPR)}, 2015, pp. 815--823.

\bibitem{globerson2004euclidean}
A.~Globerson, G.~Chechik, F.~Pereira, and N.~Tishby, ``Euclidean embedding of co-occurrence data,'' \emph{Advances in neural information processing systems}, vol.~17, 2004.

\bibitem{weinberger2009distance}
K.~Q. Weinberger and L.~K. Saul, ``Distance metric learning for large margin nearest neighbor classification,'' \emph{J. Mach. Learn. Res.}, vol.~10, p. 207–244, jun 2009.

\bibitem{recognito}
\BIBentryALTinterwordspacing
Recognito: Introduction to nist frvt top 1 face recognition technology. [Online]. Available: \url{https://recognito.vision/index.php/2023/12/21/recognito-introduction-to-nist-frvt-top-1-face-recognition-technology/}
\BIBentrySTDinterwordspacing

\bibitem{FRTE}
\BIBentryALTinterwordspacing
Face recognition technology evaluation (frte) 1:1 verification. [Online]. Available: \url{https://pages.nist.gov/frvt/html/frvt11.html}
\BIBentrySTDinterwordspacing

\bibitem{fair_classification}
J.~Buolamwini and T.~Gebru, ``Gender shades: Intersectional accuracy disparities in commercial gender classification,'' in \emph{Proceedings of the 1st Conference on Fairness, Accountability and Transparency}, ser. Proceedings of Machine Learning Research, S.~A. Friedler and C.~Wilson, Eds., vol.~81.\hskip 1em plus 0.5em minus 0.4em\relax PMLR, 23--24 Feb 2018, pp. 77--91.

\end{thebibliography}

\newpage

\onecolumn

\appendices

\section{}
    In this section we introduce the \textbf{confusion matrices} computed during the comparison methods to validate and prove the similarity of our synthetic poster / augmented pictures with respect to the original photos of the suspects / celebrities.

    We show a reduced version of our experimental results. The larger confusion matrices will be published and made available to researchers in the dedicated webpage.
    
    \

    \subsection{Mugshots}

        \begin{figure}[ht] 
            \centering
            \onecolumn\includegraphics[width=1\linewidth]{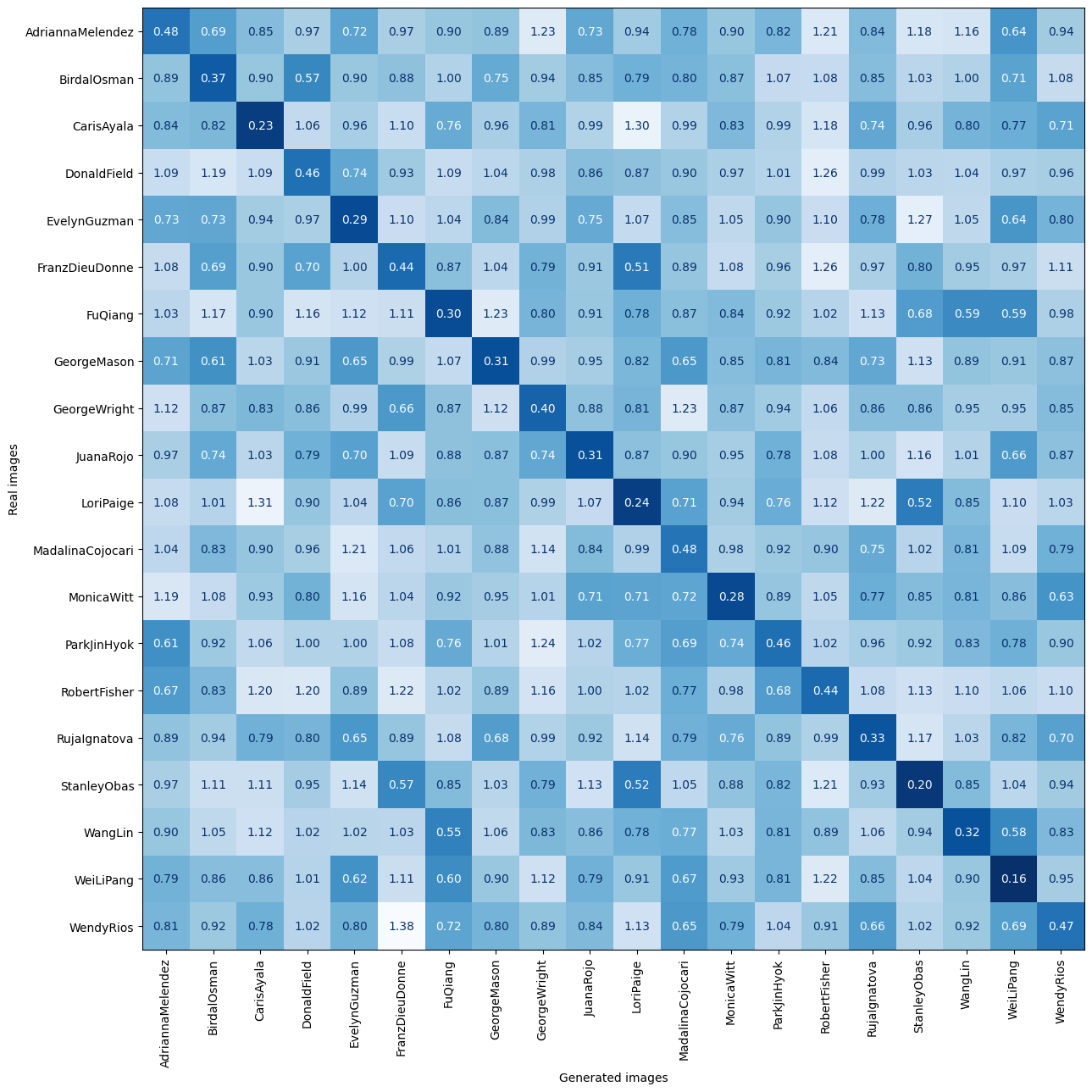}%
            \caption{Confusion matrix obtained from the results of tests performed over images generated starting from only the original mugshot. Tests were conducted using DeepFace as re-identification network. Values reported in the matrix stand for the distances between the two represented subjects, measured by DeepFace. }
            \label{fig:mugshot_a}
        \end{figure}
        \newpage
        \begin{figure}[ht] 
            \centering
            \onecolumn\includegraphics[width=1\linewidth]{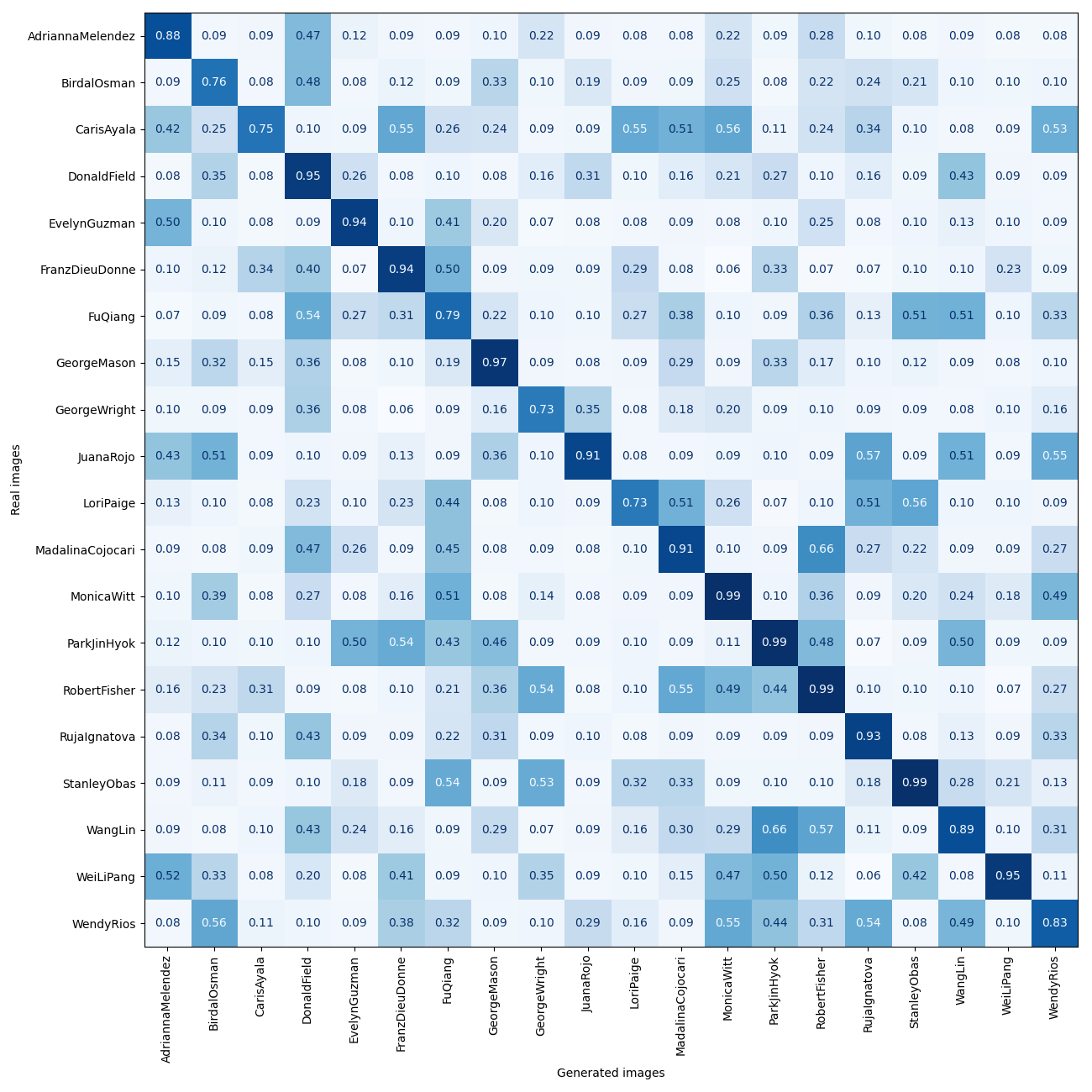}%
            \caption{Confusion matrix obtained from the results of tests performed over images generated starting from only the original mugshot. Tests were conducted using Recognito as re-identification network. Values reported in the matrix stand for the similarity values between the two represented subjects, measured by Recognito.}
            \label{fig:mugshot_b}
        \end{figure}
        \newpage
        \begin{figure}[ht] 
            \centering
            \onecolumn\includegraphics[width=1\linewidth]{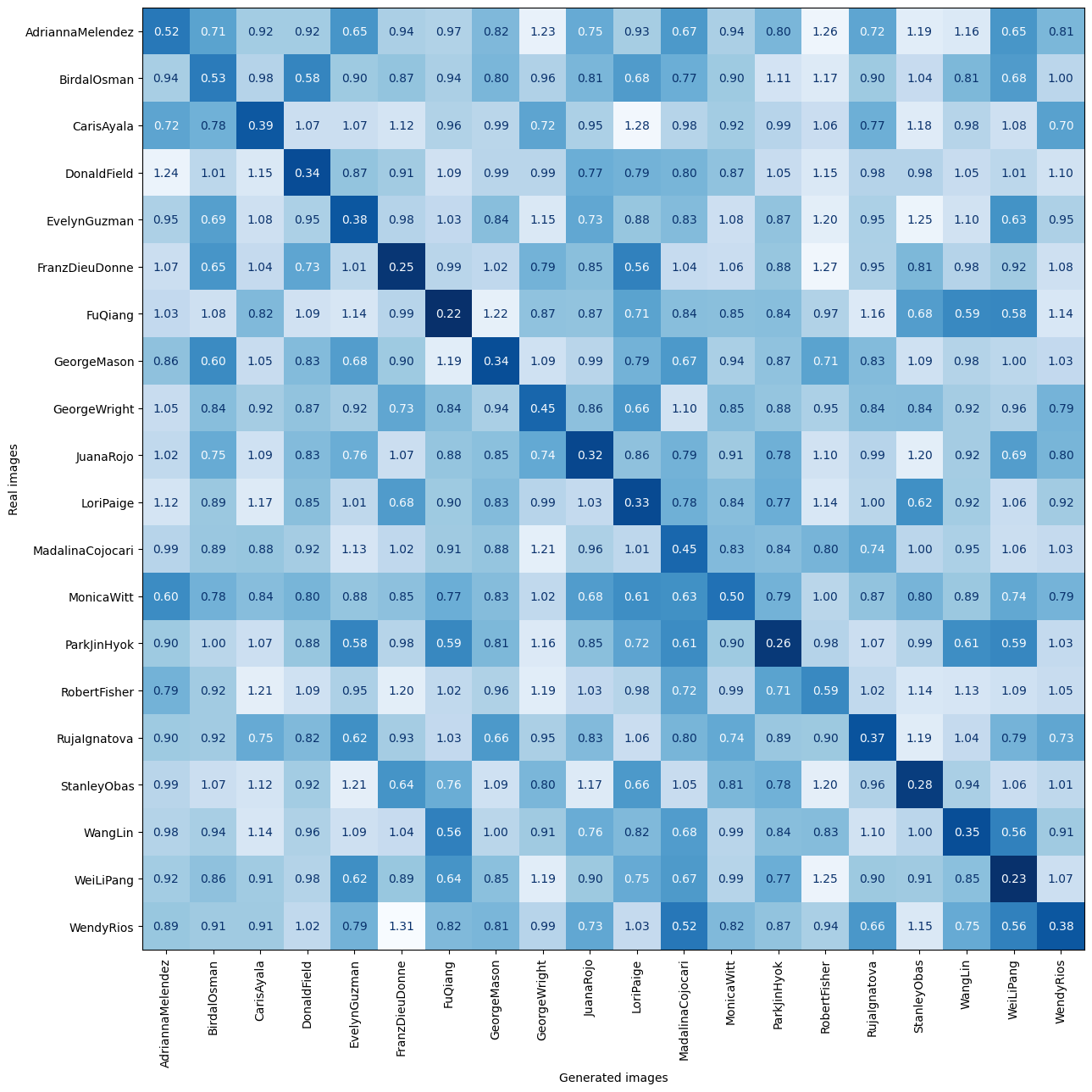}%
            \caption{Confusion matrix obtained from the results of tests performed over images generated starting from a combination of the original mugshot and others previously generated. Tests were conducted using Deepface as re-identification network. Values reported in the matrix stand for the distances between the two represented subjects, measured by DeepFace.}
            \label{fig:mugshot_c}
        \end{figure}
        
    \newpage

    \subsection{Aging}

        \

        \begin{figure}[ht] 
            \centering
            \onecolumn\includegraphics[width=1\linewidth]{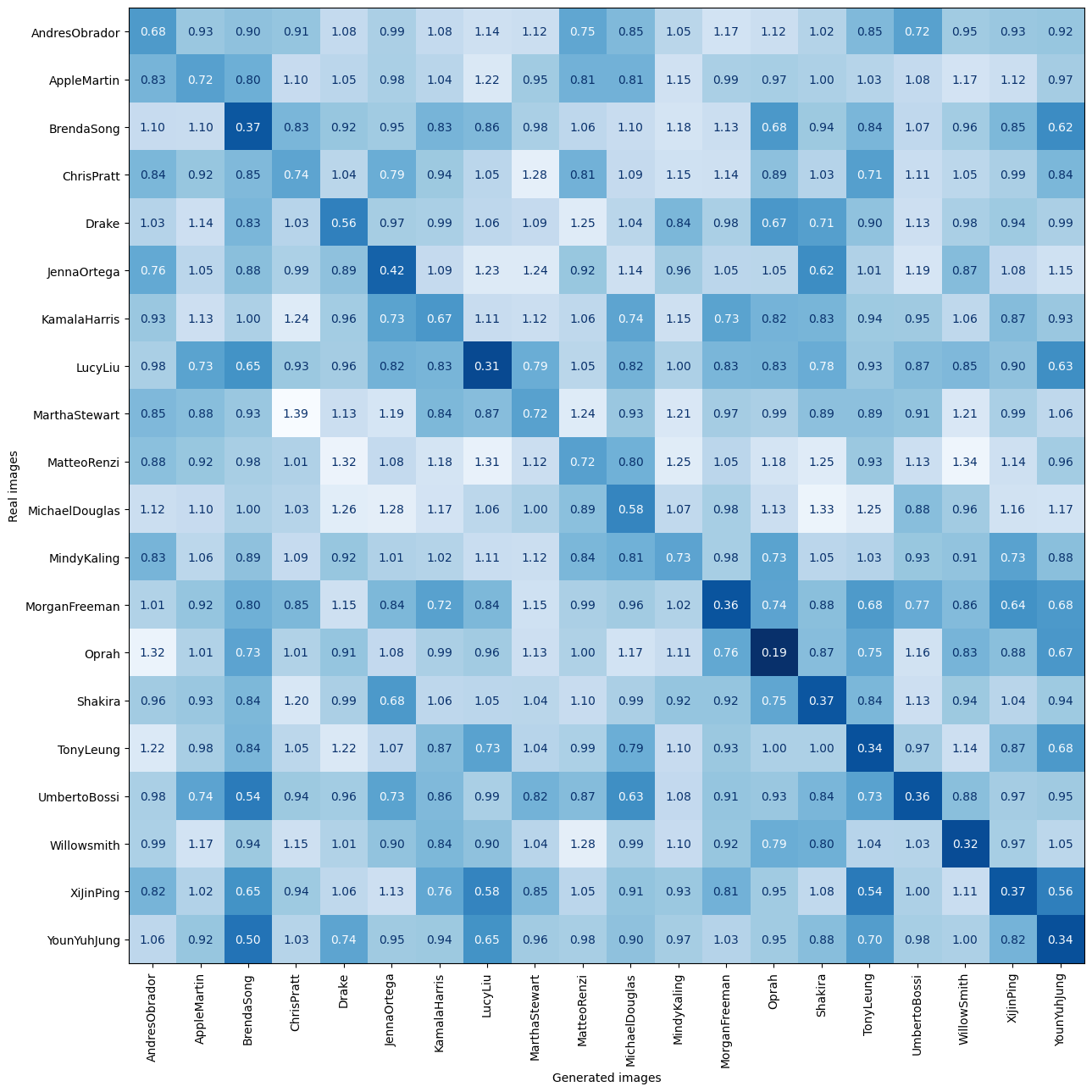}%
            \caption{Confusion matrix obtained from the results of tests performed over aged images. Tests were conducted using DeepFace as re-identification network. Values reported in the matrix stand for the distances between the two represented subjects, measured by DeepFace.}
            \label{fig:ageing_a}
        \end{figure}
        \newpage
        \begin{figure}[ht] 
            \centering
            \onecolumn\includegraphics[width=1\linewidth]{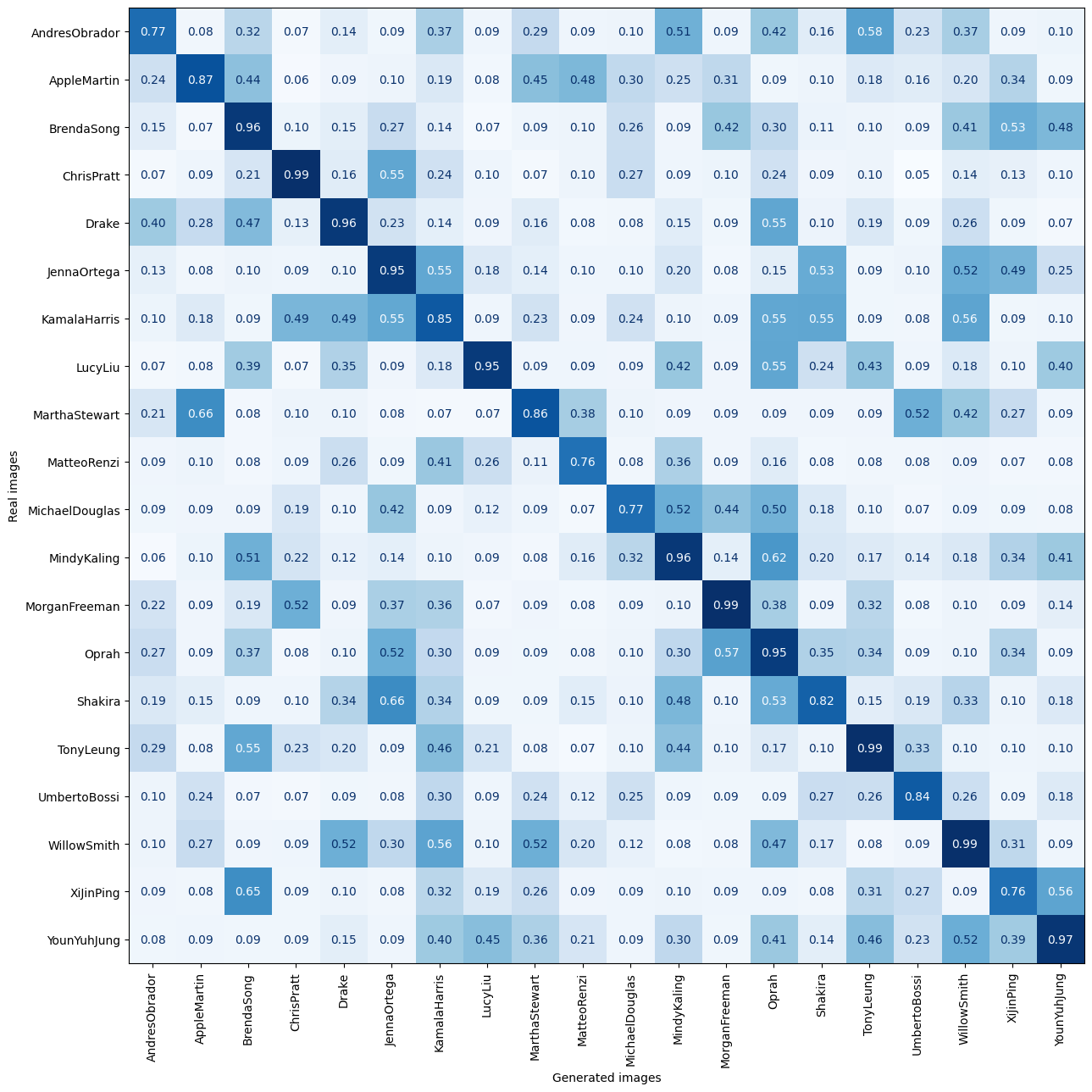}%
            \caption{Confusion matrix obtained from the results of tests performed over aged images. Tests were conducted using Recognito as re-identification network. Values reported in the matrix stand for the similarity values between the two represented subjects, measured by Recognito.}
            \label{fig:ageing_b}
        \end{figure}
        \newpage
        \subsection{De-aging}

        \begin{figure}[ht] 
            \centering
            \onecolumn\includegraphics[width=1\linewidth]{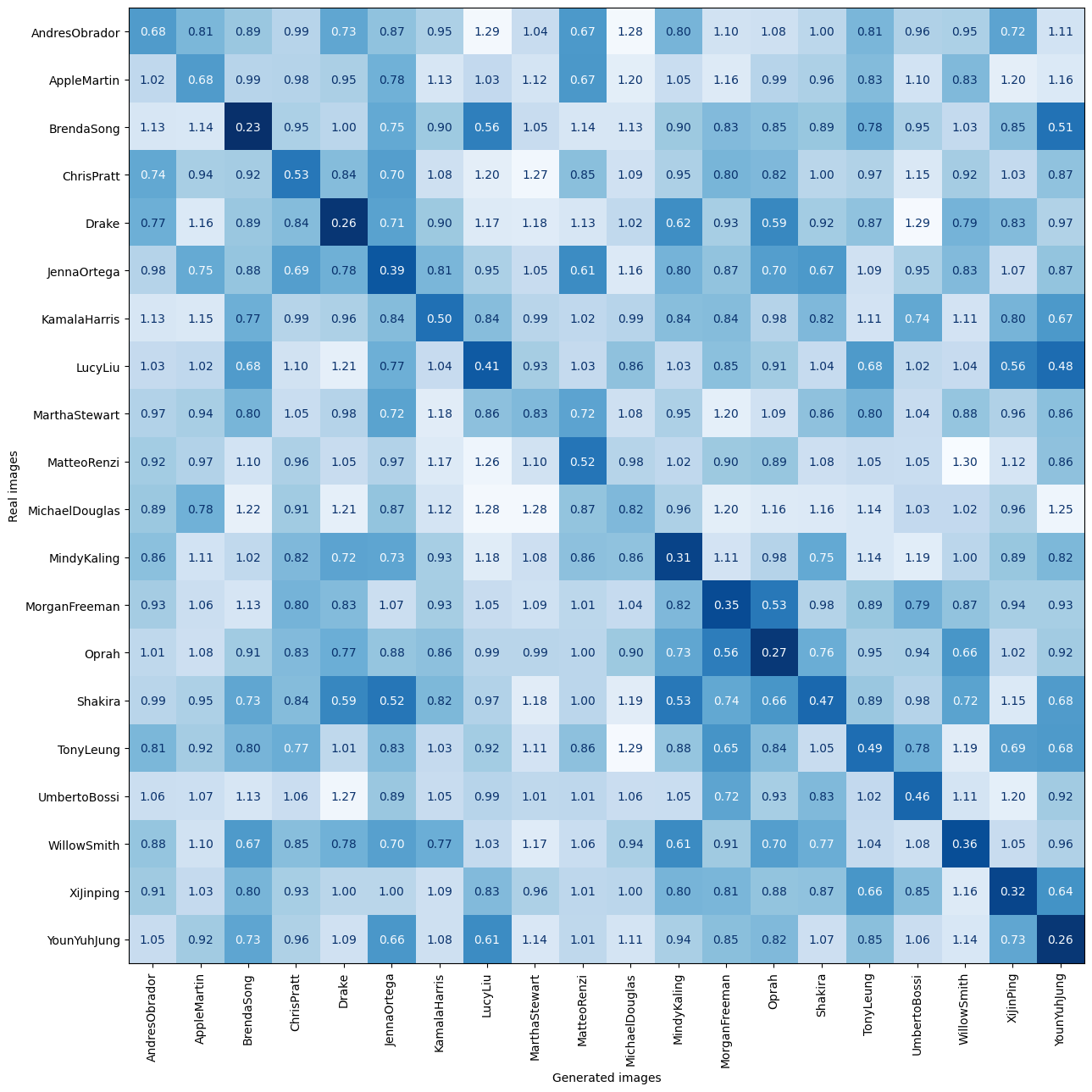}%
            \caption{Confusion matrix obtained from the results of tests performed over de-aged images. Tests were conducted using DeepFace as re-identification network. Values reported in the matrix stand for the distances between the two represented subjects, measured by DeepFace.}
            \label{fig:deageing_a}
        \end{figure}
        \newpage
        \begin{figure}[ht] 
            \centering
            \onecolumn\includegraphics[width=1\linewidth]{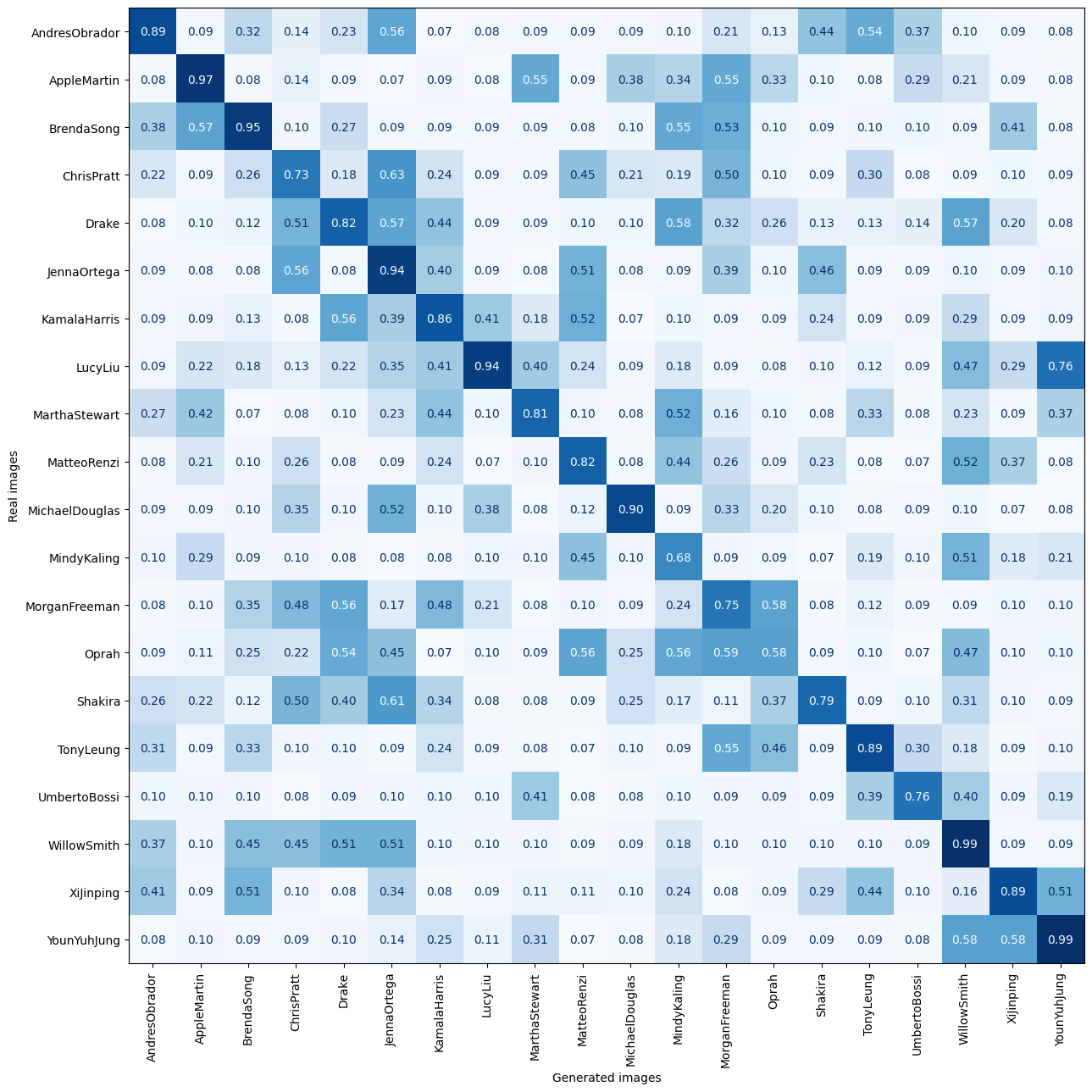}%
            \caption{Confusion matrix obtained from the results of tests performed over de-aged images. Tests were conducted using Recognito as re-identification network. Values reported in the matrix stand for the similarity values between the two represented subjects, measured by Recognito.}
            \label{fig:deageing_b}
        \end{figure}

        \newpage

\section{}
    In this section, we show several additional \textbf{synthetic mugshots} generated using our proposed framework. These examples showcase the effectiveness of our method across a diverse range of subjects, including variations in age, ethnicity, and input image quality. The images illustrate the results of both our image enhancement techniques and the subsequent synthetic augmentation process. Even more examples will be made available in a dedicated repository upon acceptance.

    \begin{figure}[ht]
        \centering
        \subfloat[Original image.]{%
            \includegraphics[width=.1275\linewidth]{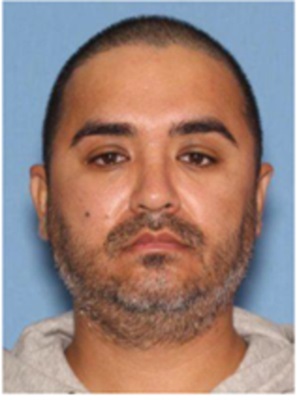}%
            \label{fig:mug_a}} \quad
        \subfloat[Augmentation from original.]{%
            \includegraphics[width=.17\linewidth]{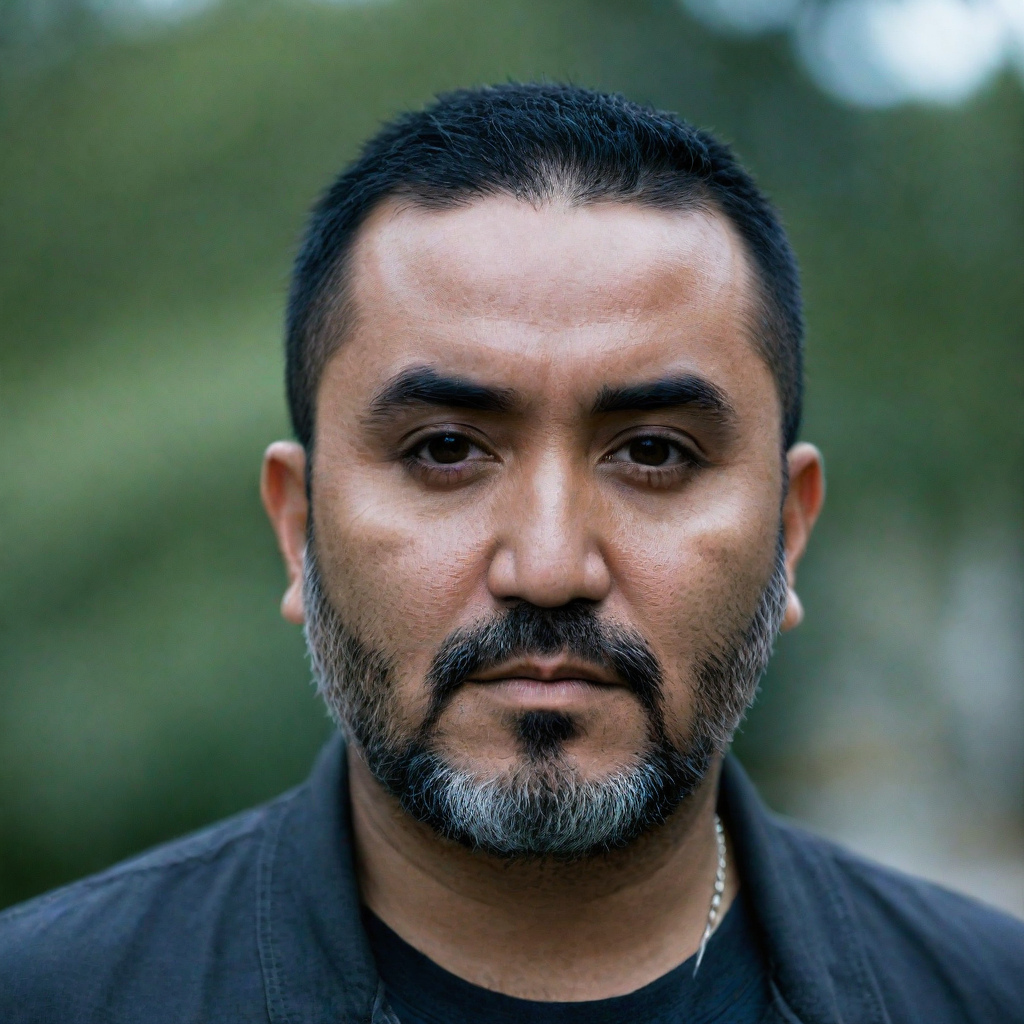}%
            \label{fig:mug_b}} \quad
        \subfloat[Augmentation from original + enhanced.]{%
            \includegraphics[width=.17\linewidth]{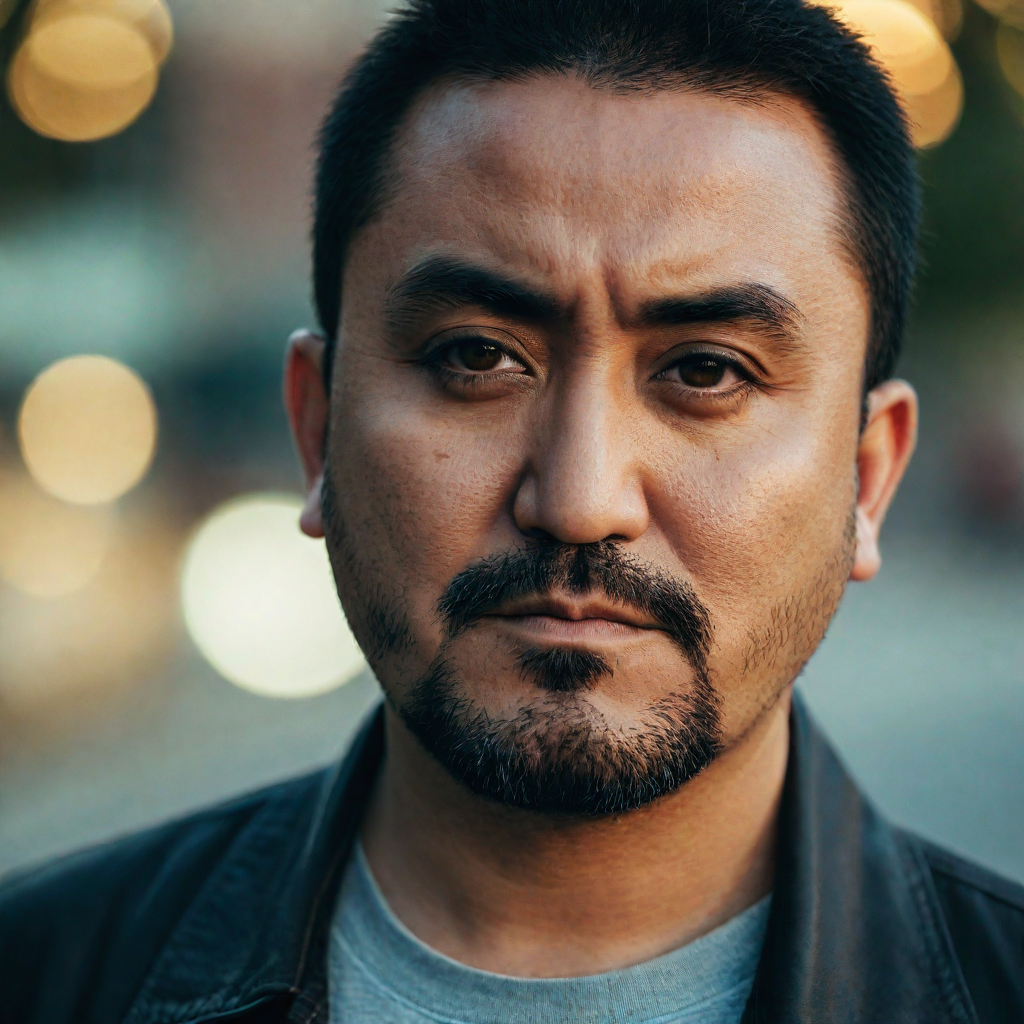}%
            \label{fig:mug_c}}\quad
            
        \subfloat[Original image]{%
            \includegraphics[width=.13\linewidth]{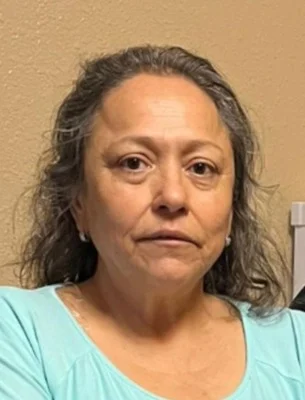}%
            \label{fig:mug_d}} \quad
        \subfloat[Augmentation from original.]{%
            \includegraphics[width=.17\linewidth]{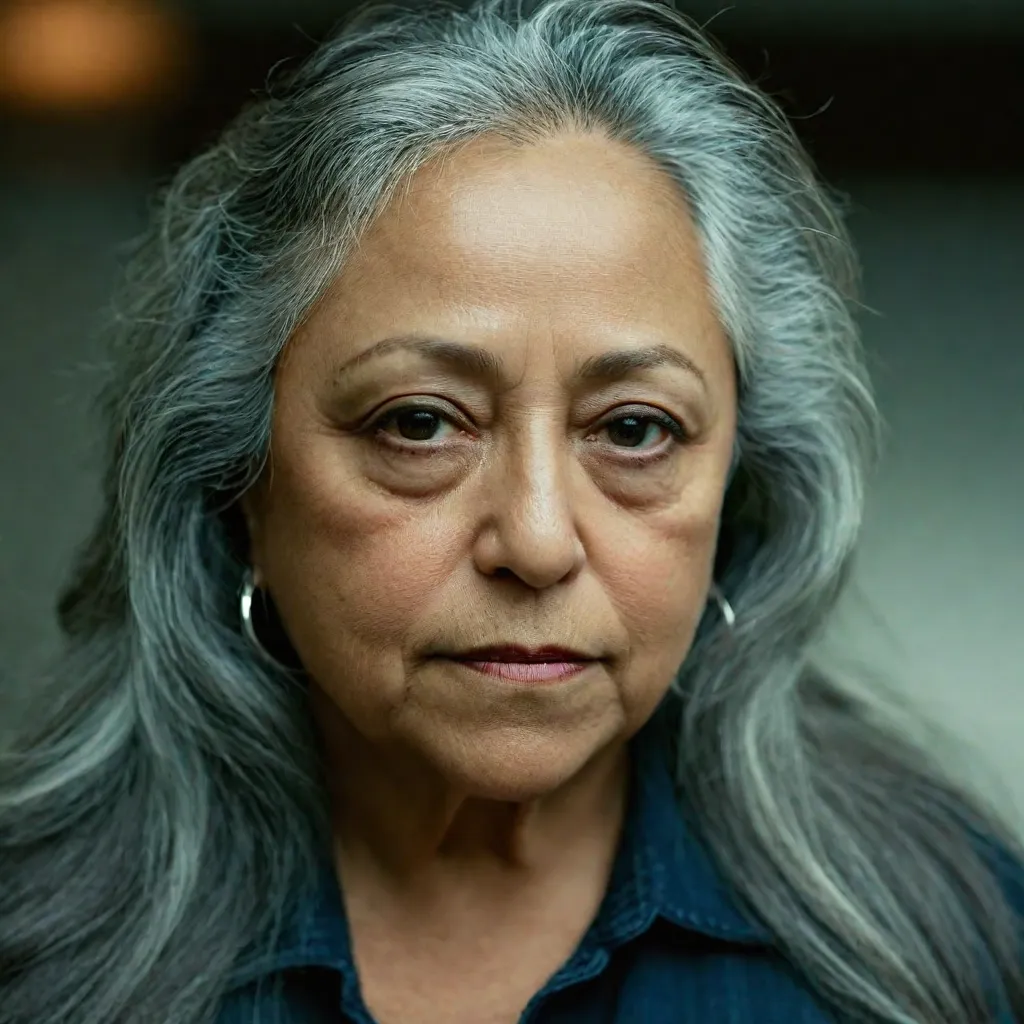}%
            \label{fig:mug_e}} \quad
        \subfloat[Augmentation from original + enhanced.]{%
            \includegraphics[width=.17\linewidth]{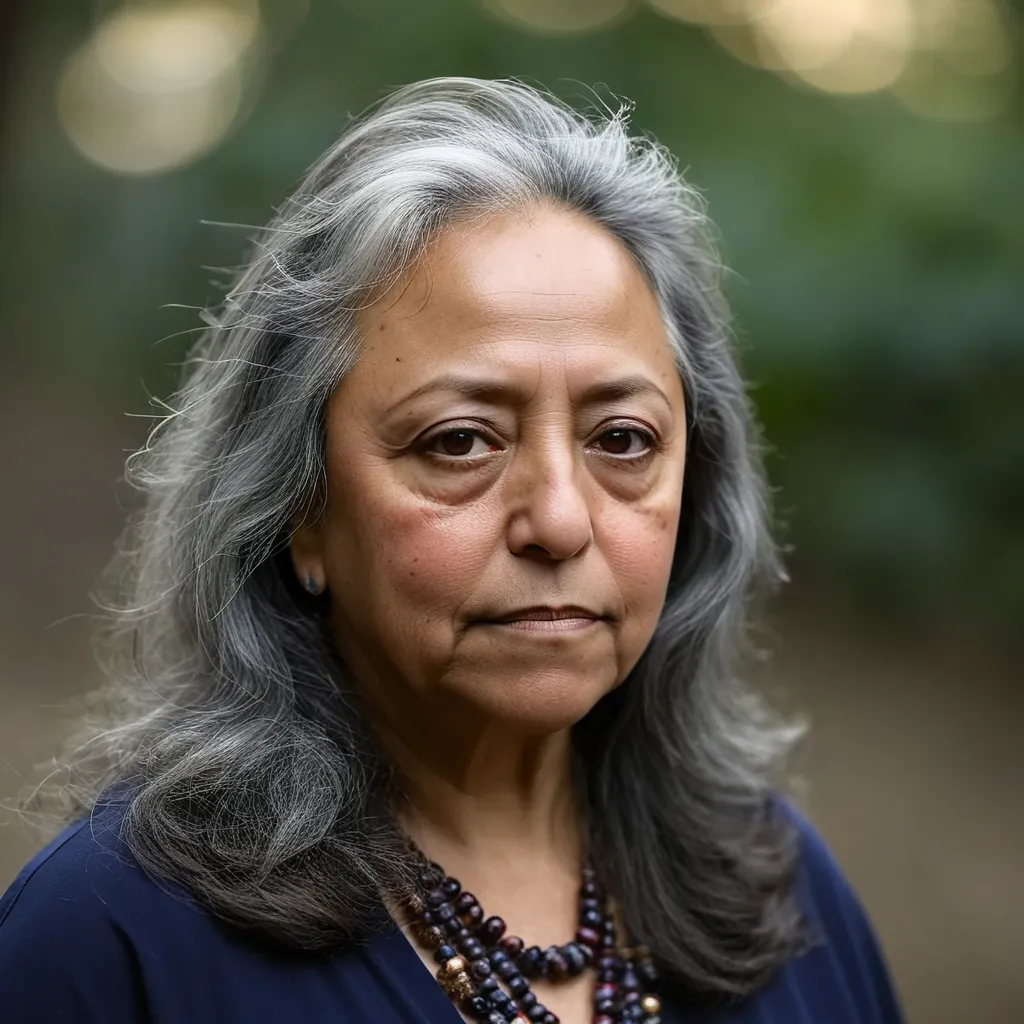}%
            \label{fig:mug_f}}\quad
            
        \subfloat[Original image.]{%
            \includegraphics[width=.11\linewidth]{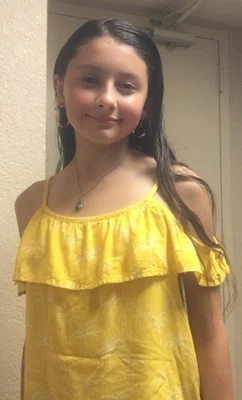}%
            \label{fig:mug_g}} \quad
        \subfloat[Augmentation from original.]{%
            \includegraphics[width=.17\linewidth]{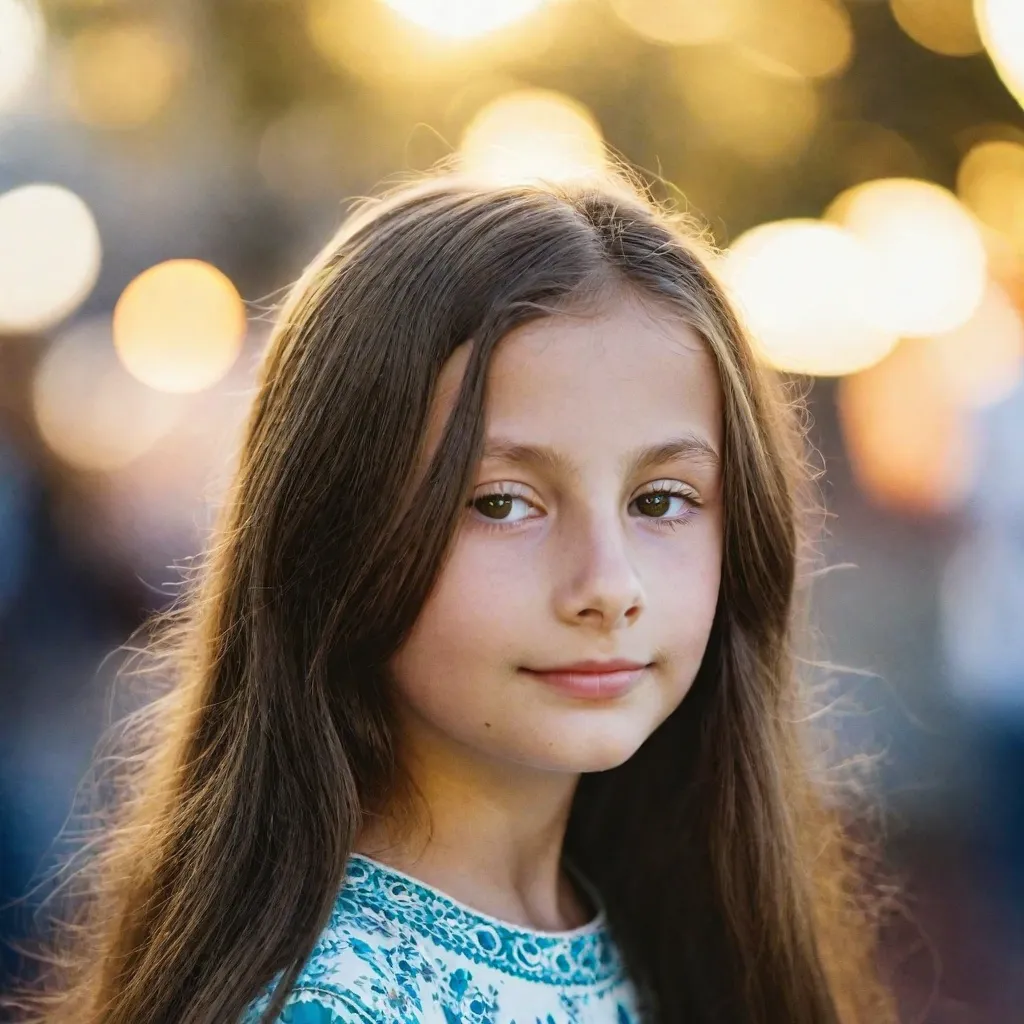}%
            \label{fig:mug_h}} \quad
        \subfloat[Augmentation from original + enhanced.]{%
            \includegraphics[width=.17\linewidth]{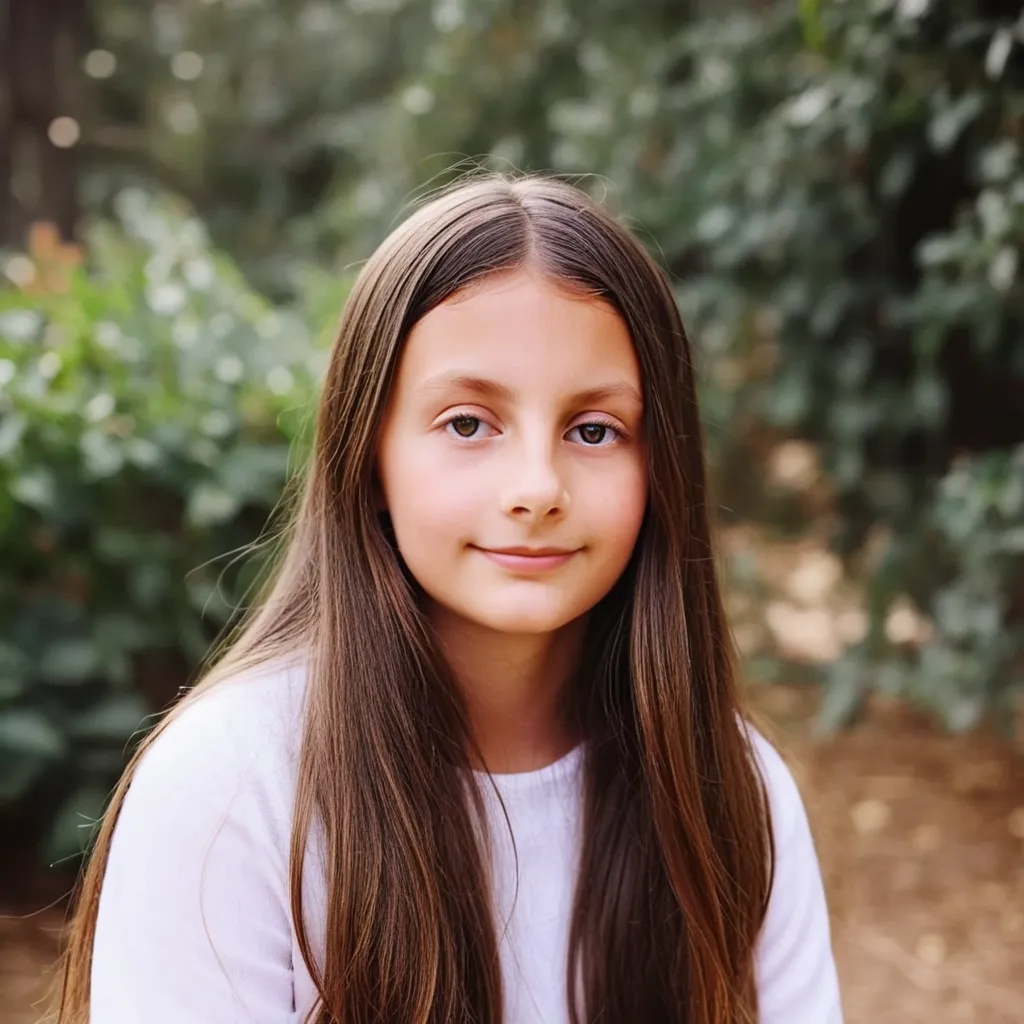}%
            \label{fig:mug_i}}\quad
            
        \subfloat[Original image.]{%
            \includegraphics[width=.1375\linewidth]{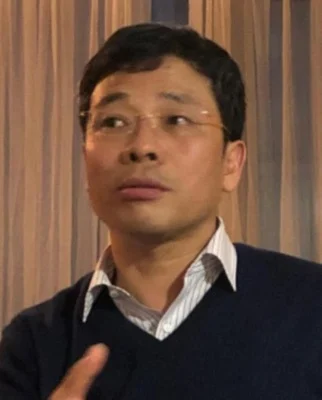}%
            \label{fig:mug_l}} \quad
        \subfloat[Augmentation from original.]{%
            \includegraphics[width=.17\linewidth]{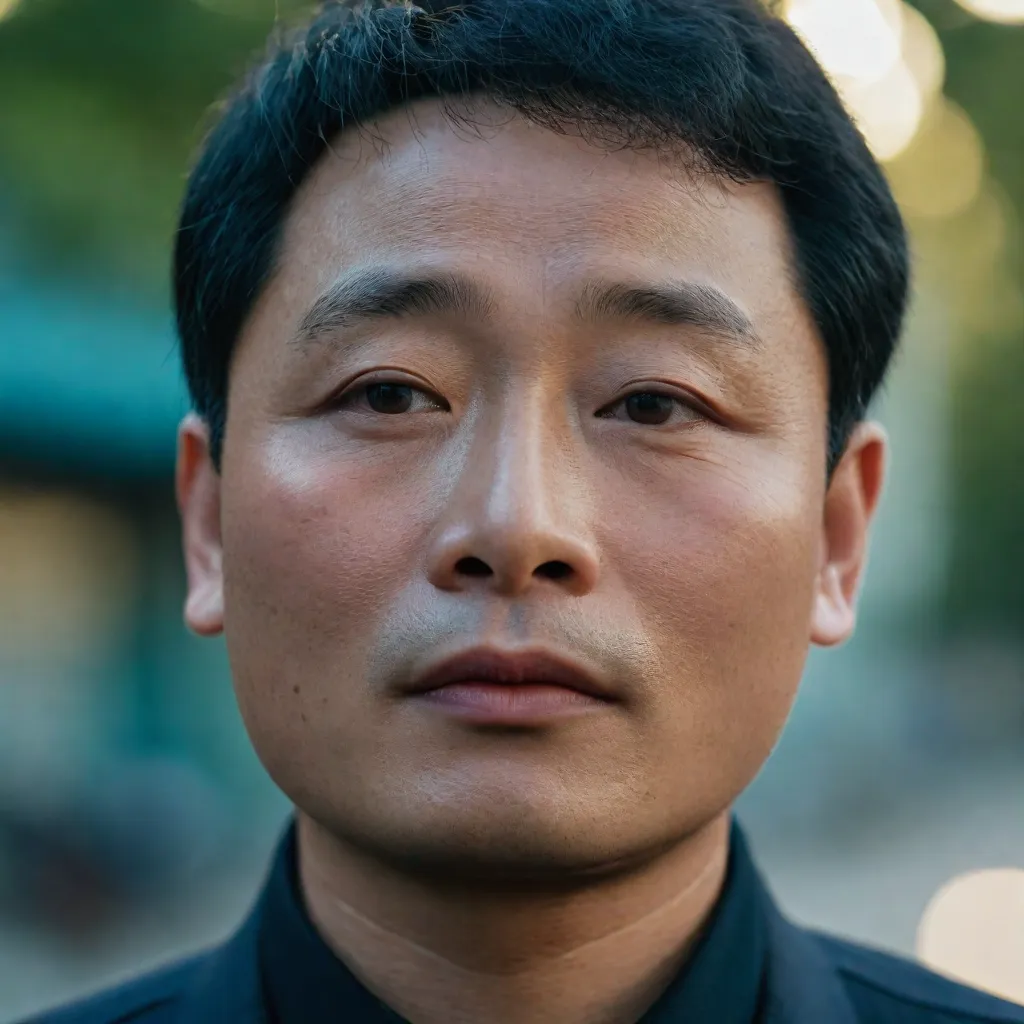}%
            \label{fig:mug_m}} \quad
        \subfloat[Augmentation from original + enhanced.]{%
            \includegraphics[width=.17\linewidth]{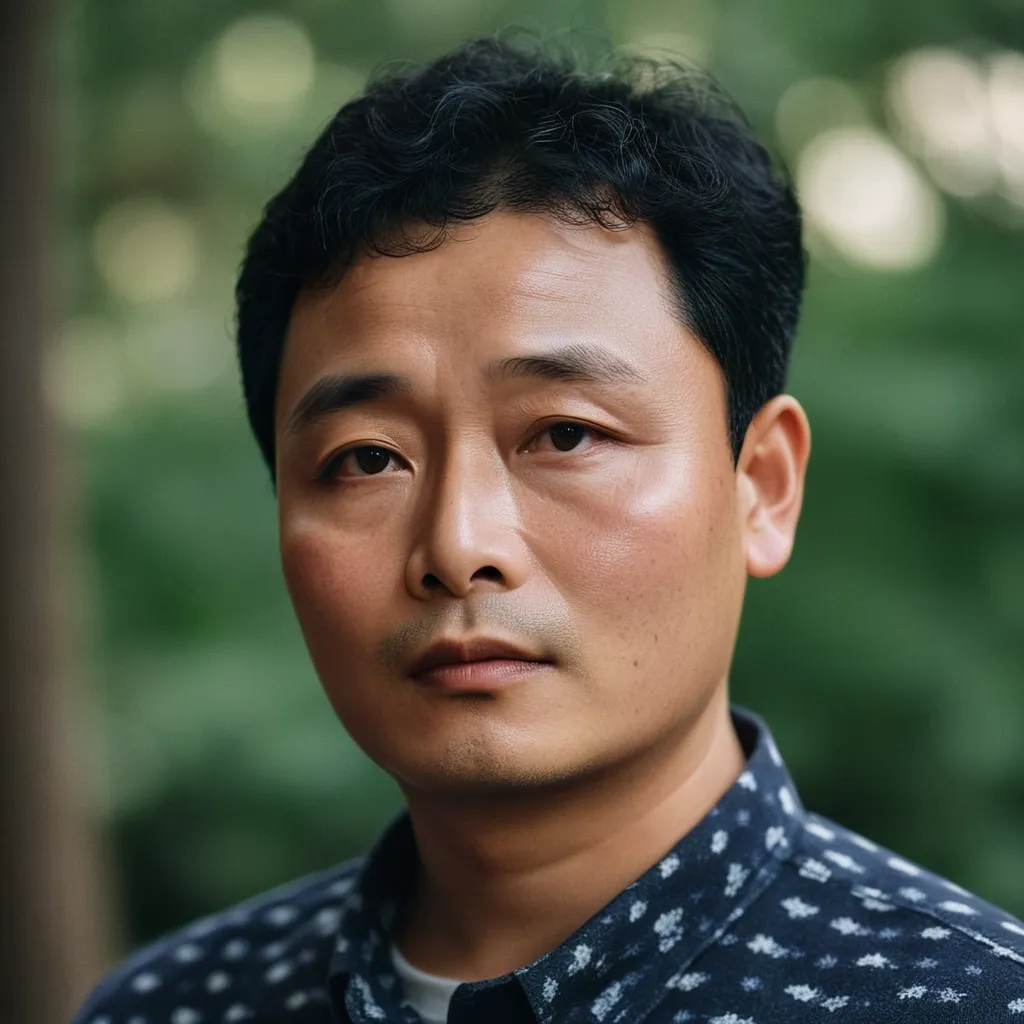}%
            \label{fig:mug_n}}\quad
            
        \caption{Additional examples of the mugshot generation task we performed. Individuals depicted represent both cases of criminal perpetrators and kidnapping victims.}
        \label{fig:mugshot_examples}
    \end{figure}

    \begin{figure}[ht]
        \centering
        \subfloat[Original image.]{%
            \includegraphics[width=.25\linewidth]{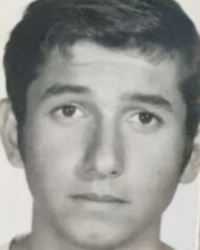}%
            \label{fig:age_a}} \quad
        \subfloat[Target image.]{%
            \includegraphics[width=.25\linewidth]{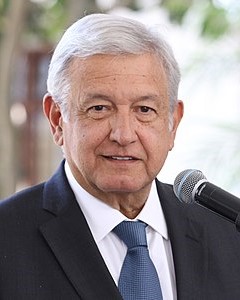}%
            \label{fig:age_b}} \quad
        \subfloat[Aged image.]{%
            \includegraphics[width=.25\linewidth]{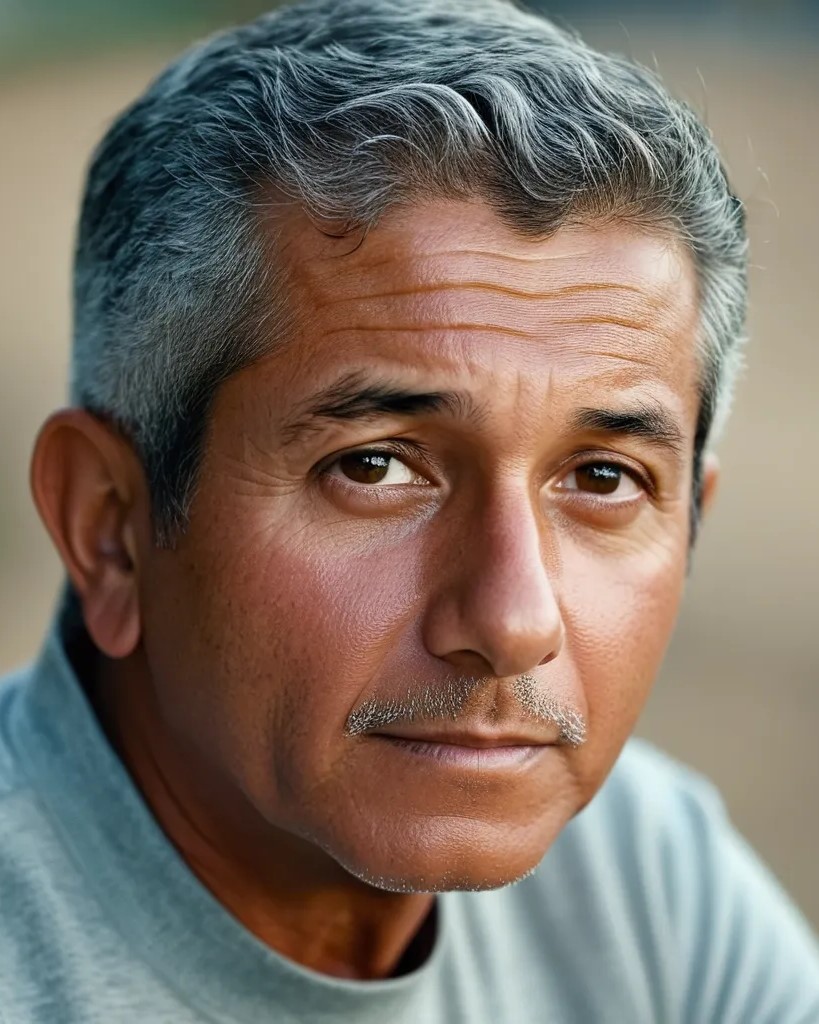}%
            \label{fig:age_c}}\quad
            
        \subfloat[Original image.]{%
            \includegraphics[width=.25\linewidth]{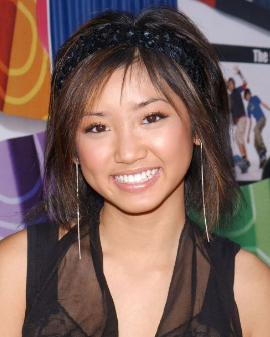}%
            \label{fig:age_d}} \quad
        \subfloat[Target image.]{%
            \includegraphics[width=.25\linewidth]{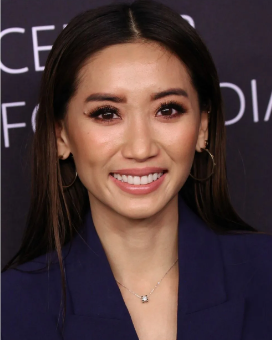}%
            \label{fig:age_e}} \quad
        \subfloat[Aged image.]{%
            \includegraphics[width=.25\linewidth]{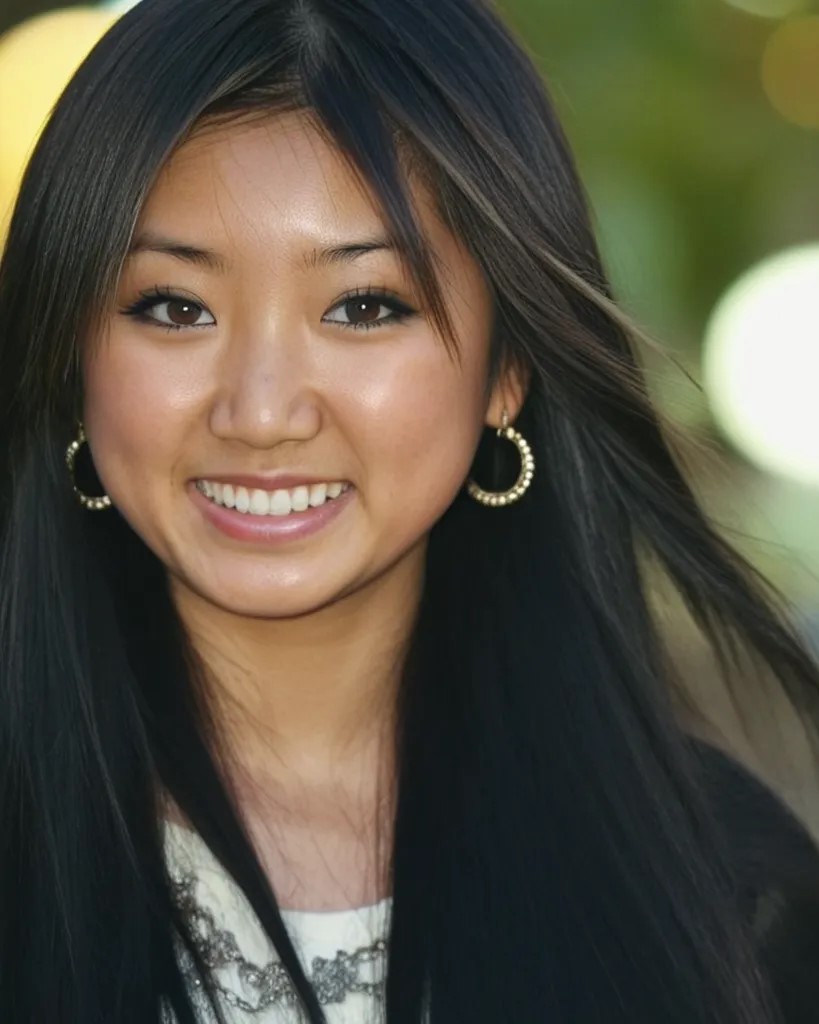}%
            \label{fig:age_f}}\quad
        \caption{Additional ageing examples}
        \label{fig:aging_unsuccessful_examples}
    \end{figure}
            
    \begin{figure}
        \centering
        \subfloat[Original image.]{%
            \includegraphics[width=.25\linewidth]{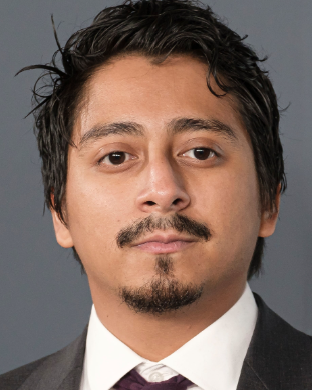}%
            \label{fig:deaged_a}} \quad
        \subfloat[Target image.]{%
            \includegraphics[width=.25\linewidth]{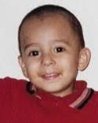}%
            \label{fig:deaged_b}} \quad
        \subfloat[De-aged image.]{%
            \includegraphics[width=.25\linewidth]{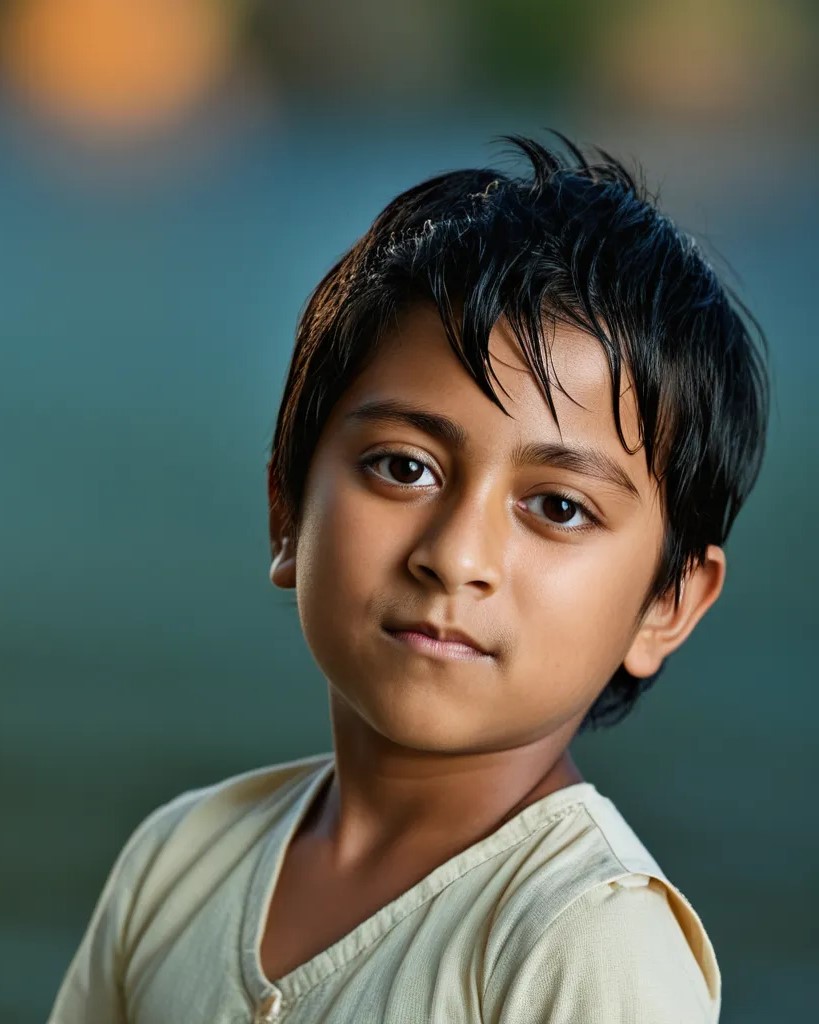}%
            \label{fig:deaged_c}}
        \caption{Example of de-aging unsuccessful test: image~\protect\subref{fig:deaged_a} represent the starting subject given in input, image~\protect\subref{fig:deaged_b} the younger subject in the real world, while image~\protect\subref{fig:deaged_c} represents the result of our tool classified as a different individual from~\protect\subref{fig:deaged_b}.}
        \label{fig:deaging}
    \end{figure}
            
\end{document}


\newpage

\onecolumn

\appendices

\section{}

In this section we introduce the \textbf{confusion matrices} computed during the comparison methods to validate and prove the similarity of our synthetic poster / augmented pictures with respect to the original photos of the suspects / celebrities.

We show a reduced version of our experimental results. The larger confusion matrices will be published and made available to researchers in the dedicated webpage.

\

\subsection{Mugshots}

\begin{figure}[ht] 
\centering
\onecolumn\includegraphics[width=1\linewidth]{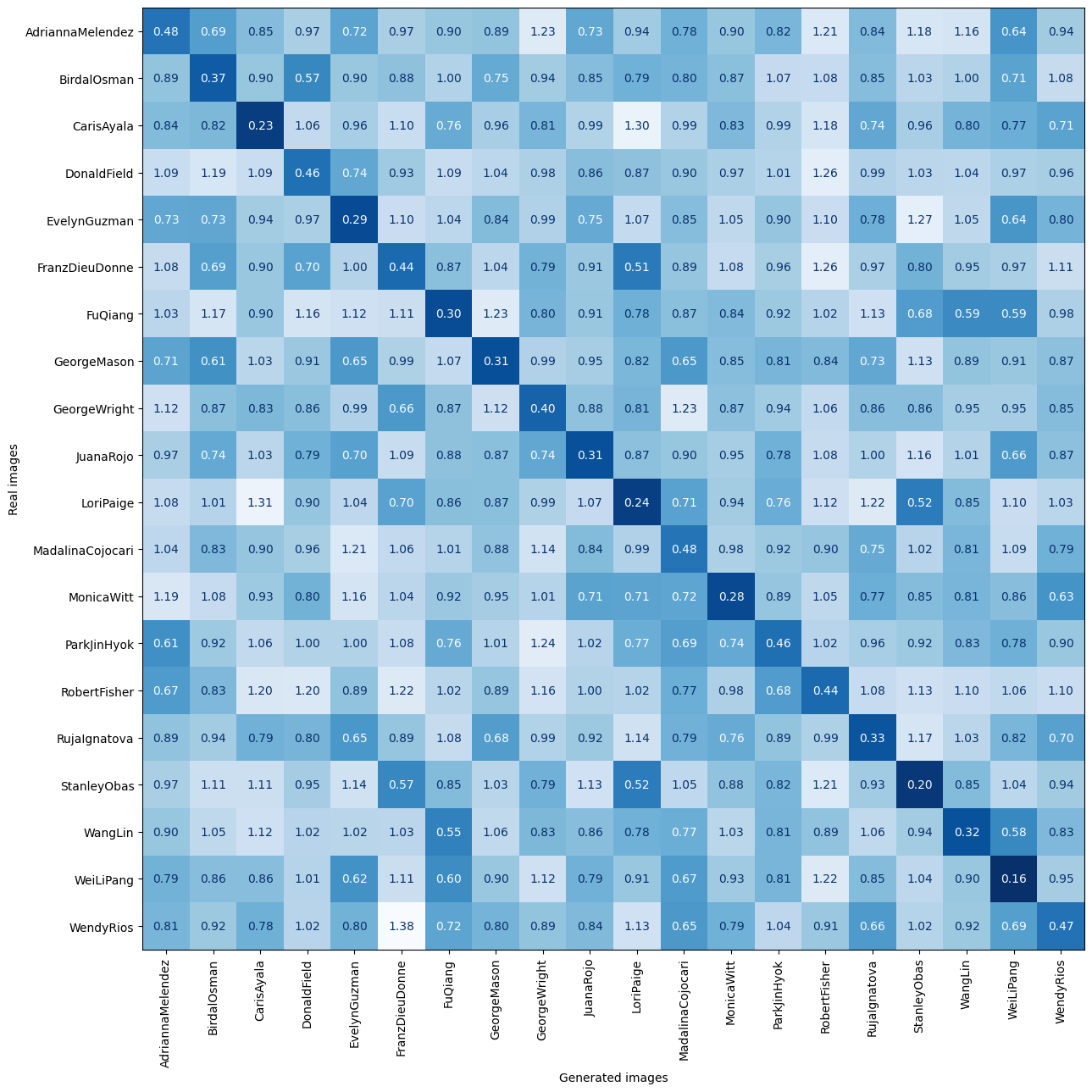}%
 \caption{Confusion matrix obtained from the results of tests performed over images generated starting from only the original mugshot. Tests were conducted using DeepFace as re-identification network. Values reported in the matrix stand for the distances between the two represented subjects, measured by DeepFace. }
                    \label{fig:mugshot_a}
\end{figure}
\newpage
\begin{figure}[ht] 
\centering
\onecolumn\includegraphics[width=1\linewidth]{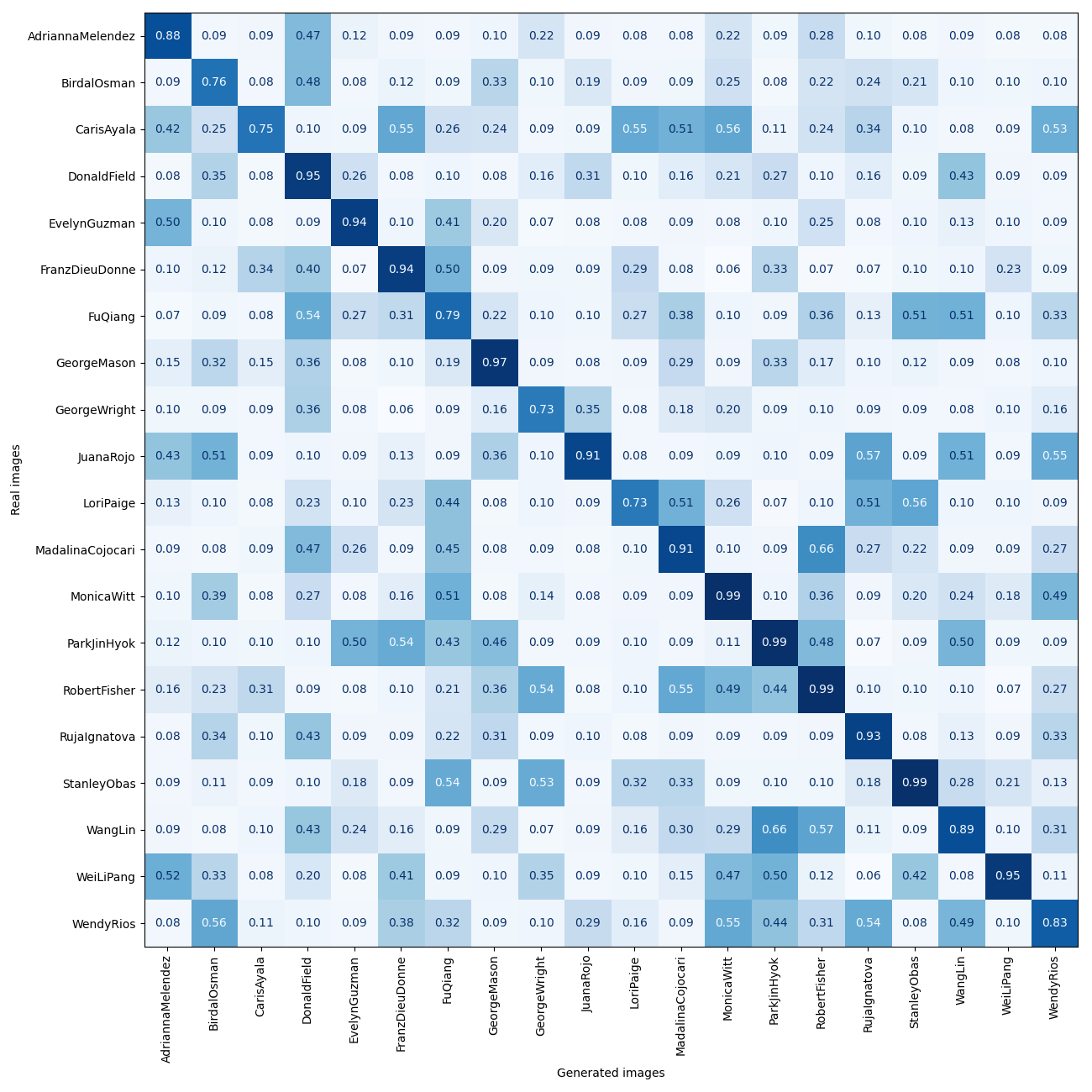}%
\caption{Confusion matrix obtained from the results of tests performed over images generated starting from only the original mugshot. Tests were conducted using Recognito as re-identification network. Values reported in the matrix stand for the similarity values between the two represented subjects, measured by Recognito.}
                    \label{fig:mugshot_b}
\end{figure}
\newpage
\begin{figure}[ht] 
\centering
\onecolumn\includegraphics[width=1\linewidth]{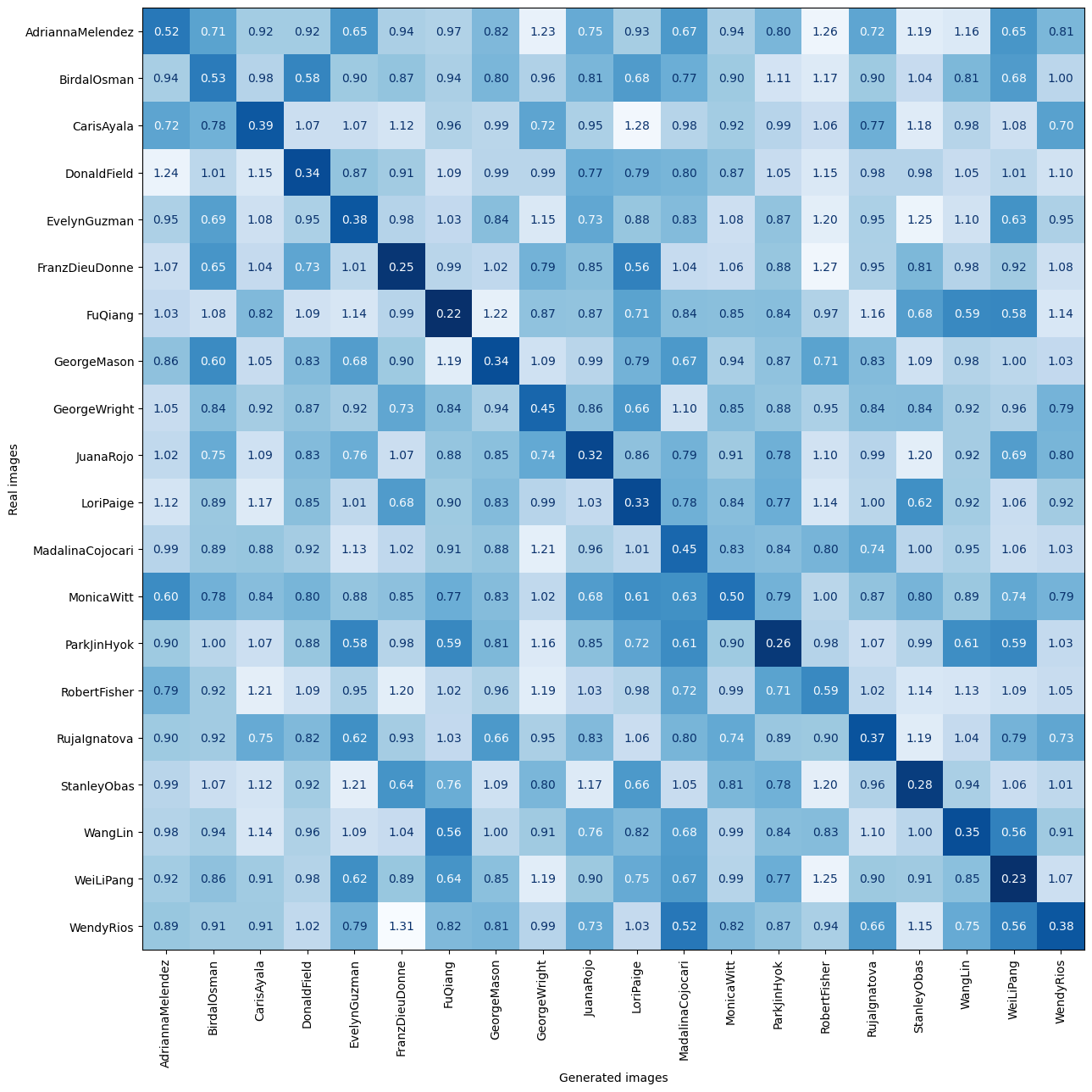}%
\caption{Confusion matrix obtained from the results of tests performed over images generated starting from a combination of the original mugshot and others previously generated. Tests were conducted using Deepface as re-identification network. Values reported in the matrix stand for the distances between the two represented subjects, measured by DeepFace.}
                    \label{fig:mugshot_c}
\end{figure}
\newpage

\subsection{Aging}

\

\begin{figure}[ht] 
\centering
\onecolumn\includegraphics[width=1\linewidth]{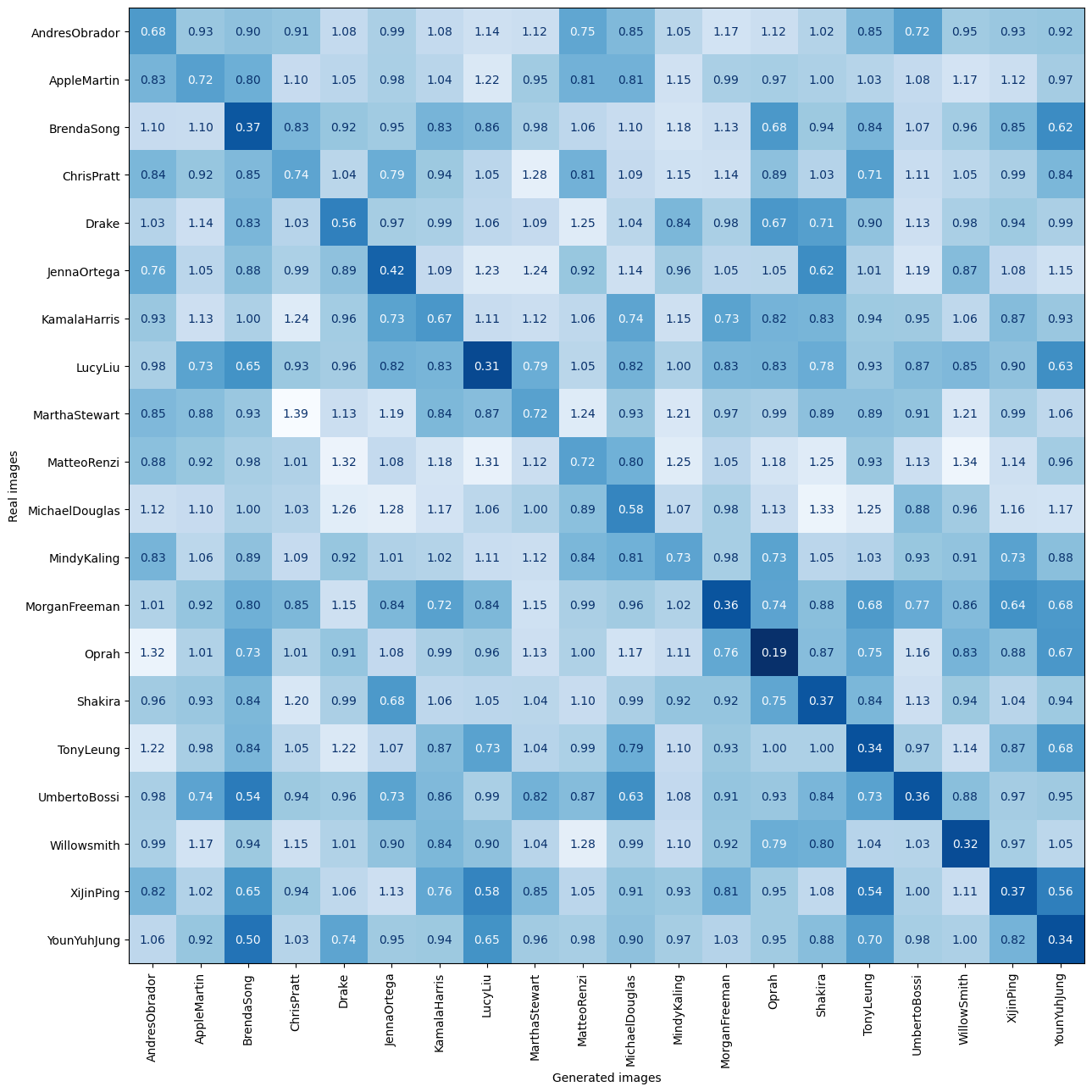}%
\caption{Confusion matrix obtained from the results of tests performed over aged images. Tests were conducted using DeepFace as re-identification network. Values reported in the matrix stand for the distances between the two represented subjects, measured by DeepFace.}
                    \label{fig:ageing_a}
\end{figure}
\newpage
\begin{figure}[ht] 
\centering
\onecolumn\includegraphics[width=1\linewidth]{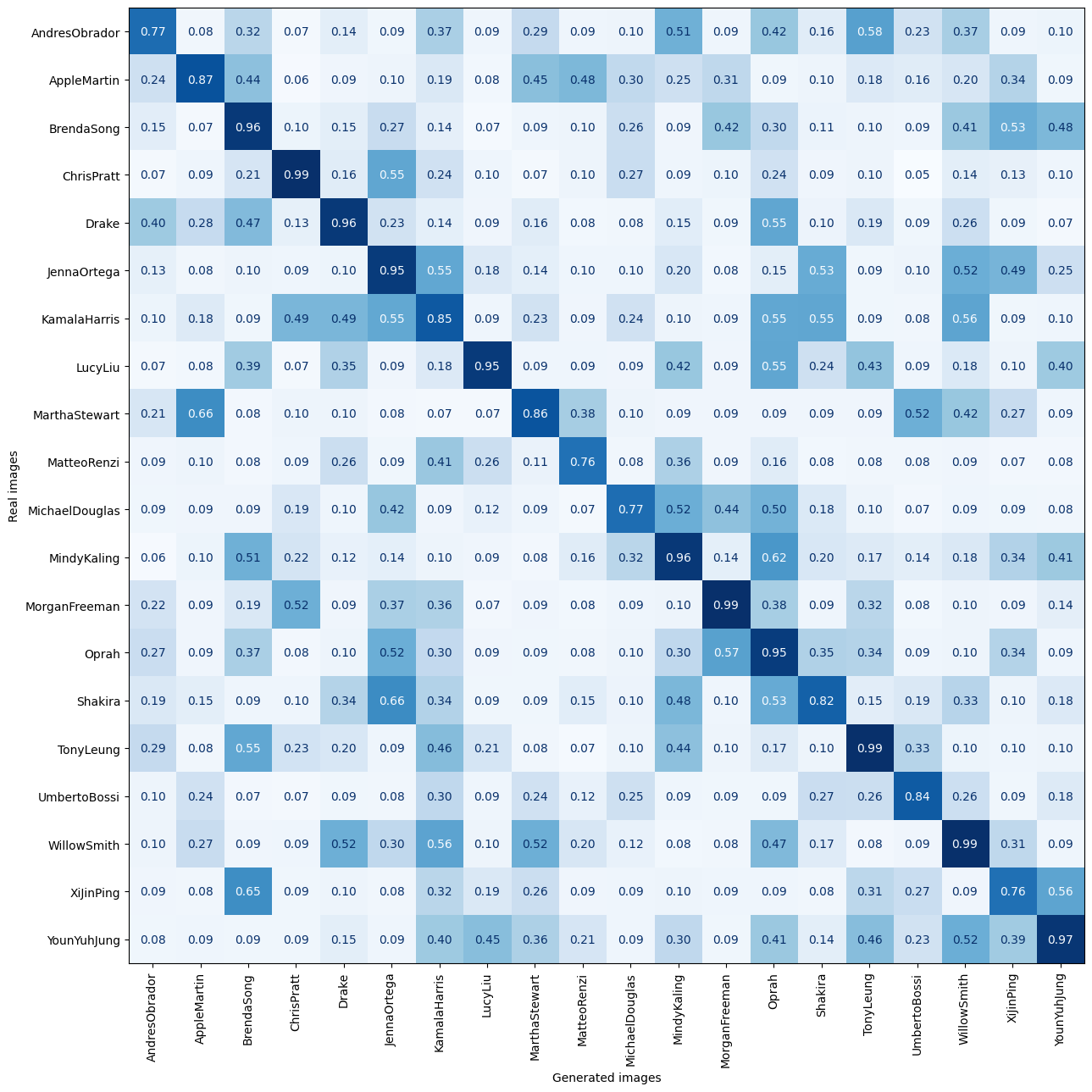}%
\caption{Confusion matrix obtained from the results of tests performed over aged images. Tests were conducted using Recognito as re-identification network. Values reported in the matrix stand for the similarity values between the two represented subjects, measured by Recognito.}
                    \label{fig:ageing_b}
\end{figure}
\newpage
\subsection{De-aging}

\begin{figure}[ht] 
\centering
\onecolumn\includegraphics[width=1\linewidth]{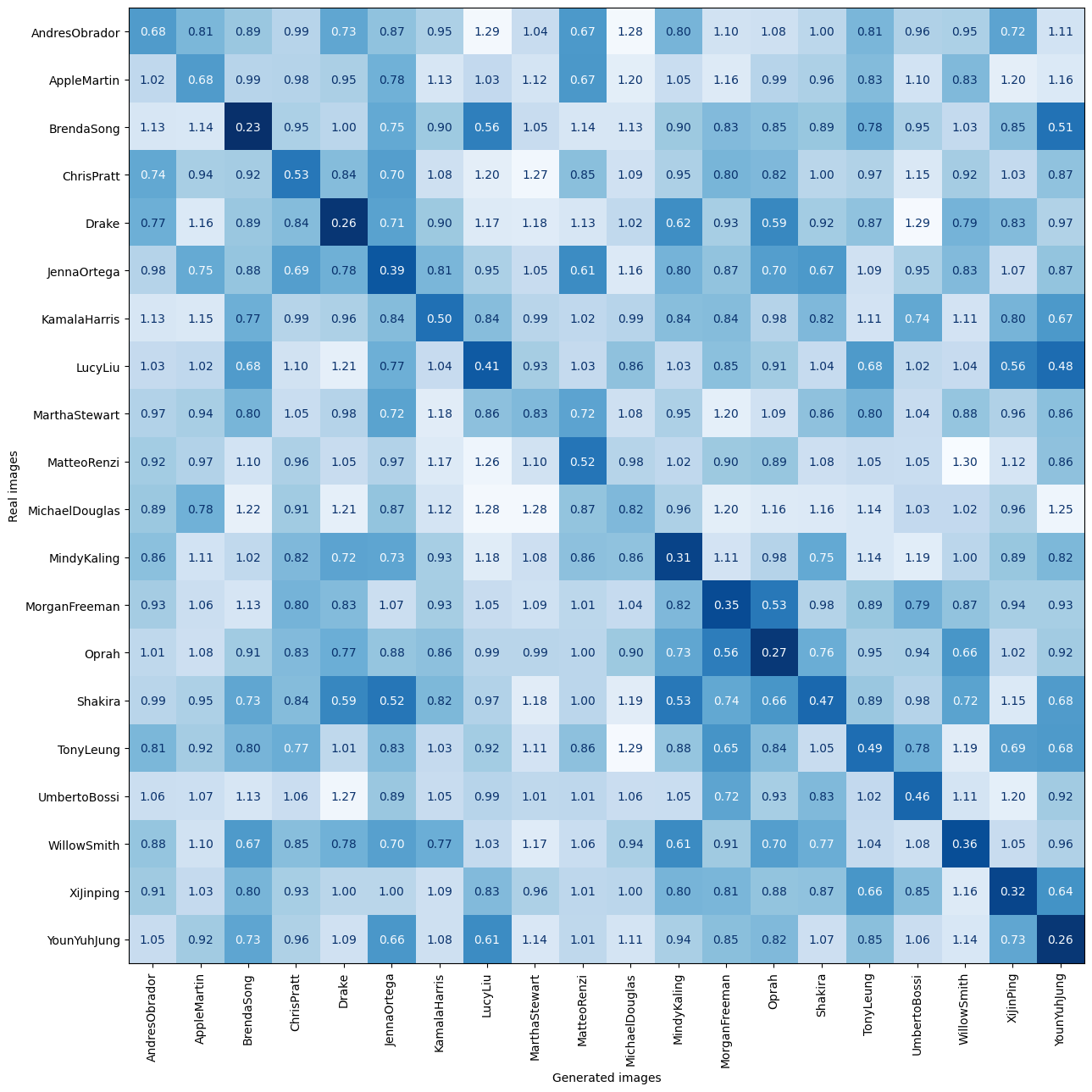}%
\caption{Confusion matrix obtained from the results of tests performed over de-aged images. Tests were conducted using DeepFace as re-identification network. Values reported in the matrix stand for the distances between the two represented subjects, measured by DeepFace.}
                    \label{fig:deageing_a}
\end{figure}
\newpage
\begin{figure}[ht] 
\centering
\onecolumn\includegraphics[width=1\linewidth]{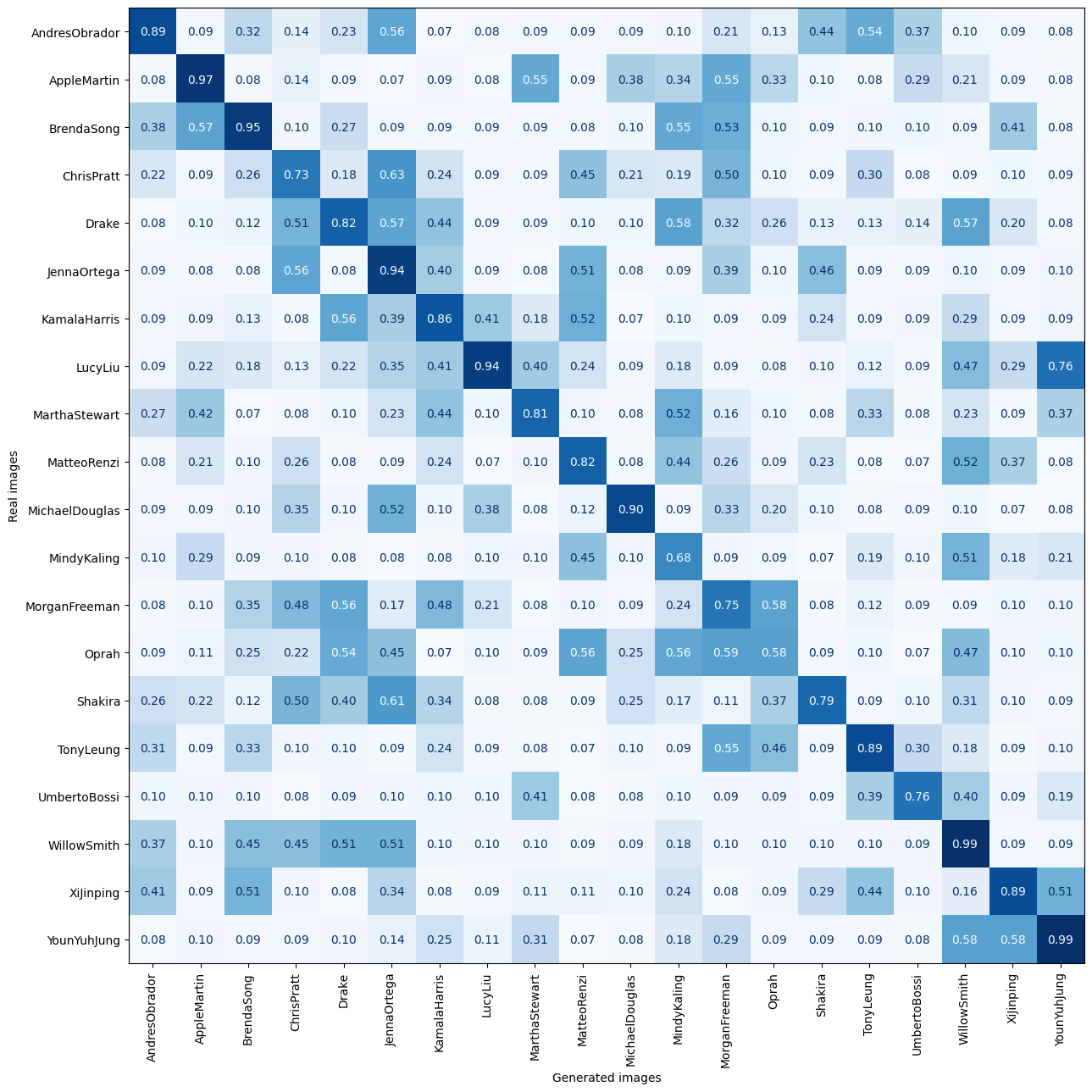}%
\caption{Confusion matrix obtained from the results of tests performed over de-aged images. Tests were conducted using Recognito as re-identification network. Values reported in the matrix stand for the similarity values between the two represented subjects, measured by Recognito.}
                    \label{fig:deageing_b}
\end{figure}

\newpage

\section{}

In this section, we show several additional \textbf{synthetic mugshots} generated using our proposed framework. These examples showcase the effectiveness of our method across a diverse range of subjects, including variations in age, ethnicity, and input image quality. The images illustrate the results of both our image enhancement techniques and the subsequent synthetic augmentation process. Even more examples will be made available in a dedicated repository upon acceptance.

\begin{figure}[h]
                \centering
                \subfloat[Original image.]{%
                    \includegraphics[width=.1275\linewidth]{images/Appendix B/BirdalOsman.jpg}%
                    \label{fig:mug_a}} \quad
                \subfloat[Augmentation from original.]{%
                    \includegraphics[width=.17\linewidth]{images/Appendix B/BirdalOsman_or.jpg}%
                    \label{fig:mug_b}} \quad
                \subfloat[Augmentation from original + enhanced.]{%
                    \includegraphics[width=.17\linewidth]{images/Appendix B/BirdalOsman_or+gen.jpg}%
                    \label{fig:mug_c}}\quad
                    
                \subfloat[Original image]{%
                    \includegraphics[width=.13\linewidth]{images/Appendix B/JuanaRojo.jpg}%
                    \label{fig:mug_d}} \quad
                \subfloat[Augmentation from original.]{%
                    \includegraphics[width=.17\linewidth]{images/Appendix B/JuanaRojo_or.jpg}%
                    \label{fig:mug_e}} \quad
                \subfloat[Augmentation from original + enhanced.]{%
                    \includegraphics[width=.17\linewidth]{images/Appendix B/JuanaRojo_or+gen.jpg}%
                    \label{fig:mug_f}}\quad
                    
                \subfloat[Original image.]{%
                    \includegraphics[width=.105\linewidth]{images/Appendix B/MadalinaCojocari.jpg}%
                    \label{fig:mug_g}} \quad
                \subfloat[Augmentation from original.]{%
                    \includegraphics[width=.17\linewidth]{images/Appendix B/MadalinaCojocari_or.jpg}%
                    \label{fig:mug_h}} \quad
                \subfloat[Augmentation from original + enhanced.]{%
                    \includegraphics[width=.17\linewidth]{images/Appendix B/MadalinaCojocari_or+gen.jpg}%
                    \label{fig:mug_i}}\quad
                    
                \subfloat[Original image.]{%
                    \includegraphics[width=.1375\linewidth]{images/Appendix B/WangLin.jpg}%
                    \label{fig:mug_l}} \quad
                \subfloat[Augmentation from original.]{%
                    \includegraphics[width=.17\linewidth]{images/Appendix B/WangLin_or.jpg}%
                    \label{fig:mug_m}} \quad
                \subfloat[Augmentation from original + enhanced.]{%
                    \includegraphics[width=.17\linewidth]{images/Appendix B/WangLin_or+gen.jpg}%
                    \label{fig:mug_n}}\quad
                    
                \caption{Additional examples of the mugshot generation task we performed. Individuals depicted represent both cases of criminal perpetrators and kidnapping victims.}
                \label{fig:aging_examples}
            \end{figure}

\begin{figure}[h]
                \centering
                \subfloat[Original image.]{%
                    \includegraphics[width=.25\linewidth]{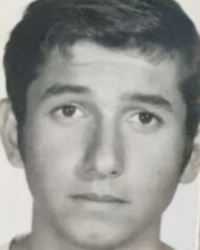}%
                    \label{fig:age_a}} \quad
                \subfloat[Target image.]{%
                    \includegraphics[width=.25\linewidth]{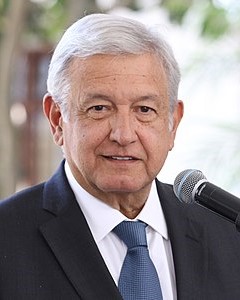}%
                    \label{fig:age_b}} \quad
                \subfloat[Aged image.]{%
                    \includegraphics[width=.25\linewidth]{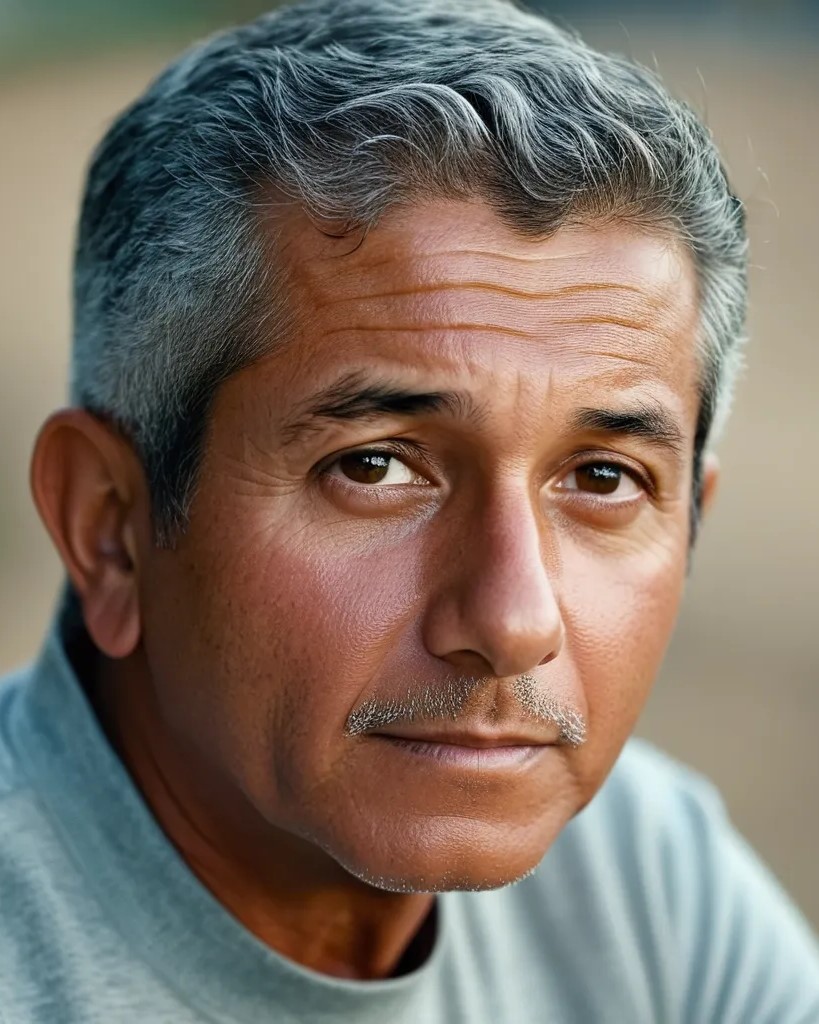}%
                    \label{fig:age_c}}\quad
                    
                \subfloat[Original image.]{%
                    \includegraphics[width=.25\linewidth]{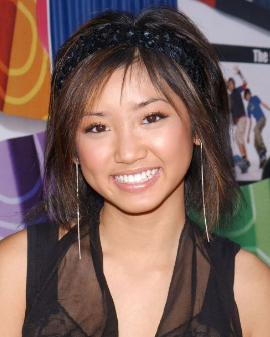}%
                    \label{fig:age_d}} \quad
                \subfloat[Target image.]{%
                    \includegraphics[width=.25\linewidth]{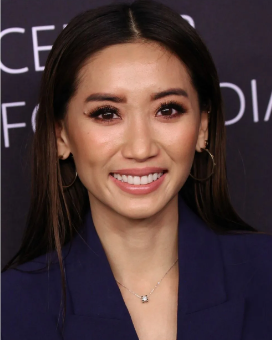}%
                    \label{fig:age_e}} \quad
                \subfloat[Aged image.]{%
                    \includegraphics[width=.25\linewidth]{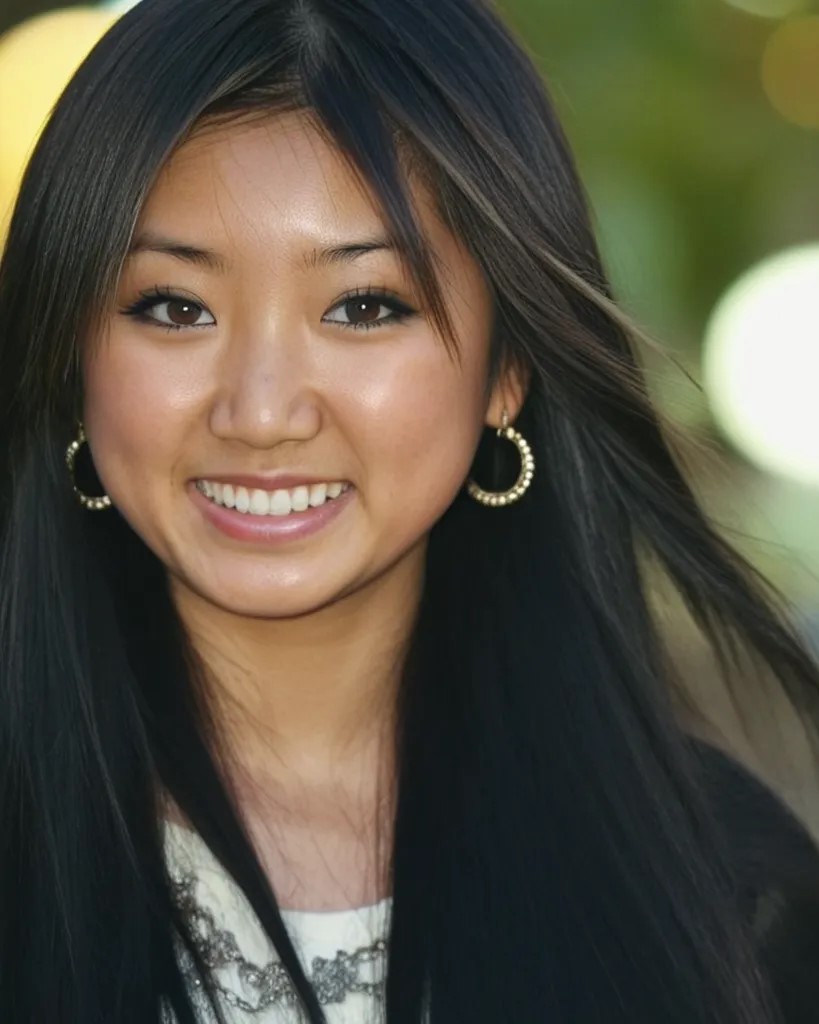}%
                    \label{fig:age_f}}\quad
\caption{Additional ageing examples}
                \label{fig:aging_examples}
            \end{figure}

    \begin{figure}
                \centering
                \subfloat[Original image.]{%
                    \includegraphics[width=.25\linewidth]{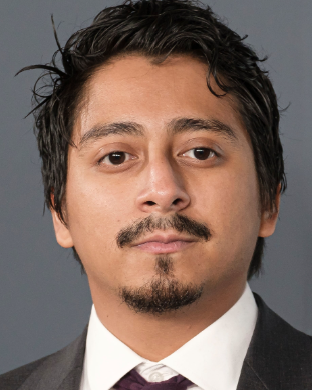}%
                    \label{fig:deaged_a}} \quad
                \subfloat[Target image.]{%
                    \includegraphics[width=.25\linewidth]{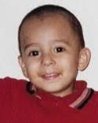}%
                    \label{fig:deaged_b}} \quad
                \subfloat[De-aged image.]{%
                    \includegraphics[width=.25\linewidth]{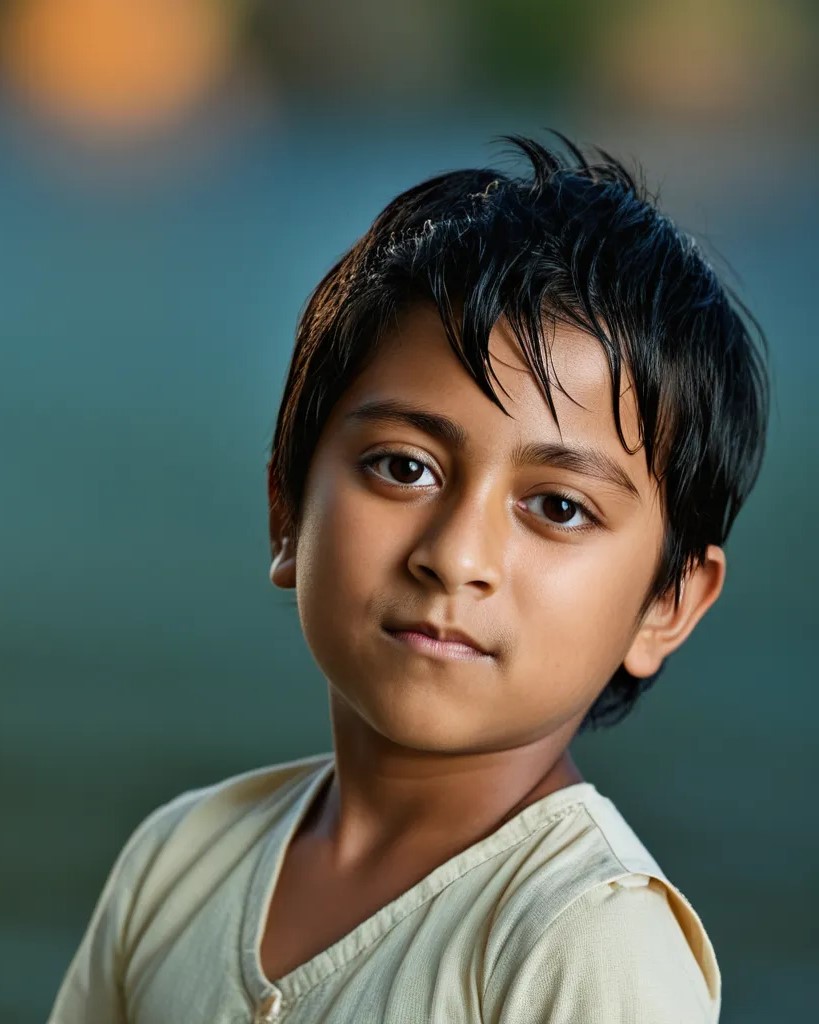}%
                    \label{fig:deaged_c}}
                \caption{Example of de-aging unsuccessful test: image~\protect\subref{fig:deaged_a} represent the starting subject given in input, image~\protect\subref{fig:deaged_b} the younger subject in the real world, while image~\protect\subref{fig:deaged_c} represents the result of our tool classified as a different individual from~\protect\subref{fig:deaged_b}.}
                \label{fig:deaging}
            \end{figure}